\documentclass[lettersize,journal]{IEEEtran}

\usepackage{amsmath,amsfonts}
\usepackage{array}
\usepackage{textcomp}
\usepackage{stfloats}
\usepackage{url}
\usepackage{verbatim}
\usepackage{graphicx}
\usepackage{cite}
\usepackage{multirow}

\usepackage[compact]{titlesec} 

\usepackage[T1]{fontenc}
\usepackage{mathtools}
\usepackage{balance}
\usepackage{amssymb}
\usepackage{lineno}
\usepackage{times}
\usepackage{soul}
\usepackage{pifont}
\usepackage[hidelinks]{hyperref}
\usepackage[utf8]{inputenc}
\usepackage{graphicx, wrapfig, subcaption, setspace, booktabs}
\usepackage{booktabs}
\usepackage{microtype}
\usepackage{xcolor}
\usepackage{enumitem}
\usepackage{algorithm, algpseudocode}
\usepackage{amsthm}
\usepackage[symbol]{footmisc}
\usepackage[normalem]{ulem}

\newcolumntype{P}[1]{>{\centering\arraybackslash}p{#1}}
\newcommand{\RNum}[1]{\uppercase\expandafter{\romannumeral #1\relax}}
\newcommand{\cmark}{\ding{51}}%
\newcommand{\xmark}{\ding{55}}%

\theoremstyle{definition}

\newtheorem*{assumption*}{\assumptionnumber}
\providecommand{\assumptionnumber}{}
\makeatletter

\DeclarePairedDelimiter{\abs}{\lvert}{\rvert}%
\DeclarePairedDelimiterX{\inp}[2]{\langle}{\rangle}{#1, #2}
\makeatletter
\let\oldabs\abs
\def\abs{\@ifstar{\oldabs}{\oldabs*}}
\ifCLASSOPTIONcompsoc
  \usepackage[nocompress]{cite}
\else
  \usepackage{cite}
\fi
\ifCLASSINFOpdf
\else
\fi

\begin{document}

%\title{Beyond The Evidence Lower Bound: \\A Contrastive Variational Graph Auto-Encoder For Node Clustering}
\title{A Contrastive Variational Graph Auto-Encoder\\ for Node Clustering}

\author{Nairouz~Mrabah,~
        Mohamed~Bouguessa,~
        Riadh~Ksantini
\IEEEcompsocitemizethanks{\IEEEcompsocthanksitem N. Mrabah, M. Bouguessa are with the Department
of Computer Science, University of Quebec at Montreal, Montreal, QC, Canada.
\protect\\
E-mails: mrabah.nairouz@courrier.uqam.ca, ~bouguessa.mohamed@uqam.ca
\IEEEcompsocthanksitem R. Ksantini is with the Department of Computer Science, College of IT, University of Bahrain, Kingdom of Bahrain.
\protect\\
E-mail: rksantini@uob.edu.bh}
}
%\author{IEEE Publication Technology,~\IEEEmembership{Staff,~IEEE,}
        % <-this % stops a space
%\thanks{This paper was produced by the IEEE Publication Technology Group. They are in Piscataway, NJ.}% <-this % stops a space
%\thanks{Manuscript received April 19, 2021; revised August 16, 2021.}}



\maketitle

\begin{abstract}
%Variational graph auto-encoders (VGAEs) have shown strong potential in various graph-related applications. Some recent VGAEs incorporate the clustering inductive bias in a principled way by selecting a multimodal prior distribution. %introducing a clustering objective within the evidence lower bound (ELBO).

Variational Graph Auto-Encoders (VGAEs) have been widely used to solve the node clustering task. However, the state-of-the-art methods have numerous challenges. First, existing VGAEs do not account for the discrepancy between the inference and generative models after incorporating the clustering inductive bias. Second, current models are prone to degenerate solutions that make the latent codes match the prior independently of the input signal (i.e., Posterior Collapse). Third, existing VGAEs overlook the effect of the noisy clustering assignments (i.e., Feature Randomness) and the impact of the strong trade-off between clustering and reconstruction (i.e., Feature Drift). To address these problems, we formulate a variational lower bound in a contrastive setting. Our lower bound is a tighter approximation of the log-likelihood function than the corresponding Evidence Lower BOund (ELBO). Thanks to a newly identified term, our lower bound can escape Posterior Collapse and has more flexibility to account for the difference between the inference and generative models. Additionally, our solution has two mechanisms to control the trade-off between Feature Randomness and Feature Drift. Extensive experiments show that the proposed method achieves state-of-the-art clustering results on several datasets. We provide strong evidence that this improvement is attributed to four aspects: integrating contrastive learning and alleviating Feature Randomness, Feature Drift, and Posterior Collapse. 

%Variational graph auto-encoders (VGAEs) have shown strong potential in various graph-related applications. Some recent VGAEs incorporate the clustering inductive bias in a principled way by introducing a clustering objective within the evidence lower bound (ELBO). However, these methods do not account for the discrepancy between the inference and generative models after integrating the clustering meta-prior. They suffer from degenerate solutions, where the latent codes match the prior independently from the corresponding input (i.e., Posterior Collapse). Furthermore, they lack the capacity to reduce the effect of the noisy clustering assignments (i.e., Feature Randomness) and the effect of the strong trade-off between clustering and reconstruction (i.e. Feature Drift). To address these problems, we propose a tighter variational lower bound (compared with the corresponding ELBO) that establishes a contrastive learning framework. Thanks to a newly-identified term, our lower bound can escape Posterior Collapse and has more flexibility to account for the difference between the true and variational posteriors. Additionally, our solution incorporates two mechanisms to control the trade-off between Feature Randomness and Feature Drift. Extensive experiments show that the proposed method achieves state-of-the-art clustering results on several datasets. We provide strong evidence that the obtained improvement is attributed to four aspects: integrating a contrastive learning strategy, alleviating Feature Randomness, Feature Drift, and Posterior Collapse. 

\end{abstract}


\section{Introduction}

Abundant graph data remains largely unlabeled in the wild. Unlike supervised learning, unsupervised methods relinquish the need for expensive data labeling to train Graph Neural Networks (GNNs). In this context, deep clustering has emerged as a new paradigm for performing joint clustering and embedded feature learning. %by exploiting the large amounts of unlabeled samples. 
%This paradigm has achieved promising performance \cite{paper50, paper51} in clustering large-scale, high-dimensional, and high-semantic image datasets such as ImageNet. %The recent progress in performance can be mainly explained by the inductive bias inherent to the over-parametrized neural networks \cite{paper52}, which makes them capture semantic similarities and hence learn generalizable hypotheses. 
In essence, the deep clustering paradigm resorts to two main strategies to offset the absence of training labels: self-supervision and pseudo-supervision. The main idea of self-supervision is to perform a well-designed pretext task that involves learning high-level representations. Unlike self-supervision, pseudo-supervision focuses on solving the primary task by identifying the semantic categories of the input data. More precisely, it constructs pseudo-labels by applying a clustering algorithm to the embedded representations and then leverages the obtained labels to train the model. 

% Pseudo-supervision techniques and limitations in the context of the deep clustering paradigm (FR)
Neural networks can perfectly fit random labels without significant time overhead, as shown by Zhang et al. \cite{paper40}. Therefore, spotting the false labels among the full set of pseudo-labels during the training process is a challenging task. Furthermore, some recent work \cite{paper105} has observed that it is not possible to revoke the effect of noisy labels after training with them, even if the model is retrained with the true labels uniquely. The accumulated error associated with pseudo-supervision causes Feature Randomness (FR) \cite{paper39}.

% Self-supervision techniques: Random walk, adjacency reconstruction, mutual information maximization, and data likelihood maximization. 
A possible solution for alleviating the effect of FR consists of adjusting the gradient of the pseudo-supervised loss using self-supervision. Graph self-supervision learning relies on several strategies. Random walk \cite{paper4, paper42, paper43}, adjacency reconstruction \cite{paper1, paper44}, and mutual information maximization \cite{paper3, paper34, paper35} are the dominant self-supervision techniques for node clustering. All of these strategies can be unified under the pairwise learning framework, which captures high-level similarities by training an encoder to project similar pairs close to each other. Contrastive learning is a specific case of the pairwise learning framework; it has the ability to push dissimilar pairs far apart. In \cite{paper45}, the authors have reformulated several deep metric learning objectives %, including the triplet loss and the softmax cross-entropy loss,
based on a unified pairwise loss. In \cite{paper46}, the authors have connected a variant of pairwise losses through bound relationships and have shown the link between these pairwise losses and mutual information. We classify existing self-supervision methods for node clustering into four categories according to the granularity of pairwise learning: node level, proximity level, cluster level, and graph level.

% Node-level self-supervised learning and limitations (employs heavy and task-specific data augmentation techniques)
% Instance-level contrastive learning prepares positive pairs by building two augmented views of the same image and leverages the remaining images for building the negative pairs.
Inspired by the success of instance-level contrastive learning in computer vision \cite{paper58}, node-level contrastive learning establishes the same idea for graph datasets. It proceeds by pulling together the latent codes of the same node in two views. The new graphs are obtained by applying augmentation techniques. However, graph augmentation methods provide limited semantic invariance (they may behave arbitrarily and change graph semantics) \cite{paper33}, are highly dependent on the dataset \cite{paper57}, and are very sensitive to the hyperparameters of the augmentation scheme \cite{paper57}. Furthermore, contrastive learning at the node level overlooks cluster-level information by treating all nodes except the target node as negatives.

% Neighbor-level self-supervised learning and limitations (over-emphasize the proximity information within the embedded representation of each node)
Random walk and adjacency reconstruction are proximity-level self-supervised methods that encode the representations of neighbor nodes (not necessarily first-order neighbors) close to each other and push the representations of the other nodes far away. In our previous work \cite{paper8}, we have shown that adjacency reconstruction amounts to a pairwise loss weighted by the graph adjacency coefficients and a regularization term. 
%we write the reconstruction loss of a GAE model in the form of a linear combination between a graph Laplacian regularization term.
%we have shown that minimizing the adjacency reconstruction amounts to minimizing a combination between a pairwise loss weighted by the coefficients of the graph adjacency matrix and a regularization term to escape degenerate solutions. 
Similarly, performing short random walks captures the proximity structural properties of the graph. For example, DeepWalk \cite{paper4} makes the nodes on the same short walk have similar embeddings, and contrasts these representations with the ones of the other nodes using noise-contrastive estimation \cite{paper54}. %In \cite{paper47}, the authors have identified the metric space behind the random-walk-based representations. They have shown that the Flow distance defined on the open-flow network is equivalent to the Euclidean distance between the embedded representations generated by random-walk-based approaches. 
However, random walk and adjacency reconstruction suffer from well-known limitations. Both strategies overemphasize proximity information that might be captured by the graph convolution operation. Moreover, random-walk-based methods are highly sensitive to the hyper-parameter choice \cite{paper4, paper42}. %We distance ourselves from random walks and the typical adjacency reconstruction by deriving an alternative contrastive objective that constitutes a novel variational lower bound of the graph likelihood.

% Graph-level self-supervised learning (makes the node embeddings mindful of the global properties of the graph) and limitations (over-emphasize the global information (i.e., the global graph summary using the readout function) within the embedded representation of each node and overlook the fine-grained)
Graph-level contrastive learning builds a graph summary using a readout function. Then, it leverages this summary to assist the encoder in building latent codes, mindful of the global characteristics of the graph. %The graph-level representation is derived by averaging all node embeddings. 
The main process of this category consists of maximizing the similarity between each node and the graph-level outline and contrasting this outline with the perturbed latent codes. However, this strategy overlooks the fine-grained structure. For the clustering task, it is crucial to learn clustering-oriented representations \cite{paper12}.

% Discuss the limitation of node-level, proximity-level, global-level self-supervised learning compared with cluster-level self-supervised learning (are not designed to learn clustering-oriented features, FD)
Node-level, proximity-level, and graph-level self-supervised techniques are insufficient to solve the clustering task. Instead, these strategies are used to pretrain the model. They are also used as an auxiliary objective to reduce the effect of FR. However, combining pseudo-supervision and self-supervision leads to a strong trade-off between the two tasks. The discriminative features learned by pseudo-supervision are vulnerable to the drifting effect caused by self-supervision. This problem is referred to as Feature Drift (FD) \cite{paper39}. For instance, the clustering objective aims to decrease each cluster variance, while the node-level task treats all nodes within the same cluster as negatives. Furthermore, the strong competition between pseudo-supervision and self-supervision increases the sensitivity to the balancing hyper-parameter \cite{paper39}.
%Node-level, proximity-level, and graph-level self-supervised techniques are insufficient to solve the clustering task. Instead of that, these strategies are used for pretraining the model and also as an auxiliary objective to reduce the effect of FR. However, combining pseudo-supervision and self-supervision leads to a strong trade-off between the two tasks. More precisely, the discriminative features learned by pseudo-supervision are vulnerable to the drifting effect caused by self-supervision. This problem is referred to as Feature Drift (FD) \cite{paper39}. For instance, the clustering objective aims at decreasing the cluster-variance, while the node-level contrastive learning treats all nodes within the same cluster as negatives. Furthermore, the strong competition between pseudo-supervision and self-supervision increases the sensitivity to the balancing hyper-parameter \cite{paper39}.

% Variational auto-encoders 
Unlike several self-supervised techniques, Variational Graph Autoencoders (VGAEs) have a solid probabilistic foundation. %The general framework of VGAEs consists of maximizing the ELBO (i.e. Evidence Lower BOund) of the input data log-likelihood. 
Since the posterior distribution is intractable, variational methods are applied to approximate the true posterior. %To this end, 
A differentiable and tractable variational posterior parameterized by an encoding network is introduced to derive the ELBO. %of the log-likelihood function. 
The typical lower bounds of existing VGAE models generally amount to a reconstruction and a regularization term. Regularization is expressed based on Kullback-Leibler (KL) divergence between the prior and the variational posterior.
%Maximizing the ELBO amounts to minimizing the reconstruction and another regularization term expressed based on the Kullback Leibler (KL) divergence between the prior and the variational posterior.

% Limitations of ELBO (Unrealistic hypothesis related to the lower bound, Posterior collapse, lacks the ability to exploit the advancement achieved by contrastive learning)
Unfortunately, the learning capacity of the ELBO is limited by the Posterior Collapse (PC) problem \cite{paper59}. Minimizing the regularization term makes the variational posterior match the prior independently from the input variable, which in turn leads to a degenerate local minimum. In this case, the decoder learns to reconstruct the input from completely random noise. Another limitation of the ELBO is that maximizing this term is equivalent to minimizing the KL divergence between the true and variational posteriors. This is a strong hypothesis because the variational posterior is limited by: (1) the encoding flexibility, (2) the restriction to the family of conditionally conjugate exponential distributions to ensure tractability, and (3) the requirement to match the prior distribution during the training process as induced by the regularization term. Therefore, we assume that a good variational lower bound should account for the difference between the true and variational posteriors. Last but not least, the ELBO objective fails to establish a contrastive learning framework, which is the main strategy for the most recent unsupervised methods.

% Our proposed approach and advantages over existing Variational graph auto-encoders:
% 1) Contrastive learning derived in a principled way to maintain an adequate lower bound of the data likelihood.
% 2) Escapes Posterior collapse: more encoding flexibility because the posterior distribution of each latent sample is no longer restricted to match a fixed prior distribution. 
% 3) Accounts for the difference between the true posterior and the variational posterior (discriminative-- clustering-oriented suitable for inference).
The previous VGAEs \cite{paper26, paper27, paper13} leverage Jensen's inequality to obtain the ELBO. In this work, we propose a tighter variational lower bound. More precisely, we formulate a term \textit{tractable} and \textit{differentiable} term that can account for the difference between the inference and generative models. This new term is derived from the quantity discarded by applying Jensen's inequality (i.e., the KL divergence between the true and variational posteriors). Therefore, it can distance the variational posterior (discriminative aspect of the clustering task) from the true posterior (generative aspect). It also establishes a contrastive learning framework by introducing negative samples. Another interesting aspect of our approach is the ability to handle the PC problem. Typical VAEs do not have any mechanism to avoid PC. For instance, the authors of \cite{paper91} have proposed combining mutual information with ELBO to avoid this problem. By introducing an external loss function, the final objective of \cite{paper91} is no longer guaranteed to be a lower bound of data log-likelihood. Furthermore, an additional hyper-parameter is required in \cite{paper91} to control the trade-off between the two components. In our case, we show empirically that our newly discovered term helps escape the PC problem. %Intuitively, maximizing the between-cluster variance avoids the case where the latent points collapse to the Gaussian random noise.

% Our proposed approach and advantages over existing contrastive learning methods:
% 1) Our approach has mechanisms against FR and FD, which are derived in a principled way within the variational framework to ensure an appropriate lower-bound of the data likelihood.
% 2) Unlike the node-level, proximity-level, and graph-level contrastive learning, our approach offers a cluster-level contrastive learning framework. Clustering-friendly embedded representations.
Node-level, proximity-level, and graph-level contrastive learning strategies fail to consider cluster-level information. To this end, we propose a variational auto-encoder that gradually goes from proximity-level to cluster-level learning. We progressively form latent clusters based on a clustering-oriented graph constructed during the training process, and we contrast the learned representations with the ones from an anchor graph. Furthermore, we equip the proposed model with two mechanisms to alleviate the effect of FR and FD. In our previous work \cite{paper8}, we have proposed to supply graph auto-encoders with two operators that can control the trade-off between FR and FD. %However, the proposed operators are not theoretically justified. 
In this work, we demonstrate that these operators can be analytically derived within a variational framework that maximizes the graph log-likelihood. Other mechanisms for controlling FR and FD can be naturally integrated into our variational framework while still optimizing the same objective function (i.e., graph likelihood). The merits of our method can be described from three perspectives: 

\begin{itemize}
\item \textbf{Variational graph auto-encoders}: We propose a novel variational lower bound of the graph log-likelihood that generalizes the ELBO. Our solution has three advantages compared to existing VGAEs. First, our lower bound lines up with contrastive learning. Second, it constitutes a tighter lower bound than the corresponding ELBO. Third, the new term in our lower bound can alleviate the PC problem and gives more flexibility to account for the difference between the true and variational posteriors.
\item \textbf{Contrastive learning}: We propose a clustering-oriented contrastive learning approach that deviates from the node-level, proximity-level, and graph-level methods. Our model has two operators to alleviate the effect of FR and FD. These operators are analytically derived under a variational framework that maximizes the log-likelihood function.
\item \textbf{Experiments}: We conducted extensive experiments to explore the benefits of our method compared to (1) existing VGAE models and (2) contrastive learning strategies. Our results provide strong evidence that the proposed method brings considerable improvement in clustering effectiveness and alleviates the FR, FD and PC problems.
\end{itemize}

%Furthermore, we establish the theoretical link between variational auto-encoders and the most recent deep metric learning approaches by introducing pairwise learning within the variational framework. Referring to the previous point, our approach can be seen as a contrastive method, which explicitly pulls similar embedded pairs close to each other while pushing dissimilar ones far away. A brief comparison between our proposed approach BELBO-VGAE (Beyond the Evidence Lower BOund Variational Graph Auto-Encoder) and the most prominent VGAE models is provided in Table \ref{table:variational_autoencoders}.

\section{Related Work} \label{sec_2}
%Our model proposes a unified view of variational auto-encoders and contrastive learning. To this end, we start by discussing the most relevant unsupervised and contrastive learning techniques for the clustering task. We shed the light on the limitations of these methods to highlight the contributions of our approach. Then, we discuss the advantages of the VGAEs over the other unsupervised graph representation methods, and we criticize their limitations from the ELBO perspective.
In line with the focus of this work, we review the previous VGAEs and contrastive learning methods.  

\subsection{Unsupervised Graph Representation Learning}

% Node-level unsupervised learning methods: GRACE, GCA, BGRL
GRACE \cite{paper34} is one of the most prominent graph contrastive learning methods. First, it constructs two augmented views of the input data by performing corruption at the level of the network topology and the node attributes. The adopted augmentation scheme for this method includes edge dropping and feature masking. Then, a lower bound of mutual information between the encoded views is maximized using the noise-contrastive estimator (a.k.a. InfoNCE \cite{paper92}). Based on the same objective function, GCA \cite{paper35} improves GRACE performance by replacing the random augmentation scheme with an adaptive one. The adaptive scheme aims to circumvent the destruction of important edges and features. Unlike GRACE and GCA, BGRL \cite{paper32} does not require the use of negative samples to avoid degenerate solutions, making it more scalable than the other node-level methods.   

% Proximity-level unsupervised learning methods: GAE, ARGE, MGAE, AGC, GALA, GMI
Graph augmentation schemes provide limited semantic invariance \cite{paper33}. Unlike node-level contrastive learning methods, proximity-level techniques do not require data augmentation. The similarities are captured from the neighbors rather than using a perturbed representation of the same node. Among the fundamental approaches of this category, GAE \cite{paper1} minimizes the reconstruction of the adjacency matrix. ARGE \cite{paper2} has an additional adversarial training scheme to make latent codes match a Gaussian distribution. MGAE \cite{paper7} makes the reconstruction objective more robust to input perturbations by marginalizing the corrupted features. GALA \cite{paper10} has a symmetric decoder that reconstructs the input features using a Laplacian sharpening operation on the graph structure. AGC \cite{paper17} extracts proximity-level representations by applying a k-order graph convolutional operation. DeepWalk \cite{paper4} performs short truncated random walks and builds latent codes using the contrastive skip-gram algorithm \cite{paper93}. GMI \cite{paper38} maximizes mutual information between each latent code of each node and an associated support graph that only considers proximity information of the corresponding node. The authors of COLES \cite{paper38} perform a reformulation of Laplacian eigenmaps in a contrastive setting. However, proximity-level methods are insufficient to capture cluster-level information. To overcome this limitation, our approach gradually transforms the proximity-level objective into a clustering-oriented contrastive learning task. This aspect is the key to alleviating FD.

% Graph-level unsupervised learning methods: DGI, MVGRL, COLES
%Graph-level contrastive learning captures the similarity between the local and global representations from the same graph, and pushes away the representations from other graphs.
%Graph-level techniques captures the similarity between the local and global representations. 
Among graph-level methods, DGI \cite{paper3} maximizes a mutual information objective using Jensen-Shannon divergence. In particular, a discriminator network learns to increase the similarity score between local and global representations of the same graph and to decrease the similarity score between global and local representations of different graphs. Based on a similar objective function, MVGRL \cite{paper31} improves DGI by performing graph diffusion and subgraph sampling to obtain better views for local and global representations. However, the graph-level methods push the node representations near the same latent code (i.e., graph summary), which in turn reduces the between-cluster variance.

% Hybrid including the cluster-level unsupervised learning methods: GIC, SDCN, DAEGC, AGE, AFGRL, R-DGAE 
The cluster-level task is generally associated with another self-supervised task from a higher or lower level of abstraction to reduce FR. For instance, GIC \cite{paper12} combines the graph-level objective of DGI with a cluster-level contrastive task. SDCN \cite{paper21} outsources a clustering-oriented task with proximity-level information by exploiting the high-order structure and adjacency reconstruction. DAEGC \cite{paper9} harnesses an attention-based encoder to assess the importance of neighboring nodes, and establishes a training that involves joint clustering and reconstruction. AGE \cite{paper11} integrates proximity-level information by exploiting a Laplacian smoothing filter, then trains a fully-connected neural network to construct a clustering-oriented graph. AFGRL \cite{paper33} introduces a filtering mechanism to reduce the number of false positive neighbors discovered by K-NN for the proximity-level task and trains the model based on a clustering-oriented loss. R-DGAE \cite{paper8} advocates the integration of two operators into the clustering-reconstruction loss to reduce the effect of FR and FD. However, these operators are not theoretically derived within a variational framework.

\subsection{Variational Graph Auto-Encoders}

% Advantages of the VGAE models
Unlike the methods discussed in the previous subsection, VGAEs %have strong probabilistic foundations. They  
do not assume a deterministic encoding process. Instead, they account for the uncertainty with random variables from predefined distributions parameterized by neural networks. Furthermore, VGAEs are among the most relevant graph generative models and are widely used for important applications such as the generation of molecules \cite{paper94}. Additionally, VGAEs can easily incorporate the inductive bias clustering in a principled way by changing the prior distribution \cite{paper95}. This aspect avoids an unjustified combination between different losses and the incumbent hyper-parameters to balance between them. 

% Encoding flexibility: N-VGAE, D-VGAE, S-VGAE, GMM-VGAE, SIG-VGAE
We start by discussing existing VGAEs based on their flexibility for encoding and incorporating different inductive biases. For example, $\mathcal{N}$-VGAE \cite{paper1} imposes Gaussian distributions on the prior and variational posterior. %On the one hand, the variational posterior has restricted capacity when the true posterior violates the Gaussian assumption. 
On the one hand, some methods attempt to enhance the encoding flexibility by learning non-Gaussian variational posteriors. %To solve this problem, 
$\mathcal{D}$-VGAE \cite{paper27} leverages the Dirichlet distribution to model the variational posterior. SIG-VGAE \cite{paper28} does not specify an explicit distribution and relies on semi-implicit variational inference to increase the expressiveness of the variational posterior. %On the other hand, the Gaussian prior restricts the capacity to capture the clustering inductive bias. 
On the other hand, some methods try non-Gaussian priors to capture the clustering inductive bias. $\mathcal{S}$-VGAE \cite{paper26} proposes a von Mises-Fisher distribution instead of the Gaussian assumption, which leads to a uniform prior on a hypersphere suitable for data clustering. GMM-VGAE \cite{paper13} captures the clustering meta-prior by imposing Gaussian mixture models on the prior. However, all these models do not establish a contrastive learning framework, do not account for the difference between the true and variational posteriors after introducing the clustering inductive bias, and neglect the PC problem.

% Decoding flexibility: GRAPHITE-VGAE, GDC-VGAE 
Some other VGAEs focus on improving the decoder. Instead of using the inner product, GRAPHITE \cite{paper29} uses a multilayer iterative decoding process. In GDN-VAE \cite{paper30}, the authors construct the decoder using graph deconvolutional operations. If the expressiveness of the decoder is improved without addressing the PC problem, then it is easier to learn the decoding distribution independently of the latent codes \cite{paper104}.

%For instance, GRAPHITE \cite{paper29} relies on a multi-layer iterative decoding process instead of the simple inner-product operation. GDN-VGAE \cite{paper30} proposes a graph deconvolutional operation to build the decoder. However, improving the decoder expressiveness without tackling PC makes the model learn the output distribution by ignoring the latent codes.

\section{Proposed Approach}
Our model proposes a unified view of contrastive learning and variational auto-encoders %beyond the typical ELBO 
for maximizing the data likelihood. For a matrix $M$, the expression $M[i,:]$ denotes the $i^{th}$ row, and $M[:,j]$ denotes the $j^{th}$ column of this matrix. Given an undirected attributed graph $\mathcal{G} = (\mathcal{V}, \mathcal{E}, X)$, where $\mathcal{V} = \left \{v_{1}, v_{2}, ..., v_{N} \right \}$ is a set of $N$ nodes, $\mathcal{E} = \left \{ e_{ij}, 1 \leqslant i,j \leqslant N \right \}$ is the set of edges, and $X \in \mathbb{R}^{N \times J}$ is the feature matrix. $x_{i} \in \mathbb{R}^{J}$ denotes the characteristic vector associated with the $i^{th}$ node, and $J$ is the input space dimension. The topology of $\mathcal{G}$ is captured by the adjacency matrix $A = \left ( a_{ij} \right ) \in \mathbb{R}^{N \times N}$, such that $a_{ij} = 1$ if $(v_{i}, v_{j}) \in \mathcal{E}$ and $a_{ij} = 0$ otherwise. We denote $D$ as the degree matrix of $A$, such that $D = \text{diag}(d_{1}, ..., d_{N})$, where $d_{i} = \sum_{j=1}^{N} A_{ij}$. We consider that the nodes of $\mathcal{G}$ can be grouped into $K$ clusters. The vector $C \in \left \{ 1, ..., K \right \}^{N}$ defines the cluster membership of the different nodes such that $c_{i}$ represents the cluster of the $i^{\text{th}}$ node. 

Our approach CVGAE (Contrastive Variational Graph Auto-Encoder) has two components: a graph convolutional encoder \cite{paper49} denoted by $E$ and a decoder denoted by $U$. The encoder maps the input graph to a low-dimensional matrix $Z \in \mathbb{R}^{N \times d}$, where $d$ is the dimension of the latent space. The expression of the embedded representations is computed according to the layer-wise propagation rule:

\begin{equation}
\resizebox{.91\linewidth}{!}{$
    \displaystyle
    Z^{(l)} = f_{\phi}(Z^{(l-1)}, \, A \, | \, W^{(l)}) =  \phi(\widetilde{D}^{-\frac{1}{2}}\widetilde{A}\widetilde{D}^{-\frac{1}{2}}Z^{(l-1)}W^{(l)}),
$}
\label{eq:1}
\end{equation}

\noindent where $Z^{(l)}$ and $W^{(l)}$ denote the output and the weight matrix of the $l^{th}$ layer, respectively, and $Z^{(0)} = X$. The function $\phi$ represents the activation of this layer. Laplacian smoothing is defined using the matrices $\widetilde{A}$ and $\widetilde{D}$, such that $\widetilde{A} = A + I_{n}$ and $\widetilde{D} = D + I_{n}$. Let $W=\left \{ W^{(l)} \right \}$ be the set of all trainable weights. Our encoder consists of two layers specified as:

\begin{equation}
Z^{(1)} = f_{\text{ReLU}}(X, \, A \, | \, W^{(1)}),
\label{eq:2}   
\end{equation}

\begin{equation}
Z^{(2)}_{\mu} = f_{\text{Linear}}(Z^{(1)}, \, A \, | \, W^{(2)}_{\mu}) \in  \mathbb{R}^{N \times d},
\label{eq:3}   
\end{equation}

\begin{equation}
Z^{(2)}_{\sigma} = f_{\text{Linear}}(Z^{(1)}, \, A \, | \, W^{(2)}_{\sigma}) \in \mathbb{R}^{N \times d}.
\label{eq:4}   
\end{equation}

Our model has two training phases. The first phase consists of maximizing an ELBO similar to $\mathcal{N}$-VGAE. We provide a full description of the pretraining phase to make this work self-contained. However, our contributions are related to the clustering phase. The optimized bounds and the inference and generative models are specified in the following subsections.

\subsection{Pretraining}
We assume that the graph topology can be generated from a random process involving the distribution $p(A^{gen}, \: Z)$, where $A^{gen}$ is the generated graph structure. This distribution factorizes as $p(A^{gen}, \, Z) = p(A^{gen} \, | \, Z) \, p(Z)$ such that: 

\begin{equation}
p(Z) = \prod_{i=1}^{N}  p(z_{i}) = \prod_{i=1}^{N} \mathcal{N}(z_{i} \, | \, \textbf{0}, \, I),
\label{eq:5}
\end{equation}

\begin{equation}
p(A^{gen}|Z) = \prod_{i=1}^{N} \prod_{j=1}^{N} p(a_{ij}^{gen} \, | \, z_{i}, \, z_{j}) = \prod_{i=1}^{N} \prod_{j=1}^{N} \mathcal{B}\text{er}(\beta_{ij}),
\label{eq:6}
\end{equation}

\noindent where $p(a_{ij}^{gen} \, | \, z_{i}, \, z_{j})$ is a Bernoulli distribution $\mathcal{B}\text{er}(\beta_{ij})$ parameterized by $\beta_{ij} = \text{Sigmoid}(z_{i}^{T}z_{j})$ for each generated edge. For the pretraining phase, we set $A^{gen}=A$. The prior distribution $p(Z)$ can be factorized according to the independence of the latent variables, so that each $p(z_{i})$ represents a Gaussian distribution $\mathcal{N}(z_{i} \, | \, \textbf{0}, \, I)$.

The general framework of variational auto-encoders consists of maximizing the log-likelihood of the input data. Since the posterior distribution is intractable, typical variational methods aim to approximate the true posterior. Specifically, a well-defined variational posterior $q(Z \, | \, X, \, A)$ parameterized by an encoding network forms the inference model and is used for this approximation. The distribution $q(Z \, | \, X, \, A)$ is described:

\begin{equation}
\resizebox{.91\linewidth}{!}{$
    \displaystyle
    q(Z \, | \, X, \, A) = \prod_{i=1}^{N} q(z_{i} \, | \, X, \, A) = \prod_{i=1}^{N} \mathcal{N}(z_{i} \, | \, \mu_{z_{i}}, \, \text{diag}(\sigma^{2}_{z_{i}})),
$}
\label{eq:7}
\end{equation}

\noindent where $\mu_{z_{i}} = Z^{(2)}_{\mu}[i,:]$ represents the mean vector of the multivariate Gaussian distribution corresponding to $z_{i}$ and $\sigma^{2}_{z_{i}} = Z^{(2)}_{\sigma}[i,:]$ captures the variance of this distribution. 

Based on the factorization and specification provided for the inference and generative distributions $q(Z \, | \, X, \, A)$ and $p(A^{gen}, \, Z)$, respectively, the generated graph log-likelihood can be expressed as:

\begin{equation}
\resizebox{.89\linewidth}{!}{$
\begin{split}
\text{log}\big(p(A^{gen})\big) &= \sum_{i,j=1}^{N} KL\big(q(z_{i}, \, z_{j} \,| \, X, \, A) \: || \: p(z_{i}, \, z_{j} \, | \, a_{ij}^{gen})\big) \\
&+ \mathcal{L}_{\text{ELBO}}^{(pre)}(X, \, A),
\end{split}
$}
\label{eq:8}
\end{equation}

\begin{equation}
\resizebox{.89\linewidth}{!}{$
\begin{split}
\mathcal{L}_{\text{ELBO}}^{(pre)}(X, \, A) &= \sum_{i,j=1}^{N} \mathbb{E}_{\substack{z_{i}, \, z_{j} \sim q(. \, | \, X, \, A)}} \bigg[ \text{log}\big(p(a_{ij}^{gen} \, | \, z_{i}, \, z_{j})\big) \bigg] \\ 
&- 2 \, N \, \sum_{i=1}^{N} KL\big(q(z_{i} \, | \, X, \, A) \, || \, p(z_{i})\big),
\end{split}
$}
\label{eq:9}
\end{equation}

\noindent where $\mathcal{L}_{\text{ELBO}}^{(pre)}(X, \: A)$ represents the ELBO for the pretraining phase. We provide the full derivation of Eq. (\ref{eq:8}) and Eq. (\ref{eq:9}) in Appendix A (all Appendices are provided in the Supplementary Material \url{https://github.com/nairouz/CVGAE_PR}). The first term of $\mathcal{L}_{\text{ELBO}}^{(pre)}(X, \: A)$ represents the reconstruction of the adjacency matrix. The second term of $\mathcal{L}_{\text{ELBO}}^{(pre)}(X, \: A)$ represents a regularization to impose a specific structure on the latent space, which in turn ensures straightforward sampling for generative tasks. The term $KL\big(q(z_{i}, z_{j} | X, A) \, || \, p(z_{i}, z_{j} | a_{ij}^{gen}) \big)$ absorbs the difference between the true and variational posteriors. Since the KL divergence is always positive, we can see that $\mathcal{L}_{\text{ELBO}}^{(pre)}(X, \: A)$ is a lower bound of $\text{log}\big(p(A^{gen})\big)$. 

%Furthermore, there is an equality between $\mathcal{L}_{ELBO}^{(1)}$ and $\text{log}\big(p(A)\big)$ only when the KL divergence between the true and variational posteriors is equal to zero. In fact, the true posterior can be inferred (although intractable) from the distribution $p(A, \: Z)$ associated with the generative process. However, the variational posterior is limited to a specific family of distributions (i.e., the family of conditionally conjugate exponential distributions) and should capture the discriminative aspect for the clustering task (i.e., reducing within-cluster variance and increasing between-cluster-variance). This aspect is problematic for the existing models. To this end, our clustering phase accounts for the term ignored by $\mathcal{L}_{ELBO}^{(1)}$, which is important for learning a posterior distribution cable of capturing the information destroyed by the discriminative process associated with variational posterior.

\subsection{Clustering}
The first phase learns a variational posterior that approximates the true posterior without enforcing the clustering structures. However, the introduction of clustering inductive bias leads to a discrepancy between the inference and generative models. To address this problem, we introduce a second training phase to make the variational posterior destroy the non-discriminative information without affecting the decoding process. More precisely, we derive a new term that absorbs the difference between inference and generative models. Based on this term, the decoding process can restore the non-discriminative information destroyed by the encoding process. To achieve this goal, we start by presenting the inference and generative models for the clustering phase. Then, we derive our lower bound formulation.

Our generative model is set to capture the joint distribution $p(A^{gen}, \, Z, \, C)$. Compared to the pretraining phase, this model has two heads to separate the graph generation and clustering task. Accordingly, the joint distribution is factorized as $p(A^{gen}, \, Z, \, C) = p(A^{gen} \, | \, Z) \, p(C \, | \, Z) \, p(Z)$, where $p(Z)$ and $p(A^{gen} \, | \, Z)$ maintain the same distributions as in the pretraining phase according to Eq. (\ref{eq:5}) and Eq. (\ref{eq:6}), respectively. The distribution $p(C \, | \, Z)$ is defined as follows:

\begin{equation}
p(C \, | \, Z) = \prod_{i=1}^{N} p(c_{i} \, | \, z_{i}),
\label{eq:10}
\end{equation}

\begin{equation} 
p_{ij} = \frac{(1 + \left \| z_{i} - \Omega_{j} \right \|^{2} )^{-1}}{\sum_{j'}(1 + \left \| z_{i} - \Omega_{j'} \right \|^{2})^{-1}},
\label{eq:11}
\end{equation}

\noindent where $p_{ij}$ is equal to the probability $p(c_{i}=j \, | \, z_{i})$ and $\left \{ \Omega_{j} \right \}_{j=1}^{K}$ represents the set of trainable centers. Clustering centers are initialized by K-means. The distribution $p(c_{i} \, | \,  z_{i})$ computes the soft clustering assignments. We use Student's t distribution \cite{paper96} to assess the similarity between latent codes and clustering centers in Eq. (\ref{eq:11}).

The inference model captures the distribution $q(Z, \, C \, | \, X, \, A)$. Compared with the pretraining phase, our model infers the cluster of each node in addition to the latent code. Accordingly, the joint distribution factorizes as $q(Z, \, C \, | \, X, \, A) = q(Z \,  | \, X, \, A) \, q(C \, | \, Z)$, where $q(Z \,  | \, X, \, A)$ preserves the pretraining distribution defined in Eq. (\ref{eq:7}). The distribution $q(C \, | \, Z)$ is described as follows:

\begin{equation}
q(C \,| \,Z) = \prod_{i=1}^{N} q(c_{i} \, | \, z_{i}),
\label{eq:12}
\end{equation}

\begin{equation} 
q_{ij} = \left\{\begin{matrix}
1 \; \text{if} \; (\lambda_{i}^{1}-\lambda_{i}^{2}) \geq  \alpha \; \text{and} \; j = \text{arg}\max_{_{j'}}(p_{ij'}),
\\ 
0 \; \text{if} \; (\lambda_{i}^{1}-\lambda_{i}^{2}) \geq  \alpha \; \text{and} \; j \neq \ \text{arg}\max_{_{j'}}(p_{ij'}),
\\
p_{ij} \; \text{if} \; (\lambda_{i}^{1}-\lambda_{i}^{2}) <  \alpha,
\end{matrix}\right.
\label{eq:13}
\end{equation}

\noindent where $q_{ij}$ is equal to the probability $q(c_{i}=j \, | \, z_{i})$. The distribution $q(c_{i} \, | \,  z_{i})$ strengthens the discriminative aspect by transforming soft clustering assignments into hard assignments. It gradually decreases the within-cluster variance and increases the between-cluster variance. Specifically, we compare the difference between the first and second high-confidence assignment scores of $p(c_{i} \, | \,  z_{i})$ with a hyper-parameter $\alpha$ to gradually construct the hard clustering assignments of $q(c_{i} \, | \,  z_{i})$. The first and second scores associated with the $i^{th}$ node are denoted by $\lambda_{i}^{1}$ and $\lambda_{i}^{2}$, respectively, and are defined:

\begin{equation}
\begin{aligned}
\lambda_{i}^{1} = \max_{_{j \in \left \{ 1,...,K \right \}}} \left \{p_{ij} \right \},
\end{aligned}
\label{eq:14}
\end{equation}

\begin{equation}
\begin{aligned}
\lambda_{i}^{2} = \max_{_{j \in \left \{ 1,...,K \right \}}} \left \{p_{ij} \,\, | \,\, p_{ij} < \lambda_{i}^{1} \right \}.
\end{aligned}
\label{eq:15}
\end{equation}

Based on the factorization and specification provided for the inference and generative distributions $q(Z, \, C | \, X, \, A)$ and $p(A^{gen}, \, Z)$, respectively, we find that the ELBO for the clustering phase $\mathcal{L}_{\text{ELBO}}^{(clus)}(X, \, A)$ is expressed as follows (the complete derivation is provided in Appendix B):

\begin{equation}
\resizebox{.89\linewidth}{!}{$
\begin{split}
\mathcal{L}_{\text{ELBO}}^{(clus)}(X, \, A) &=  \mathcal{L}_{\text{ELBO}}^{(pre)}(X, \, A)  \\
&- 2 N \sum_{i=1}^{N} \mathbb{E}_{z_{i} \sim q(.|X,A)}\bigg[KL\Big(q(c_{i}|z_{i})||p(c_{i}|z_{i})\Big)\bigg].
\end{split}
$}
\label{eq:16}
\end{equation}

The obtained lower bound has an additional term compared to the pretraining lower bound. The new term is the clustering objective and comes with a mechanism against FR. In particular, it forces the clustering assignments to become closer to the target assignments for the nodes with a strong margin between the first and second high-confidence scores. Let $\Theta(t) =\left \{i \in \mathcal{V} | \;  (\lambda_{i}^{1}-\lambda_{i}^{2}) \geq  \alpha \right \}$ be the set of nodes with reliable clustering assignments in iteration $t$. This set gradually expands to include more reliable nodes based on the discriminative features learned in previous iterations. 

Existing evidence lower bounds do not support contrastive learning, which is the state-of-the-art strategy for unsupervised learning. To bridge this gap, we analytically establish a contrastive framework for variational auto-encoders by developing the term discarded by the ELBO-based methods. We start by specifying the negative and positive signals. We construct two attributed graphs $\mathcal{G}^{pos}$ and $\mathcal{G}^{neg}$ from the original graph $\mathcal{G}$. The first graph $\mathcal{G}^{pos}$ is defined by the adjacency matrix $A^{pos}=\left (a_{ij}^{pos} \right) \in \mathbb{R}^{N \times N}$ and the feature matrix $X^{pos}$. The second graph $\mathcal{G}^{neg}$ is defined by the adjacency matrix $A^{neg}=\left (a_{ij}^{neg} \right) \in \mathbb{R}^{N \times N}$ and the feature matrix $X^{neg}$.

\begin{figure*}
  \centerline{\includegraphics[width=0.85\textwidth]{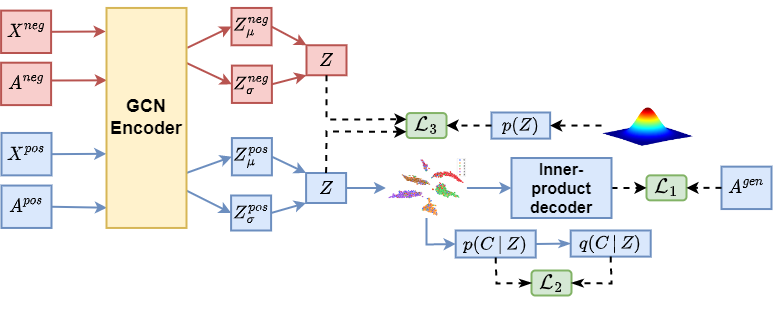}}
  \caption{Illustration of the proposed \textbf{C}ontrastive \textbf{V}ariational \textbf{G}raph \textbf{A}uto-\textbf{E}ncoder (CVGAE).}
  \label{fig:model_CVGAE}
  %\vspace{-2.5mm}
\end{figure*}

We set $X^{pos}=X$ and build a clustering-oriented graph structure $A^{pos}$ by gradually transforming the original graph structure. The adopted scheme consists of adding and removing edges based on the clustering assignments of reliable nodes. Specifically, we drop edges connecting nodes from different clusters. In addition, we add missing edges between the centroid node of each cluster and the remaining nodes of the same cluster. We identify the centroid node of each cluster using the node index of the nearest neighbor to the latent centroid. It is important to note that we only consider nodes in the set $\Theta$ for both operations (i.e., adding and removing edges). The function $\Upsilon$ to construct $A^{pos}$ is provided in Appendix D. For the second graph, we set $X^{neg}=X$ and $A^{neg}=A$ to contrast the clustering-oriented and original graphs. 

In addition to the input graphs $\mathcal{G}^{pos}$ and $\mathcal{G}^{neg}$, we construct a third graph $\mathcal{G}^{gen}$ as a self-supervisory signal for the decoding process. This graph is defined by the adjacency matrix $A^{gen}=\left (a_{ij}^{gen} \right) \in \mathbb{R}^{N \times N}$. Unlike the pretraining phase, the decoder does not reconstruct the original graph. Instead of that, our model is trained to construct an enhanced graph structure to account for the clustering task without destroying edges. Specifically, the scheme adopted to construct $A^{gen}$ consists of gradually adding edges, similar to the operation adopted to construct $A^{pos}$. The added edges increase the within-cluster similarities, which in turn reduce the effect of FD caused by the clustering-reconstruction trade-off. However, we refrain from dropping edges to preserve the between-cluster similarity, which is important for an effective generative process. The function $\Psi$ to construct $A^{gen}$ is summarized in Appendix E.

After specifying the encoding and decoding models along with the input and self-supervisory signals, we derive a contrastive variational lower bound $\mathcal{L}_{\text{CVGAE}}$ of the generated graph log-likelihood according to Theorem 1 (proof provided in Appendix C).

\newtheorem*{T1}{Theorem~1} 
\label{theorem_1}
\begin{T1} 
Given the design choices for the generative and inference models, we have the following.

\begin{equation}
%\resizebox{0.89\linewidth}{!}{$
 \mathcal{L}_{\text{ELBO}}^{(clus)}(X^{pos}, \: A^{pos}) \leqslant \mathcal{L}_{\text{CVGAE}} \leqslant  \text{log}\big(p(A^{gen})\big),
%$}
\label{eq:17}
\end{equation}

\begin{equation}
\resizebox{0.89\linewidth}{!}{$
\begin{split}
\mathcal{L}_{\text{CVGAE}} &= \mathcal{L}_{\text{ELBO}}^{(clus)}(X^{pos}, \: A^{pos}) \\ 
&+  2 \; N \; \sum_{i=1}^{N} KL\Big(q(z_{i} | X^{pos}, A^{pos}) \: \big\|  \: q(z_{i} | X^{neg}, A^{neg}) \Big).
\end{split}
$}
\label{eq:18}
\end{equation}
\end{T1}

From Theorem 1, we can see that the proposed objective function $\mathcal{L}_{\text{CVGAE}}$ is a tighter lower bound than the corresponding evidence lower bound. Furthermore, $\mathcal{L}_{\text{CVGAE}}$ establishes a contrastive learning framework due to the new term $KL(q(z_{i} | X^{pos}, A^{pos}) \: \|  \: q(z_{i} | X^{neg}, A^{neg}))$, which captures the difference between the clustering-oriented and original graphs in the latent space. Furthermore, the new term is a lower bound of the KL divergence between the true and variational posteriors. While maximizing the ELBO aims to make the variational posterior equal to the true posterior, the new term absorbs the difference between the inference and the generative models. Hence, the decoding process can restore the non-discriminative information destroyed by the encoding process.

The PC problem is associated with a significant number of hidden units becoming unable to capture any useful information from the input signal. It is caused by the regularization term. More precisely, the hidden units, which do not participate in optimizing the pseudo-supervised or self-supervised objectives of $\mathcal{L}_{\text{ELBO}}^{(clus)}$, are limited to follow a standard multivariate Gaussian distribution (considered inactive units). 

%The distribution $q(z_{i} | X^{neg}, A^{neg})$ preserves the encoding style of the pretraining phase. Since the pretraining phase does not have any mechanism to avoid PC, we can assume that the learnt distribution $q(z_{i} | X^{neg}, A^{neg}))$ is not sufficiently representative of the input signal. Therefore, we can see that contrasting $q(z_{i} | X^{pos}, A^{pos}))$ with $q(z_{i} | X^{neg}, A^{neg}))$ mitigates the PC problem. We provide strong empirical evidence of this aspect in the Experiments section.
 
The distribution $q(z_{i} | X^{neg}, A^{neg})$ preserves the encoding style of the pretraining phase. Regularization of pretraining induces $q(z_{i} | X, A)$ to follow a standard Gaussian distribution without any mechanism to avoid PC. Therefore, we can see that contrasting $q(z_{i} | X^{pos}, A^{pos}))$ with $q(z_{i} | X^{neg}, A^{neg}))$ can mitigate the PC problem. We provide empirical evidence of this aspect in the Experiments section.

%we know that the regularization term induces $q_{w}(z_{i} | X, A)$ to follow a standard Gaussian distribution during the pretraining phase. Hence, pushing $q_{u}(z_{i} | X, A^{clus})$ away from $q_{w}(z_{i} | X, A)$ increases the resistance to the PC problem.  
 
We can divide our lower bound $\mathcal{L}_{\text{CVGAE}}$ into three different components ($\mathcal{L}_{1}$, $\mathcal{L}_{2}$, and $\mathcal{L}_{3}$) according to the contributions of this work. The obtained decomposition is described in Eq. (\ref{eq:19}). The term $\mathcal{L}_{1}$ has a mechanism to alleviate FD. The term $\mathcal{L}_{2}$ has a mechanism to mitigate FR. The term $\mathcal{L}_{3}$ has a mechanism to mitigate PC. We illustrate the framework of the clustering phase in Figure \ref{fig:model_CVGAE}. 

\begin{equation}
\resizebox{0.89\linewidth}{!}{$
\begin{split}
\mathcal{L}_{\text{CVGAE}} &= 
\underbrace{\sum_{i,j=1}^{N} \mathbb{E}_{\substack{z_{i}, \, z_{j} \sim q(. \, | \, X^{pos}, \, A^{pos})}} \bigg[ \text{log}\big(p(a_{ij}^{gen} \, | \, z_{i}, \, z_{j})\big) \bigg]}_{\textbf{First term $\mathcal{L}_{1}$: construction of $A^{gen}$ + mechanism against FD}} \\ 
& \underbrace{- 2 \, N \, \sum_{i=1}^{N} \mathbb{E}_{z_{i} \sim q(.|X^{pos}, A^{pos})}\bigg[KL\Big(q(c_{i}|z_{i})||p(c_{i}|z_{i})\Big)\bigg]}_{\textbf{Second term $\mathcal{L}_{2}$: clustering objective + mechanism against FR}} \\ 
& \underbrace{- 2 \, N \, \sum_{i=1}^{N} \mathbb{E}_{z_{i} \sim q(.|X^{pos}, A^{pos})}\bigg[\text{log}\Big(\frac{p(z_{i})}{q(z_{i} | X^{neg}, A^{neg})}\Big)\bigg]}_{\textbf{Third term $\mathcal{L}_{3}$: contrastive objective + mechanism against PC}}.
\end{split}
$}
\label{eq:19}
\end{equation}

We approximate the first term $\mathcal{L}_{1}$ based on MCMC (Markov Chain Monte Carlo) and the reparameterization trick to ensure straightforward gradient-based optimization. Given $L$ is the number of Monte Carlo samples, the time complexity for computing this approximation is $\mathcal{O}(dL^{2}N^{2})$. The expression obtained for $\mathcal{L}_{1}$ is provided in Eq. (\ref{eq:20}).

\begin{equation}
\resizebox{0.89\linewidth}{!}{$
\begin{split}
\mathcal{L}_{1} & =\sum_{i,j=1}^{N} \mathbb{E}_{\substack{z_{i}, \, z_{j} \sim q(. \, | \, X^{pos}, \, A^{pos})}}, \\
&\simeq \: \frac{1}{L^{2}} \sum_{l_{1}, l_{2}=1}^{L} \sum_{i,j=1}^{N} a_{ij}^{gen} \: \text{log}\Big(\text{Sigmoid}\Big((z_{i}^{(l_1)})^{T}z_{j}^{(l_2)}\Big)\Big) \\
&+ \: \frac{1}{L^{2}} \sum_{l_{1}, l_{2}=1}^{L}  \sum_{i,j=1}^{N} (1 - a_{ij}^{gen}) \: \text{log}\Big(1-  \text{Sigmoid}\Big((z_{i}^{(l_1)})^{T}z_{j}^{(l_2)}\Big)\Big).\\
\end{split}
$}
\label{eq:20}
\end{equation}

The second term $\mathcal{L}_{2}$ is also approximated on the basis of MCMC and the reparameterization trick. The time complexity to compute this approximation is $\mathcal{O}(LKN)$ and the expression obtained is as follows: 

\begin{equation}
\resizebox{0.89\linewidth}{!}{$
\begin{split}
\mathcal{L}_{2} &= -2 \: N \: \sum_{i=1}^{N} \mathbb{E}_{\substack{z_{i} \sim q(z_{i}|X^{pos},A^{pos})}}\bigg[\sum_{c_{i}} q(c_{i}|z_{i}) \: \text{log}\Big(\frac{q(c_{i}|z_{i})}{p(c_{i}|z_{i})}\Big)\bigg], \\
&\simeq -2 \: \frac{N}{L} \sum_{l=1}^{L}  \sum_{i=1}^{N} \sum_{c_{i}} q(c_{i}|z_{i}^{(l)}) \: \text{log}\Big(\frac{q(c_{i}|z_{i}^{(l)})}{p(c_{i}|z_{i}^{(l)})}\Big). \\
\end{split}
$}
\label{eq:21}
\end{equation}

The three distributions $p(z_{i})$, $q(z_{i} | X^{pos}, A^{pos})$, and $q(z_{i} | X^{neg}, A^{neg})$ are all multivariate Gaussian distributions. Hence, we can compute a closed-form expression for $\mathcal{L}_{3}$. The time complexity for computing $\mathcal{L}_{3}$ is $\mathcal{O}(dN)$. Let $\mu_{z_{i}}^{pos}$ and $\sigma_{z_{i}}^{neg}$ be the mean and standard deviation, respectively, of $q(z_{i} | X^{pos}, A^{pos})$. Let $\mu_{z_{i}}^{neg}$ and $\sigma_{z_{i}}^{neg}$ be the mean and standard deviation, respectively, of $q(z_{i} | X^{neg}, A^{neg})$. Thus, the expression obtained for $\mathcal{L}_{3}$ is as follows: 

\begin{equation}
\resizebox{0.89\linewidth}{!}{$
\begin{split}
\mathcal{L}_{3} &= - 2 \, N \, \sum_{i=1}^{N} \mathbb{E}_{z_{i} \sim q(.|X^{pos}, A^{pos})}\bigg[\text{log}\big(\frac{p(z_{i})}{q(z_{i} | X^{neg}, A^{neg})}\big)\bigg], \\
&= \, N \, \sum_{i=1}^{N} \sum_{j=1}^{d} \bigg[ \text{log}( (\sigma_{z_{i}}^{neg}[j])^{2}) - (\mu_{z_{i}}^{pos}[j])^{2} - (\sigma_{z_{i}}^{pos}[j])^{2} \\
&+ \Big(\frac{\sigma_{z_{i}}^{pos}[j]}{\sigma_{z_{i}}^{neg}[j]}\Big)^{2} + \Big(\frac{\mu_{z_{i}}^{pos}[j]-\mu_{z_{i}}^{neg}[j]}{\sigma_{z_{i}}^{neg}[j]}\Big)^{2} \bigg]. \\
\end{split}
$}
\label{eq:22}
\end{equation}

\subsection{Algorithm}
We pretrain our model for $T_{1}$ iterations to maximize the lower bound in Eq.(\ref{eq:9}). The training weights $W$ are updated using Adam \cite{paper97} and backpropagation. For the clustering phase, we maximize the lower bound in Eq. (\ref{eq:19}). We run the optimization process until the number of reliable points represented by $\Theta$ constitutes $80\%$ of the entire set of nodes. We compute $\Theta$, $A^{pos}$ and $A^{gen}$ every $M$ iteration to avoid instability. The training weights $W$ and set of centers $\left \{ \Omega_{j} \right \}$ are updated using Adam and backpropagation. The algorithm is described in Algorithm \ref{algo_1}. 

%\begin{algorithm}
%\caption{Constructing $A^{gen}$}
%\label{algo_1}
%\begin{algorithmic}[1]
%\State {\bfseries Input:} Input adjacency matrix: $A$, Clustering assignment: $(p_{ij})$, Set of reliable nodes: $\Theta(t)$.
%\State {\bfseries Output:} Clustering-oriented graph structure: $A^{gen}$.
%\State $\Pi \leftarrow  \left [\text{1-NN}(\Omega_{j}, \Theta) \: | \:  j \in \left \{ 1,...,K \right \} \right ]$ \Comment{$\text{1-NN}(\Omega_{j}, \, \Theta)$ returns the index of the nearest neighbor of $\Omega_{j}$ among the set $\Theta(t)$.}  
%\State $A^{gen} \leftarrow  A$
%\For{$i$ in $\Theta(t)$}   
%    \State $k_{1} \leftarrow \text{arg}\max_{_{k}}(p_{ik})$
%    \State $l \leftarrow \Pi[k_{1}]$ 
%    \If{$\text{arg}\max_{_{k}}(p_{ik}) = \text{arg}\max_{_{k}}(p_{lk})$} 
%            \State $a_{il}^{gen} \leftarrow 1$
%    \EndIf
%\EndFor 
%\State \textbf{Return} $A^{gen}$
%\end{algorithmic}
%\end{algorithm}

\begin{algorithm}
\caption{CVGAE training}
\label{algo_1}
\begin{algorithmic}[1]
\State {\bfseries Input:} Features matrix: X, Adjacency matrix: A, Number of pretraining iterations: $T_{1}$, Number of clusters: $K$, Updating indicator: M, Confidence threshold: $\alpha$.

\For{$m=0$ {\bfseries to} $T_{1}$} 
\State Compute $\mathcal{L}_{\text{ELBO}}^{(pre)}$ according to Eq. (\ref{eq:9});
\State Update $W$ to maximize $\mathcal{L}_{\text{ELBO}}^{(pre)}$ using Adam optimizer; 
\EndFor
\State $m \gets 0$;
\State $\Theta \gets \varnothing$;
\State $A^{neg} \gets A$;
\State Compute $\left \{ \Omega_{j} \right \}_{j=1}^{K}$ using K-means;  
\While{$|\Theta| \leq 0.8 \; * \; |\mathcal{V}|$}
\State Compute $(p_{ij})$ according to Eq. (\ref{eq:11});
\If{$m \% M == 0$}
\State Compute $\lambda^{1}$ according to Eq. (\ref{eq:14}); 
\State Compute $\lambda^{2}$ according to Eq. (\ref{eq:15}); 
\State $\Theta \gets \left \{i \in \mathcal{V} | \;  (\lambda_{i}^{1}-\lambda_{i}^{2}) \geq  \alpha \right \}$;
\State $A^{pos} \gets \Upsilon(A, (p_{ij}), \Theta)$; 
\State $A^{gen} \gets \Psi(A, (p_{ij}), \Theta)$; 
\EndIf
\State Compute $\mathcal{L}_{\text{CVGAE}}$ according to Eq. (\ref{eq:19});
\State Update $W, \, \left \{ \Omega_{j} \right \}_{j=1}^{K}$ to maximize $\mathcal{L}_{\text{CVGAE}}$ using Adam; 
\State $m \gets m + 1$;
\EndWhile
\end{algorithmic}
\end{algorithm}

%We make three assumptions to facilitate the computational complexity analysis. First, we assume that all layers and the feature matrix have the same dimension $d$. Second, we assume that the second phase requires $T_{2}$ iterations. Third, we assume that the number of nodes $N$ $\gg$ K, $p$, and $d$, where $p$ denotes the number of encoding layers. The encoding process requires a time complexity $\mathcal{O}(p d |\mathcal{E}| + p d^{2} N)$. The computational complexity for computing $\mathcal{L}_{\text{ELBO}}^{(pre)}$ is $\mathcal{O}(p d |\mathcal{E}| + d N^{2})$. The computational complexity for computing $\mathcal{L}_{\text{BELBO}}^{(clus)}$ is also equal to $\mathcal{O}(2 p d |\mathcal{E}| + d N^{2})$. The time complexity of the additional terms that distinguish $\mathcal{L}_{\text{ELBO}}^{(pre)}$ and $\mathcal{L}_{\text{BELBO}}^{(clus)}$ are dominated by the computational complexity of the reconstruction term $\mathcal{O}(d N^{2})$. The computational complexity of Algorithm \ref{algo_1} is $\mathcal{O}(K|\mathcal{E}|N)$. The computational complexity of Algorithm 2 is $\mathcal{O}(2 p d \, (T_{1} + T_{2}) |\mathcal{E}| + K \left \lfloor T_{2} / M \right \rfloor |\mathcal{E}|N + d (T_{1} + T_{2}) N^{2})$.

To simplify the complexity estimation of our algorithm, we make three assumptions. We assume that the number of features for the input and all layers is constant $d$. We also assume that the number of iterations for the second phase is $T_{2}$. We denote the number of layers by $P$. Finally, we assume that $N$ $\gg$ $d$, $P$, and $L$. Calculating latent representations using the GCN architecture requires a computational complexity of $\mathcal{O}(d P |\mathcal{E}| + P d^{2} N)$. The time complexity to calculate the lower bound of the pretraining phase is equal to $\mathcal{O}(d P |\mathcal{E}| + d L^{2} N^{2})$. The time complexity to compute the lower bound of the clustering phase is also equal to $\mathcal{O}(2 d P |\mathcal{E}| + d L^{2} N^{2})$. The computational complexity to compute $A^{pos}$ and $A^{gen}$ is equal to $\mathcal{O}(\mathcal{E}KN)$ and $\mathcal{O}(KN)$, respectively. The time complexity of Algorithm 1 is equal to $\mathcal{O}(d P (T_{1} + 2 T_{2}) |\mathcal{E}| + K \left \lfloor T_{2} / M \right \rfloor |\mathcal{E}| N + d (T_{1} + T_{2}) L^{2}  N^{2})$.

\section{Experiments}

We conduct extensive experiments to validate the effectiveness, efficiency, and robustness of our model. We also perform specific experiments to validate the capacity of our model in mitigating FR, FD and PC. Additionally, we conducted an ablation study to understand the impact of each contribution in improving the quality of the clustering. Finally, we perform a qualitative study to support the quantitative findings.  

We compare our model with several state-of-the-art unsupervised models on the node clustering task. We divide the baselines into two groups. The first group represents the specific category of VGAE models, which is the most related group to our work. The methods selected from this category are $\mathcal{N}$-VGAE \cite{paper1}, $\mathcal{S}$-VGAE \cite{paper26}, $\mathcal{D}$-VGAE \cite{paper27}, GMM-VGAE \cite{paper13}, SI-VGAE \cite{paper28}, Graphite \cite{paper29}, and GDN-VAE \cite{paper30}. The second group represents the broad category of self-supervised GNN models. The selected methods from this broad category are GAE \cite{paper1}, ARGE/ARVGE \cite{paper2}, MGAE \cite{paper7}, AGC \cite{paper17}, DAEGC \cite{paper9}, SDCN \cite{paper21}, GALA \cite{paper10}, DGAE \cite{paper8}, AGE \cite{paper11}, DGI \cite{paper3}, GMI \cite{paper38}, GIC \cite{paper12}, MVGRL\cite{paper31}, COLES-S$^{2}$GC \cite{paper36}, GRACE \cite{paper34}, GCA \cite{paper35}, BGRL \cite{paper32}, and AFGRL \cite{paper33}. For the self-supervised approaches that do not adopt a clustering-oriented training, we run K-means on the latent codes at the end of the training process. Then, we leverage the clustering assignments to evaluate the model. All baselines are discussed in Section \ref{sec_2}. We employ the code provided by the authors for each baseline, and we set the hyperparameters according to the authors' recommendations, or we tune them in case there is no specification. 

\begin{table*}[t]
  \caption{Comparing the clustering results for different VGAE models. $-$ indicates that the code crashes. Best method in bold.}
  \begin{center}
  \begin{small}
  \scalebox{0.79}{\begin{tabular}{|P{1.2cm}|c|P{2cm}|P{2cm}|P{2cm}|P{2cm}|P{2cm}|P{2cm}|P{2cm}|P{2cm}|}
    \hline
    \textbf{Dataset} & \textbf{Metric} & \textbf{Graphite} & \textbf{GDN-VAE} & \textbf{$\mathcal{N}$-VGAE} & \textbf{$\mathcal{S}$-VGAE} & \textbf{$\mathcal{D}$-VGAE} & \textbf{GMM-VGAE} & \textbf{SI-VGAE} & \textbf{CVGAE} \\ \hline
                                   & \textbf{ACC} & 69.8 & 70.5 & 64.1 & 70.3 & 64.8 & 71.6 & 68.5 & \textbf{79.0} \\ \cline{2-10}
                                   & \textbf{NMI} & 51.8 & 51.1 & 46.2 & 51.6 & 50.2 & 53.3 & 51.0 & \textbf{59.9} \\ \cline{2-10}
    \multirow{2}{*}{\textbf{Cora}} & \textbf{ARI} & 46.4 & 46.3 & 39.3 & 46.8 & 42.2 & 48.1 & 45.9 & \textbf{58.8} \\ \cline{2-10}
                                   & \textbf{F1} & 68.0 & 66.0 & 63.6 & 68.3 & 65.3 & 64.1 & 66.4 & \textbf{77.2} \\ \cline{2-10}
                                   & \textbf{Pre} & 69.5 & 68.2 & 64.5 & 70.6 & 68.3 & 66.7 & 70.4 & \textbf{79.7} \\ \cline{2-10}
                                   & \textbf{Pur} & 70.5 & 70.5 & 64.7 & 70.6 & 69.4 & 71.8 & 69.3 & \textbf{79.0} \\  \hline  \hline
                                       & \textbf{ACC} & 52.3 & 62.2 & 51.2 & 62.2 & 55.4 & 67.6 & 54.8 & \textbf{71.8} \\ \cline{2-10}
                                       & \textbf{NMI} & 25.2 & 32.8 & 28.5 & 32.8 & 28.8 & 40.9 & 28.8 & \textbf{45.4} \\  \cline{2-10}
    \multirow{2}{*}{\textbf{Citeseer}} & \textbf{ARI} & 23.6 & 33.8 & 22.1 & 33.8 & 26.4 & 42.8 & 27.2 & \textbf{47.7} \\ \cline{2-10}
                                       & \textbf{F1} & 50.1 & 58.9 & 51.3 & 58.9 & 53.6 & 63.1 & 53.0 & \textbf{63.9} \\ \cline{2-10}
                                       & \textbf{Pre} & 52.1 & 59.1 & 59.4 & 59.1 & 56.7 & 63.4 & 55.1 & \textbf{66.6} \\ \cline{2-10}
                                       & \textbf{Pur} & 56.8 & 64.1 & 55.4 & 64.1 & 58.5 & 69.0 & 58.0 & \textbf{71.9} \\ \hline  \hline
                                     & \textbf{ACC} & 64.1 & 40.4 & 66.5 & 70.1 & 71.0 & 71.7 & 66.7 & \textbf{74.4} \\ \cline{2-10}
                                     & \textbf{NMI} & 23.8 & 01.2 & 25.4 & 29.1 & \textbf{38.7} & 29.5 & 26.0 & 34.8 \\ \cline{2-10}
    \multirow{2}{*}{\textbf{Pubmed}} & \textbf{ARI} & 23.0 & 00.1 & 25.7 & 32.2 & 29.5 & 33.5 & 25.9 & \textbf{38.2} \\ \cline{2-10}
                                     & \textbf{F1} & 64.6 & 30.3 & 66.3 & 68.9 & 71.1 & 70.7 & 66.7 & \textbf{74.1} \\ \cline{2-10}
                                     & \textbf{Pre} & 64.1 & 33.8 & 66.2 & 69.5 & \textbf{81.2} & 70.8 & 66.5 & 74.2 \\ \cline{2-10}
                                     & \textbf{Pur} & 64.1 & 43.3 & 66.5 & 70.1 & 71.0 & 71.7 & 66.7 & \textbf{74.4} \\ \hline  \hline
                                   & \textbf{ACC} & 65.4 & 57.6 & 58.9 & $-$ & 59.3 & 54.5 & 62.1 & \textbf{79.5} \\ \cline{2-10}
                                   & \textbf{NMI} & 32.6 & 27.2 & 21.2 & $-$ & 28.0 & 23.4 & 26.2 & \textbf{49.0} \\ \cline{2-10}
    \multirow{2}{*}{\textbf{DBLP}} & \textbf{ARI} & 30.8 & 15.9 & 21.6 & $-$ & 17.9 & 24.0 & 24.0 & \textbf{54.0} \\ \cline{2-10}
                                   & \textbf{F1} & 65.8 & 58.0 & 58.4 & $-$ & 59.5 & 48.3 & 61.9 & \textbf{79.0} \\ \cline{2-10}
                                   & \textbf{Pre} & 68.4 & 70.7 & 59.6 & $-$ & 71.6 & 52.3 & 65.9 & \textbf{79.0} \\ \cline{2-10}
                                   & \textbf{Pur} & 65.4 & 57.6 & 58.9 & $-$ & 59.3 & 54.5 & 62.1 & \textbf{79.5} \\ \hline \hline
                                  & \textbf{ACC} & 85.8 & 79.5 & 84.6 & 71.0 & 79.5 & 83.2 & 62.1 & \textbf{88.8} \\ \cline{2-10}
                                  & \textbf{NMI} & 56.6 & 49.3 & 54.3 & 35.4 & 48.9 & 53.9 & 32.5 & \textbf{62.3} \\ \cline{2-10}
    \multirow{2}{*}{\textbf{ACM}} & \textbf{ARI} & 62.1 & 46.1 & 59.7 & 33.9 & 47.3 & 56.2 & 29.2 & \textbf{69.4} \\ \cline{2-10}
                                  & \textbf{F1} & 85.9 & 79.8 & 84.7 & 70.7 & 79.5 & 82.9 & 61.5 & \textbf{88.8} \\ \cline{2-10}
                                  & \textbf{Pre} & 86.7 & 85.0 & 84.8 & 74.6 & 83.0 & 84.9 & 63.2 & \textbf{89.0} \\ \cline{2-10}
                                  & \textbf{Pur} & 85.8 & 79.5 & 84.6 & 71.0 & 79.5 & 83.2 & 62.1 & \textbf{88.8} \\ \hline \hline
                                  & \textbf{ACC} & 42.7 & 45.1 & 45.0 & $-$ & 46.4 & 42.6 & 42.2 & \textbf{58.8} \\ \cline{2-10}
                                   & \textbf{NMI} & 41.4 & 42.4 & 39.9 & $-$ & 40.2 & 39.6 & 40.2 & \textbf{45.7} \\ \cline{2-10} 
    \multirow{2}{*}{\textbf{Wiki}} & \textbf{ARI} & 24.2 & 24.0 & 25.2 & $-$ & 24.5 & 23.5 & 22.2 & \textbf{34.7} \\ \cline{2-10}
                                   & \textbf{F1} & 38.4 & 39.0 & 38.4 & $-$ & 39.9 & 36.6 & 37.3 & \textbf{43.2} \\ \cline{2-10}
                                   & \textbf{Pre} & 45.6 & 45.5 & 41.7 & $-$ & 44.3 & 43.7 & 43.3 & \textbf{46.0} \\ \cline{2-10}
                                   & \textbf{Pur} & 53.3 & 53.8 & 53.7 & $-$ & 51.7 & 53.7 & 52.6 & \textbf{61.1} \\ \hline 
  \end{tabular}}  	
  \end{small}
  \end{center}
  \label{Table:comparison_vgae}
\end{table*}

We perform our evaluations on six datasets: Cora \cite{paper99}, Citeseer \cite{paper99}, Pubmed \cite{paper99}, Wiki \cite{paper100}, ACM \cite{paper21}, and DBLP \cite{paper21}. We provide the statistics of these datasets and list their detailed description in Appendix F. We evaluate the clustering effectiveness using standard metrics such as ACC (Accuracy), NMI (Normalized Mutual Information), ARI (Adjusted Rand Index), F1 (Macro-averaged F1 score), Pre (Precision) and Pur (Purity). The results obtained are reported in \% and higher values indicate better clustering quality. We use the metrics $\Lambda_{FR}$ and $\Lambda_{FD}$ \cite{paper39} to estimate the level of FR and FD. Higher values for $\Lambda_{FR}$ ($\Lambda_{FD}$, respectively) indicate less FR (FD, respectively). We compute the number of Active Units in the latent space to estimate the level of PC. We use the standard metric $AU$ \cite{paper98} to quantify if a unit is active. 

We specify four modified versions of our model to understand the impact of our contributions. The first model is denoted CVGAE(-FR). It rules out the FR mechanism from CVGAE by computing $q_{ij}$ using hard-clustering assignments instead of gradual update based on the confidence threshold $\alpha$. The second model is denoted CVGAE(-FD). It rules out the FD mechanism by reconstructing the original adjacency matrix $A$ instead of the clustering-oriented structure defined in Appendix E. The third model is denoted CVGAE(-PC). It rules out the PC mechanism from CVGAE by eliminating the term $KL(q(z_{i} | X^{pos}, A^{pos}) \: \|  \: q(z_{i} | X^{neg}, A^{neg}) )$ from the lower bound $\mathcal{L}_{\text{CVGAE}}$. The fourth model is denoted CVGAE(-CL). It rules out contrastive learning from CVGAE by setting $\mathcal{G}^{pos} = \mathcal{G}^{neg} = \mathcal{G}$.
 
We provide a complete description of $\Lambda_{FR}$, $\Lambda_{FD}$, and $AU$ in Appendix G to make this work self-contained. The architecture of our model along with the required hyperparameters are specified in Appendix H. All experiments are performed under the same software and hardware environments, which are described in Appendix I. Our code is on Github \footnote[2]{\url{https://github.com/nairouz/CVGAE_PR}}.

\textbf{Effectiveness:} In Table \ref{Table:comparison_vgae}, we evaluate the effectiveness of our approach against state-of-the-art VGAE models on six datasets (Cora, Citeseer, Pubmed, Wiki, ACM, and DBLP) in terms of six metrics (ACC, NMI, ARI, F1, Pre, and Pur). The considered VGAE models maximize different evidence lower bounds of the log-likelihood function. The first two models (Graphite and GDN-VAE) focus on improving the decoding flexibility and use the same inference model as $\mathcal{N}$-VGAE. The other methods do not exploit a sophisticated decoding style (vanilla reconstruction based on the inner product). Instead, they focused on refining the encoding capacity. First, we find that enhancing the decoding flexibility can bring about significant improvement to the clustering task, which is sometimes higher than the improvement brought about by some encoding refinement. This result suggests the importance of performing a reliable decoding process on the clustering task. Second, we observe that our approach outperforms the other VGAE models by \textit{significant margin} for all the datasets and metrics considered. Unlike the other VGAE models, CVGAE stands out by a tighter lower bound than the associated ELBO. Additionally, CVGAE does not perform vanilla reconstruction and it has the capacity to tackle FR, FD and PC.

In Table \ref{Table:comparison_ss}, we perform a broader comparison with state-of-the-art self-supervised methods. We form different groups according to the abstraction level of the training process. As we can see, the category consisting of combining proximity- and cluster-level training generally yields better results than the other strategies. Among this category, our approach outperforms all other methods. Combining proximity- and cluster-level training requires controlling the trade-off between FR and FD, which was neglected by previous methods. Unlike these methods, our approach gradually transforms the proximity-level task into a clustering-oriented task to alleviate FD.

\begin{table*}[t]
  \caption{Comparing the node clustering results for different graph self-supervised methods. Best method in bold.}
  %\vspace{-1\baselineskip}
  \begin{center}
  \begin{small}
  \scalebox{0.78}{\begin{tabular}{|P{3cm}|P{1.2cm}|P{1.2cm}|P{1.2cm}|P{1.2cm}|c|c|c|c|c|c|c|c|c|c|c|c|}
    \hline
    {\textbf{Method}} & \multicolumn{4}{c|}{\textbf{Abstraction level of the training process}} & \multicolumn{3}{c|}{\textbf{Cora}} & \multicolumn{3}{c|}{\textbf{Citeseer}} & \multicolumn{3}{c|}{\textbf{Pubmed}} & \multicolumn{3}{c|}{\textbf{DBLP}} \\
    \cline{2-17}
    {\textbf{}} & \textbf{Node} & \textbf{Proximity} & \textbf{Cluster} & \textbf{Graph} & \textbf{ACC} & \textbf{NMI} & \textbf{ARI} & \textbf{ACC} & \textbf{NMI} & \textbf{ARI} & \textbf{ACC} & \textbf{NMI} & \textbf{ARI} & \textbf{ACC} & \textbf{NMI} & \textbf{ARI} \\ \hline
    \textbf{GAE} & \xmark & \cmark & \xmark & \xmark & 61.3 & 44.4 & 38.1 & 48.2 & 22.7 & 19.2 & 63.2 & 24.9 & 24.6 & 61.2 & 30.8 & 22.0 \\ \hline
    \textbf{ARGE} & \xmark & \cmark & \xmark & \xmark & 64.0 & 44.9 & 35.2 & 57.3 & 35.0 & 34.1 & 68.1 & 27.6 & 29.1 & 64.8 & 29.4 & 28.0 \\ \hline
    \textbf{ARVGE} & \xmark & \cmark & \xmark & \xmark & 63.8 & 45.0 & 37.4 & 54.4 & 26.1 & 24.5 & 63.5 & 23.2 & 22.5 & 54.4 & 25.9 & 19.8 \\ \hline
    \textbf{GMI} & \xmark & \cmark & \xmark & \xmark & 63.4 & 50.3 & 38.8 & 63.8 & 38.1 & 37.5 & 67.1 & 26.2 & 26.8 & 73.2 & 43.2 & 42.5 \\ \hline
    \textbf{MGAE} & \xmark & \cmark & \xmark & \xmark & 68.1 & 48.9 & 43.6 & 66.9 & 41.6 & 42.5 & 59.3 & 28.2 & 24.8 & 42.1 & 15.8 & 07.6 \\ \hline
    \textbf{AGC} & \xmark & \cmark & \xmark & \xmark & 68.9 & 53.7 & 48.6 & 67.0 & 41.1 & 41.9 & 69.8 & 31.6 & 31.9 & 61.0 & 30.0 & 19.3 \\ \hline
    \textbf{COLES-S$^{2}$GC} & \xmark & \cmark & \xmark & \xmark & 69.4 & 54.5 & 48.0 & 70.0 & 45.2 & 46.5 & 65.3 & 31.8 & 28.9 & 44.6 & 14.9 & 10.4 \\ \hline
    \textbf{GALA} & \xmark & \cmark & \xmark & \xmark & 74.6 & 57.7 & 53.2 & 69.3 & 44.1 & 44.6 & 69.4 & 32.7 & 32.1 & 48.0 & 20.7 & 12.8 \\ \hline \hline
    
    \textbf{DGI} & \xmark & \xmark & \xmark & \cmark & 71.3 & 56.4 & 51.1 & 68.8 & 44.4 & 45.0 & 58.9 & 27.7 & 31.5 & 55.4 & 32.9 & 27.0 \\ \hline
    \textbf{MVGRL}  & \xmark & \xmark & \xmark &  \cmark & 73.2 & 56.2 & 51.9 & 68.7 & 43.7 & 44.3 & 67.0 & 31.6 & 29.4 & 42.7 & 15.4 & 08.2 \\ \hline \hline
    
    \textbf{GRACE} & \cmark & \xmark & \xmark & \xmark & 65.8 & 51.7 & 44.0 & 67.5 & 41.9 & 42.1 & 70.2 & 36.7 & 33.6 & 62.2 & 33.3 & 22.4 \\ \hline
    \textbf{GCA} & \cmark & \xmark & \xmark & \xmark & 69.1 & 51.7 & 48.4 & 68.0 & 43.2 & 43.3 & 70.7 & \textbf{37.1} & 34.1 & 66.1 & 34.9 & 27.4 \\ \hline
    \textbf{BGRL} & \cmark & \xmark & \xmark & \xmark & 73.8 & 54.7 & 51.1 & 65.6 & 38.4 & 38.7 & 58.7 & 24.9 & 23.1 & 45.7 & 12.1 & 10.1 \\ \hline \hline
    
    \textbf{GIC} & \xmark & \xmark & \cmark & \cmark & 72.5 & 53.7 & 50.8 & 69.6 & 45.3 & 46.5 & 67.3 & 31.9 & 29.1 & 71.1 & \textbf{52.6} & 48.6 \\ \hline \hline
    
    \textbf{DAEGC} & \xmark & \cmark & \cmark & \xmark & 70.4 & 52.8 & 49.6 & 67.2 & 39.7 & 41.0 & 67.1 & 26.6 & 27.8 & 62.1 & 32.5 & 21.0 \\ \hline 
    \textbf{SDCN} & \xmark & \cmark & \cmark & \xmark & 71.2 & 53.5 & 50.6 & 65.9 & 38.7 & 40.1 & 63.7 & 28.1 & 24.5 & 68.1 & 39.5 & 39.2 \\ \hline
    \textbf{R-DGAE} & \xmark & \cmark & \cmark & \xmark & 73.7 & 56.0 & 54.1 & 70.5 & 45.0 & 47.1 & 71.4 & 34.4 & 34.6 & 60.8 & 30.7 & 23.9 \\ \hline 
    \textbf{AFGRL} & \xmark & \cmark & \cmark & \xmark & 74.6 & 58.4 & 57.6 & 67.4 & 42.2 & 42.7 & 63.9 & 27.6 & 25.4 & 55.8 & 28.8 & 18.5 \\ \hline
    %\textbf{DBGAN \cite{paper37}} &  &  &  &  & 74.8 & 56.0 & 54.0 & 67.0 & 40.7 & 41.4 & 69.4 & 32.4 & 32.7 &  &  &  \\ \hline
    \textbf{AGE} & \xmark & \cmark & \cmark & \xmark & 76.1 & 59.7 & 54.5 & 70.1 & 44.3 & 45.4 & 70.9 & 30.8 & 32.9 & 50.8 & 22.8 & 15.1 \\ \hline
    
    \textbf{CVGAE} & \xmark & \cmark & \cmark & \xmark & \textbf{79.0} & \textbf{59.9} & \textbf{58.8} & \textbf{71.8} & \textbf{45.4} & \textbf{47.7} & \textbf{74.4} & 34.8 & \textbf{38.2} & \textbf{79.5} & 49.0 & \textbf{54.0} \\ \hline
  \end{tabular}}
  \end{small}
  \end{center}
  %\vskip -0.1in
  \label{Table:comparison_ss}
\end{table*}

\begin{figure*}[t]
  \begin{subfigure}[b]{0.33\textwidth}
    \includegraphics[width=\linewidth]{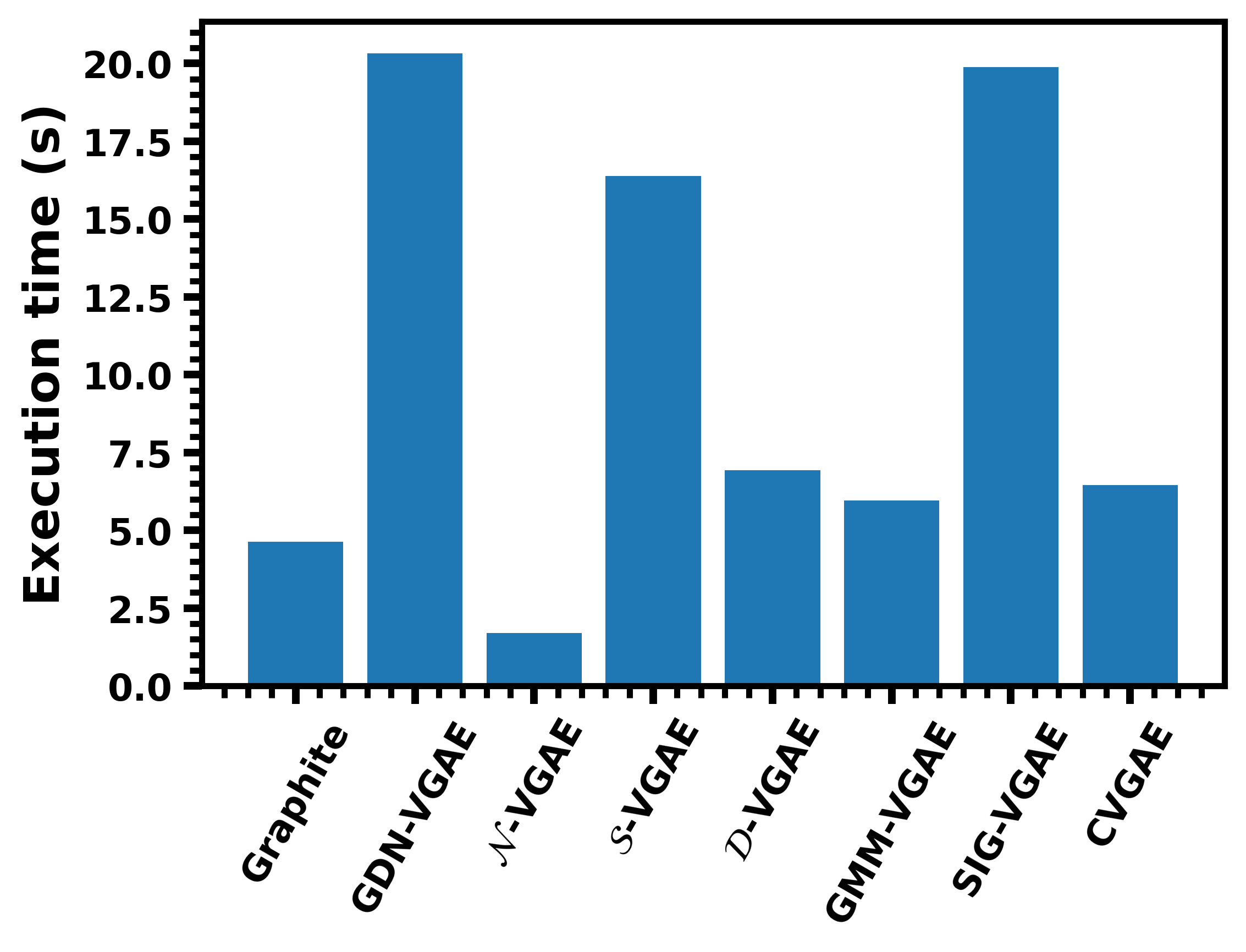}
    \caption{Cora}
  \end{subfigure}
  \begin{subfigure}[b]{0.33\textwidth}
     \includegraphics[width=\linewidth]{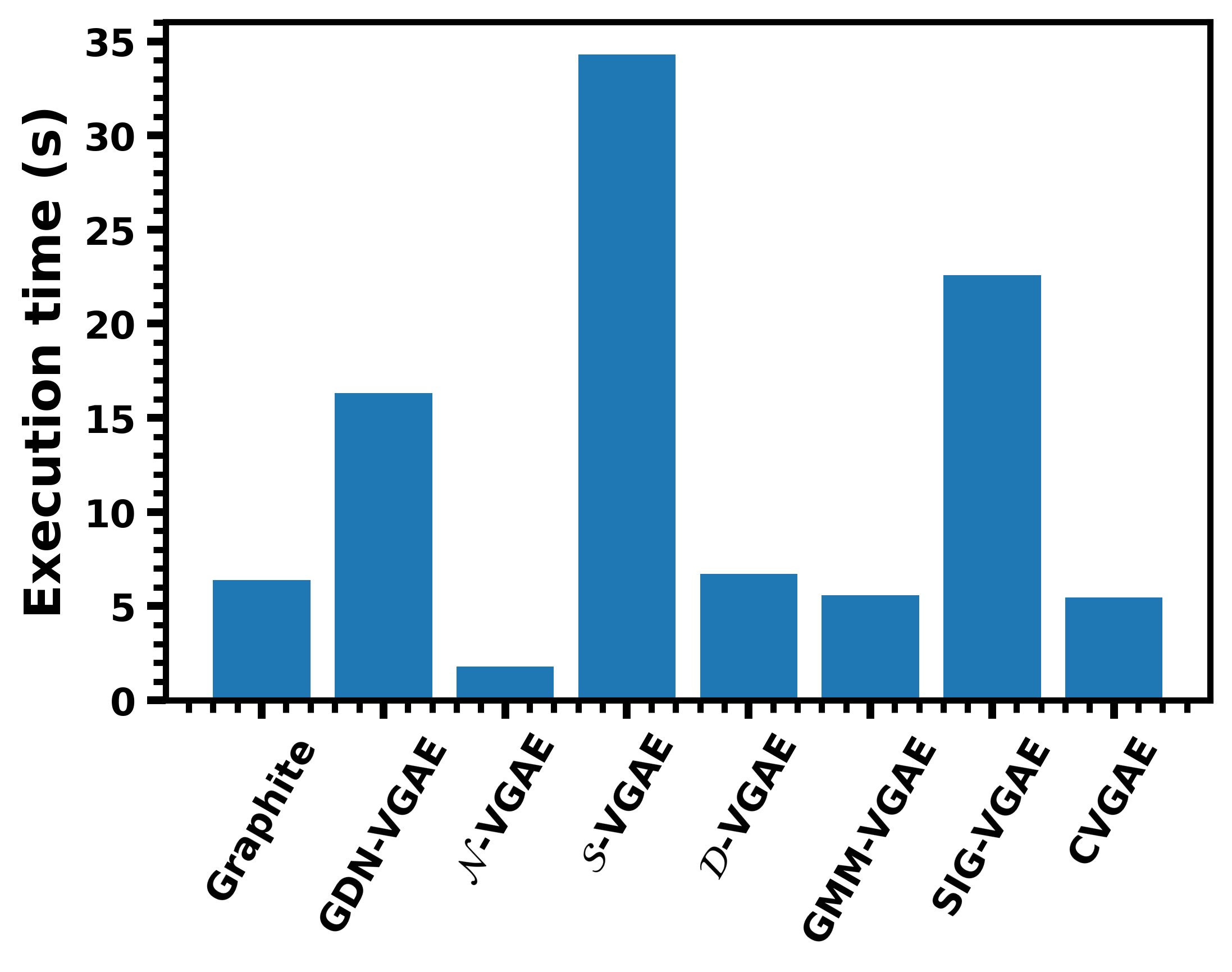}
     \caption{Citeseer}
  \end{subfigure}
  \begin{subfigure}[b]{0.33\textwidth}
    \includegraphics[width=\linewidth]{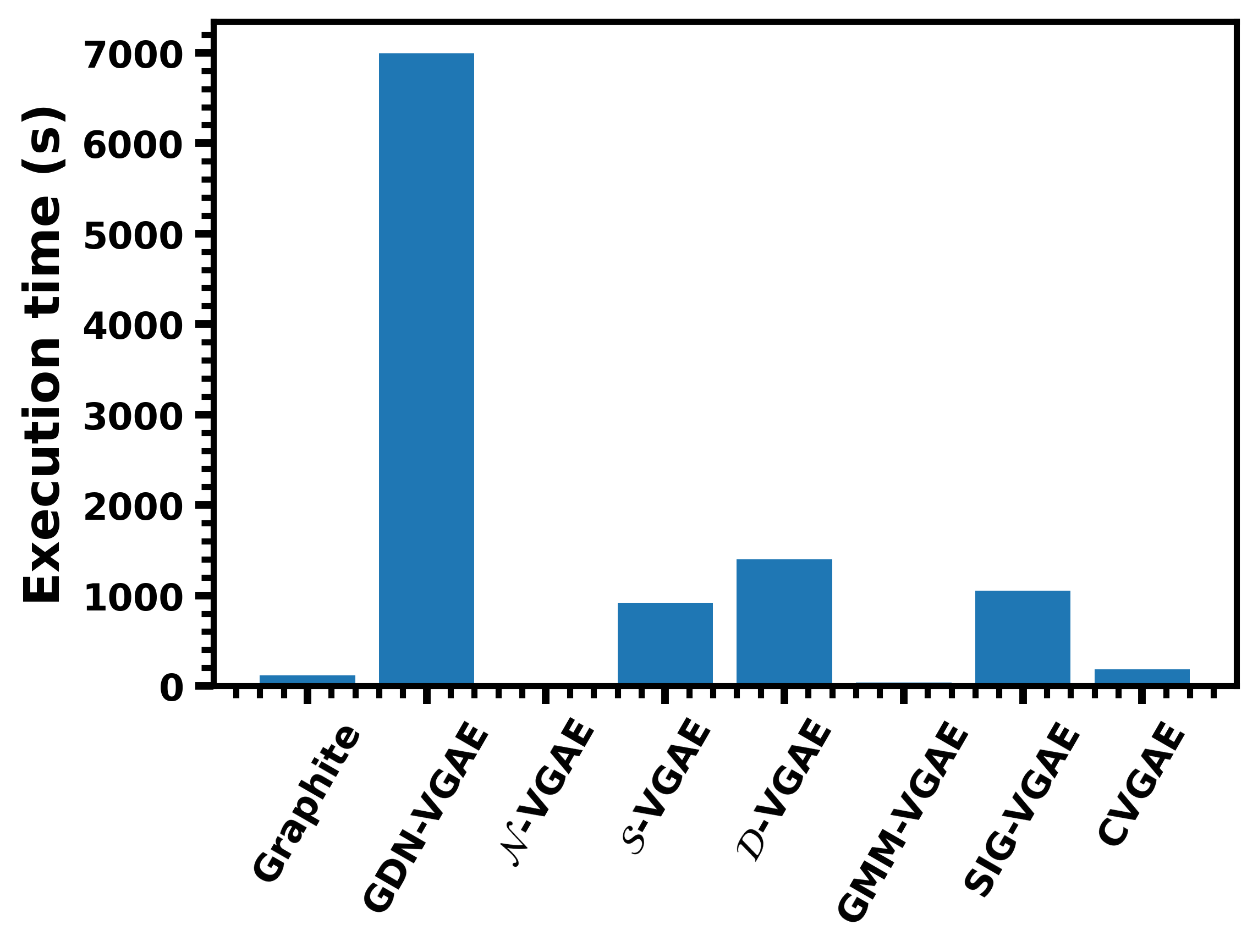}
    \caption{Pubmed}
  \end{subfigure}
  \caption{Comparing different VGAE models in terms of execution time.}
  \label{fig:efficiency_CVGAE}
\end{figure*}

\textbf{Efficiency:} We evaluate our method against its variational counterparts in terms of execution time on three datasets (Cora, Citeseer, and Pubmed). The results obtained are illustrated in Figure \ref{fig:efficiency_CVGAE}. As we can see, CVGAE is consistently more efficient than GDN-VAE, $\mathcal{S}$-VGAE, $\mathcal{D}$-VGAE, and SIG-VGAE. However, our model has a higher run-time than $\mathcal{N}$-VGAE and GMM-VGAE. To understand this aspect, we perform an ablation study. We compare the execution time of CVGAE and CVGAE(-CL-FD). CVGAE(-CL-FD) does not rely on contrastive learning and the FD mechanism. Consequently, CVGAE(-CL-FD) does not construct any additional graph.   %to track down the reason for this difference. 
We report the results in Table \ref{Table:efficiency_ablation}. We find that constructing the two graphs $\mathcal{G}^{pos}$ and $\mathcal{G}^{gen}$, which is crucial for contrastive learning and reducing FD, increases run-time significantly. Globally, our model has a reasonable training time in accordance with the complexity analysis. 

\begin{table}
  \caption{Execution time (in seconds) of $\mathcal{N}$-VGAE, GMM-VGAE, CVGAE(-CL-FD), CVGAE. }
  \begin{center}
  \begin{small}
  \begin{tabular}{|p{3cm}|c|c|c|}
    \hline
    Method & Cora & Citeseer & Pubmed \\ \hline
    \textbf{$\mathcal{N}$-VGAE \cite{paper1}} & 1.7 &  1.8 & 17.1 \\ \hline
    \textbf{GMM-VGAE \cite{paper13}} & 6.0 & 5.6 & 34.8 \\ \hline
    \textbf{CVGAE(-CL-FD)} & 3.9 & 4.3 & 35.9 \\  \hline
    \textbf{CVGAE} & 6.4 & 5.4 & 183.7 \\  \hline
  \end{tabular}
  \end{small}
  \end{center}
  \vskip -0.25in
  \label{Table:efficiency_ablation}
\end{table}

\begin{figure*}[t]
  \begin{subfigure}[b]{0.24\textwidth}
    \includegraphics[width=\linewidth]{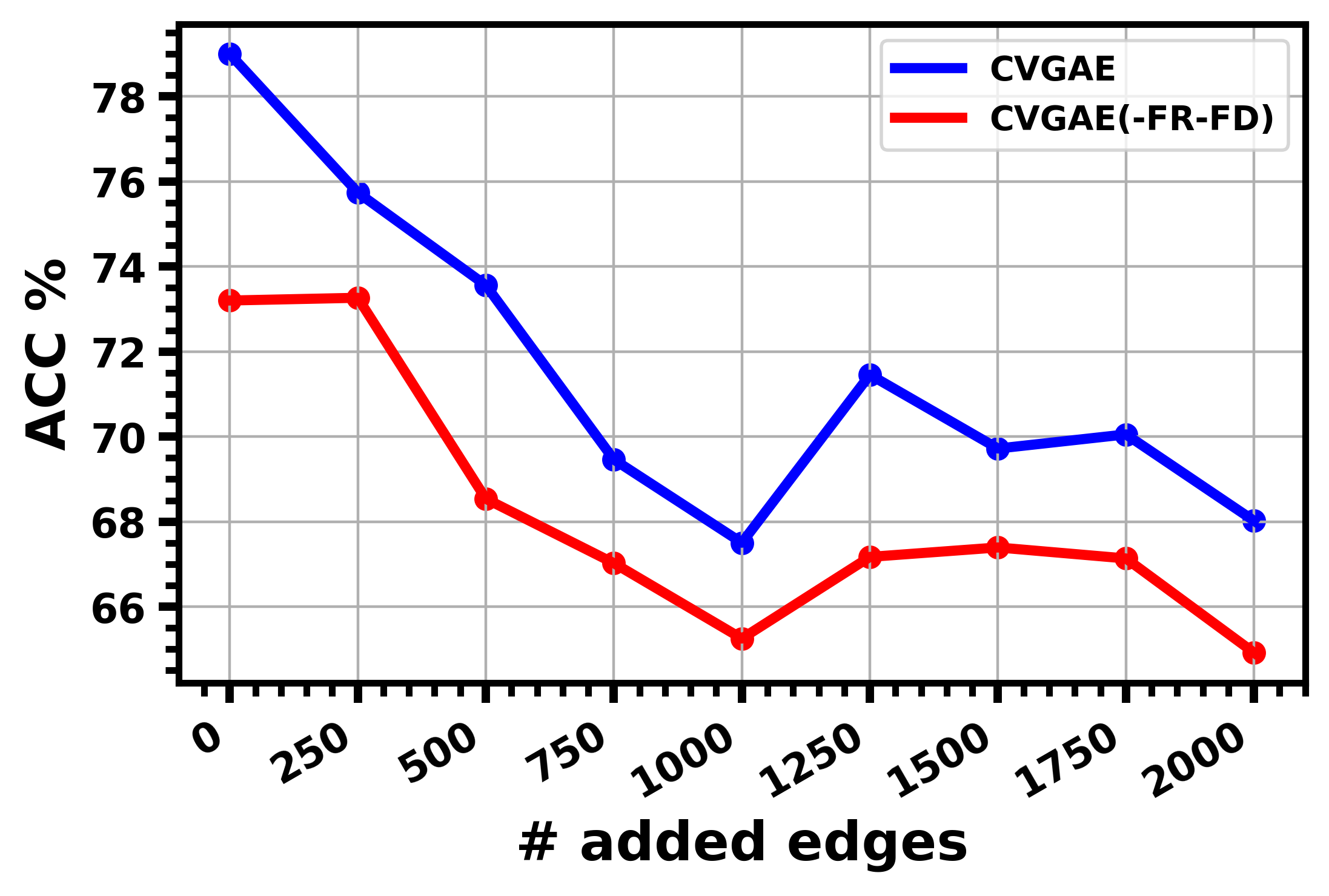}
    %\caption{ACC with noisy edges.}
  \end{subfigure}
  \begin{subfigure}[b]{0.24\textwidth}
     \includegraphics[width=\linewidth]{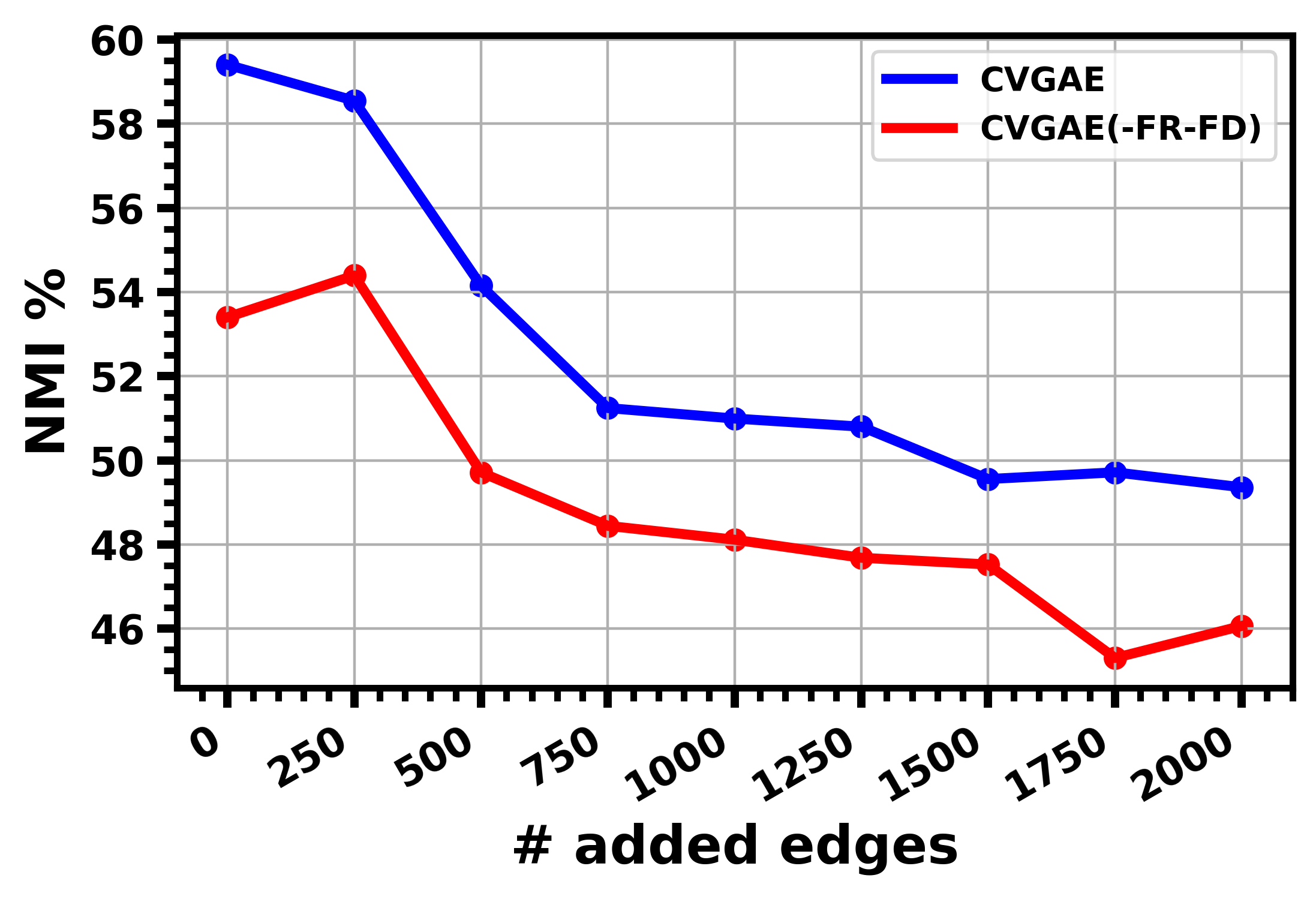}
     %\caption{ARI with noisy edges.}
  \end{subfigure}
  \begin{subfigure}[b]{0.24\textwidth}
     \includegraphics[width=\linewidth]{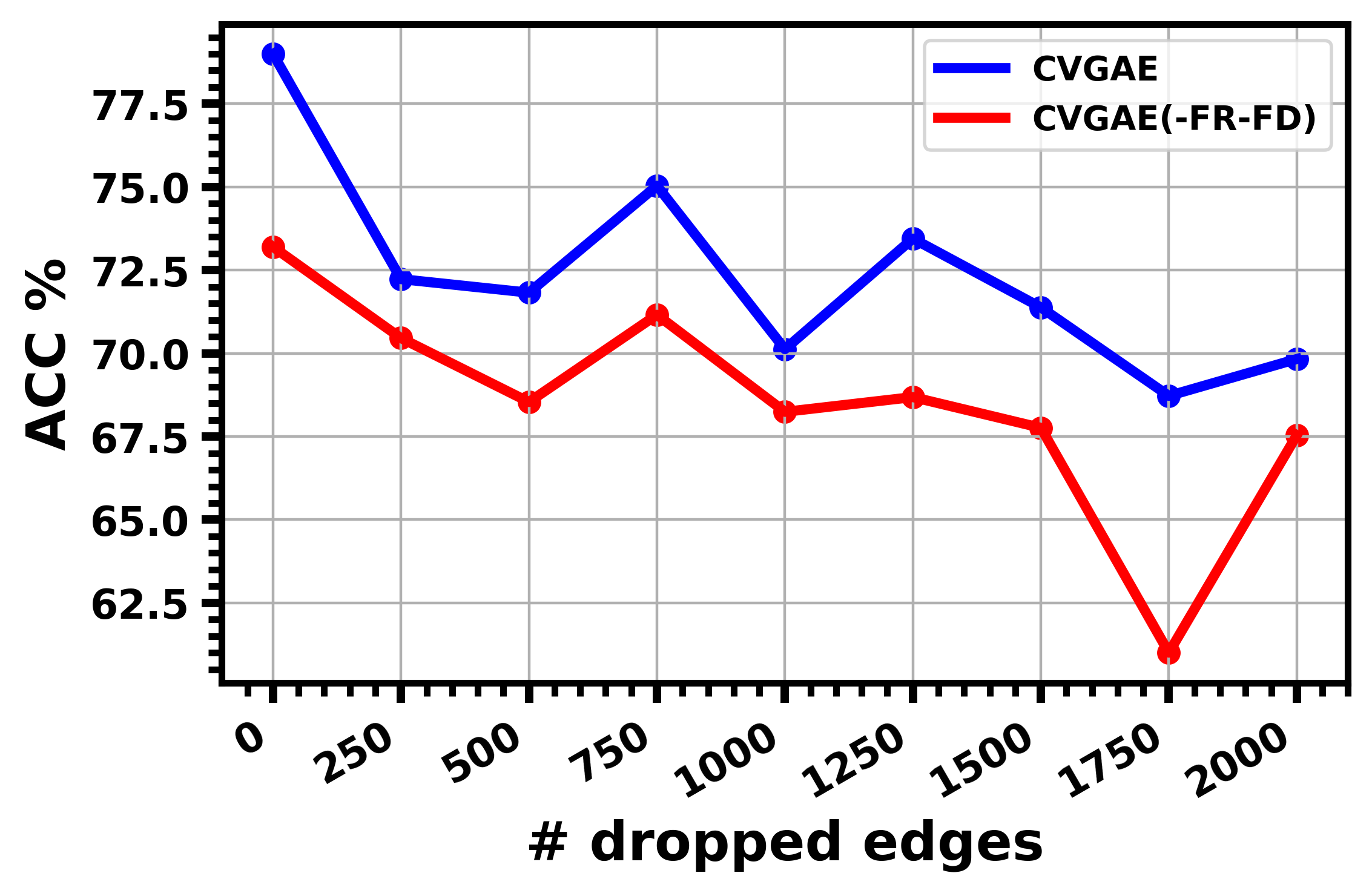}
     %\caption{ACC with noisy features.}
  \end{subfigure}
  \begin{subfigure}[b]{0.24\textwidth}
     \includegraphics[width=\linewidth]{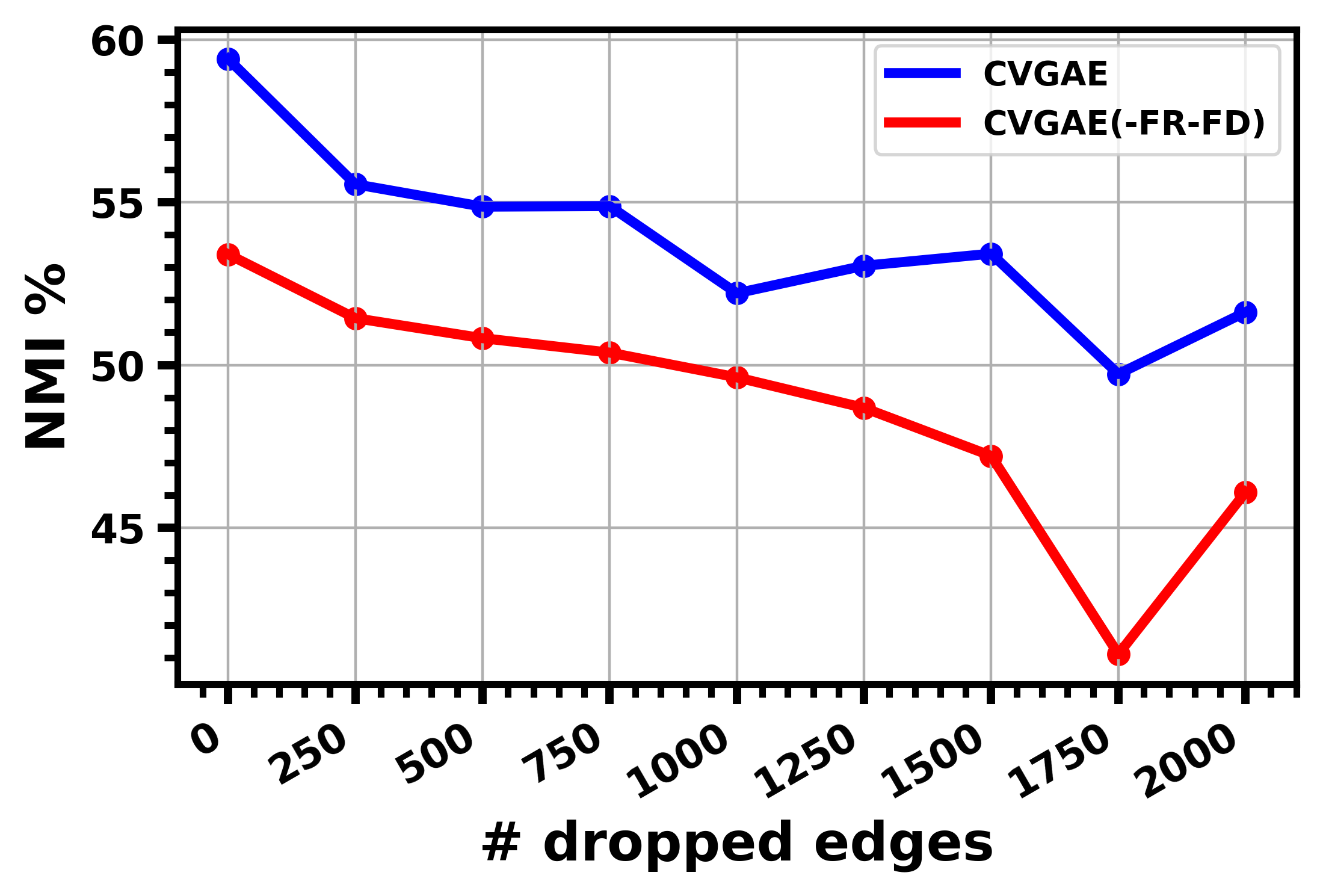}
     %\caption{ARI with noisy features.}
  \end{subfigure}
  \vskip 0.1in
  \caption{Performance of CVGAE and CVGAE(-FR-FD) on Cora in terms of ACC and NMI after adding and dropping edges.}
  \label{fig:robustness_edges_CVGAE}
\end{figure*}

\begin{figure*}[t]
  \begin{subfigure}[b]{0.24\textwidth}
    \includegraphics[width=\linewidth]{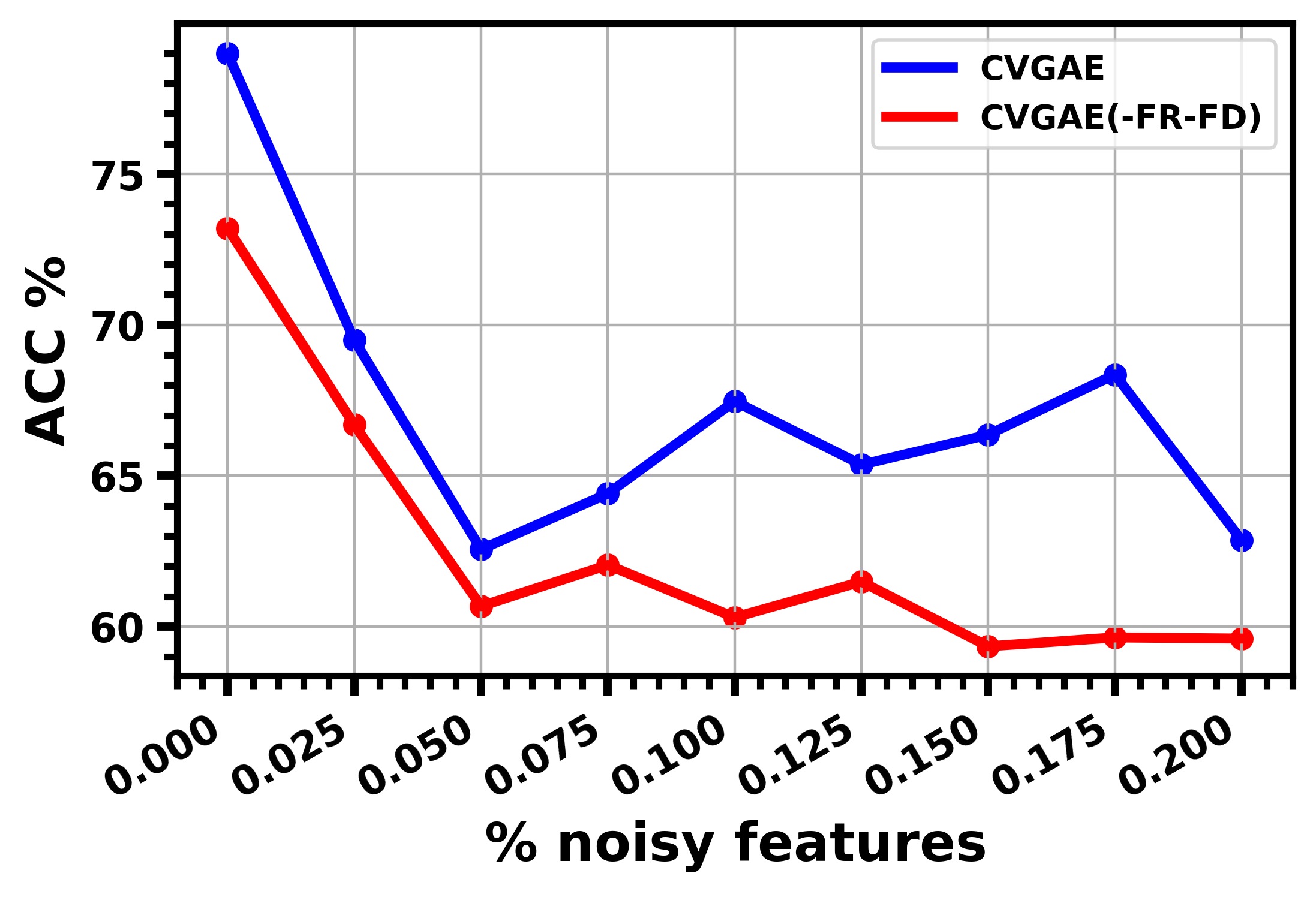}
    %\caption{ACC with noisy edges.}
  \end{subfigure}
  \begin{subfigure}[b]{0.24\textwidth}
     \includegraphics[width=\linewidth]{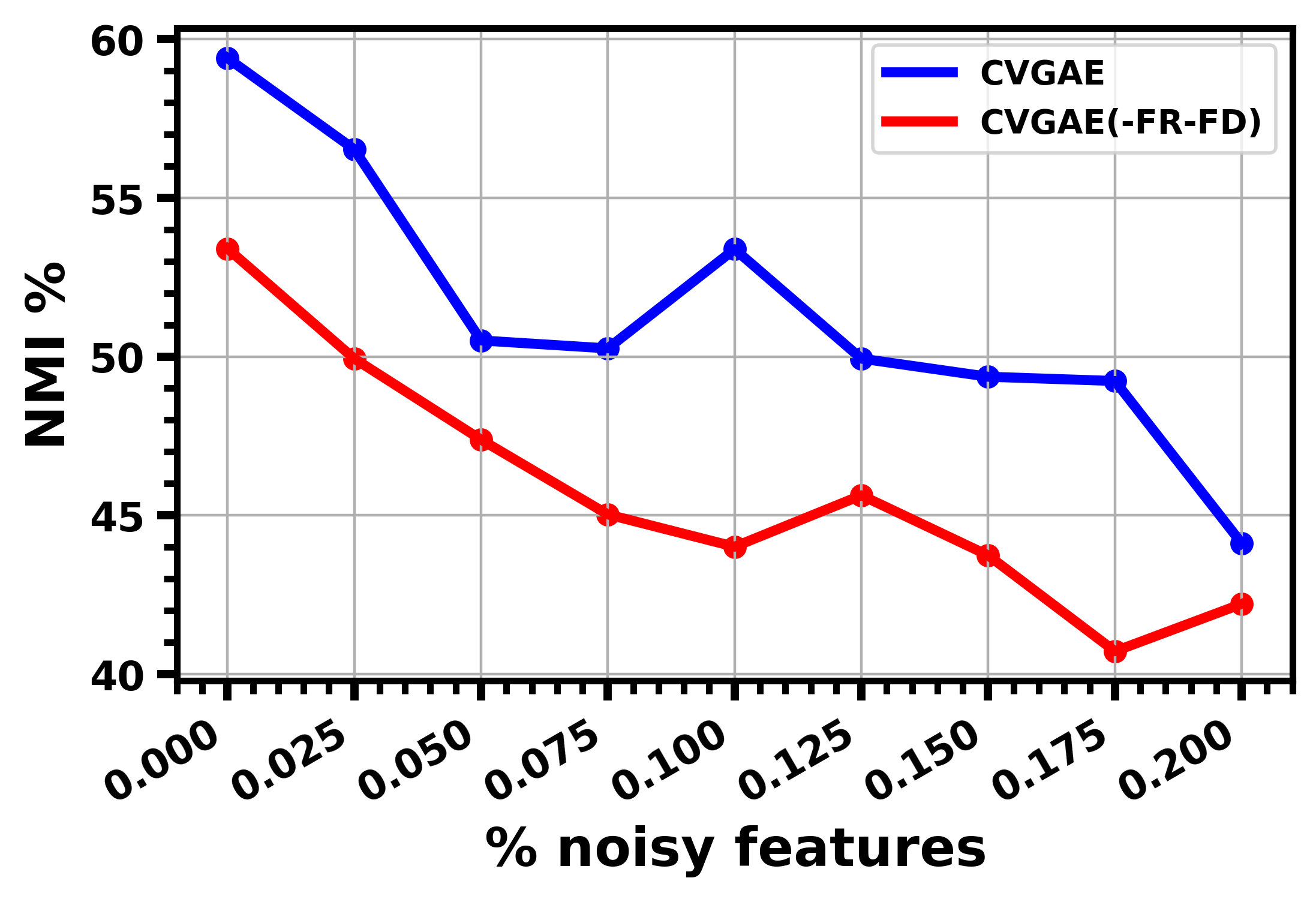}
     %\caption{ARI with noisy edges.}
  \end{subfigure}
  \begin{subfigure}[b]{0.24\textwidth}
     \includegraphics[width=\linewidth]{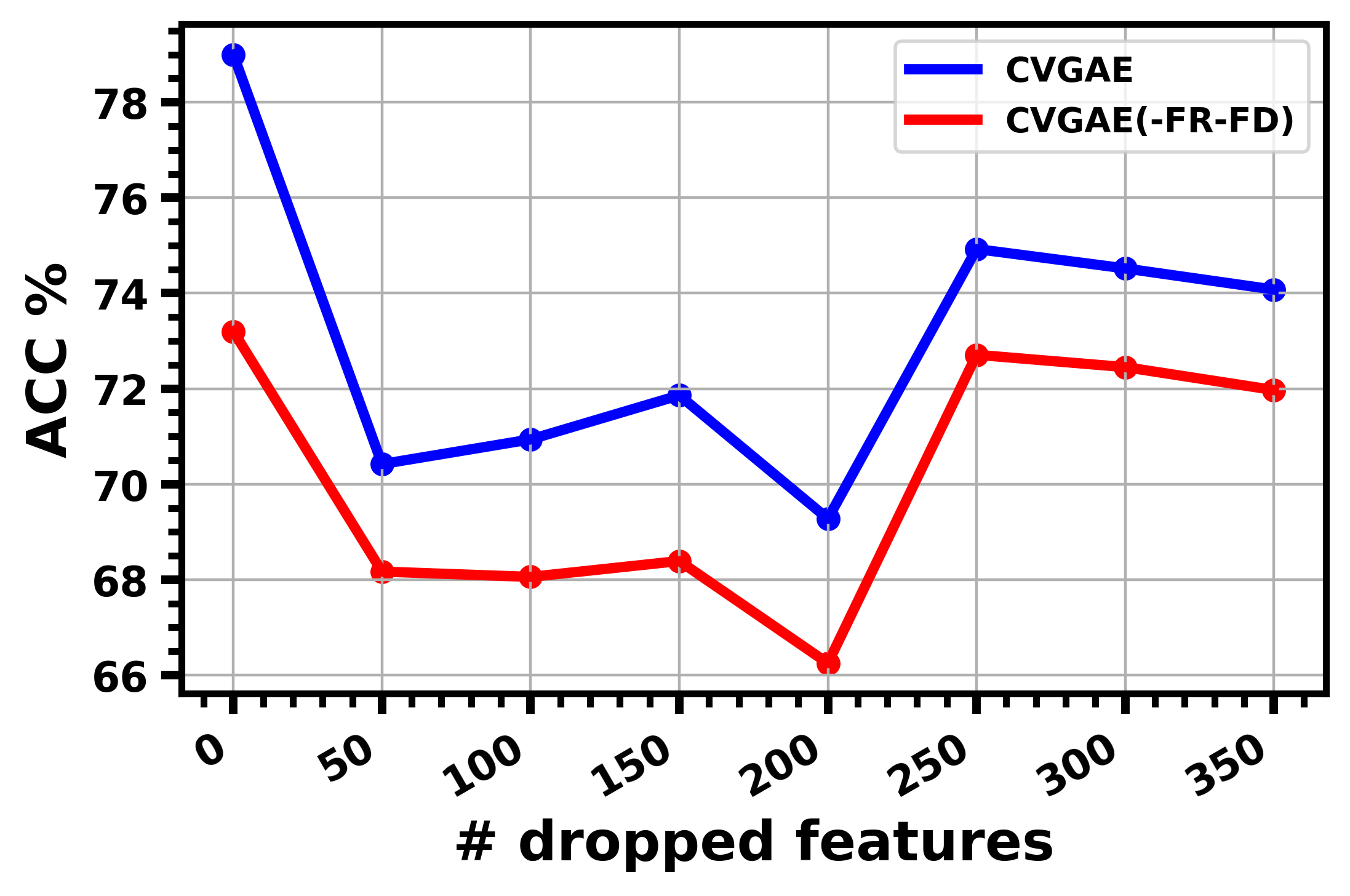}
     %\caption{ACC with noisy features.}
  \end{subfigure}
  \begin{subfigure}[b]{0.24\textwidth}
     \includegraphics[width=\linewidth]{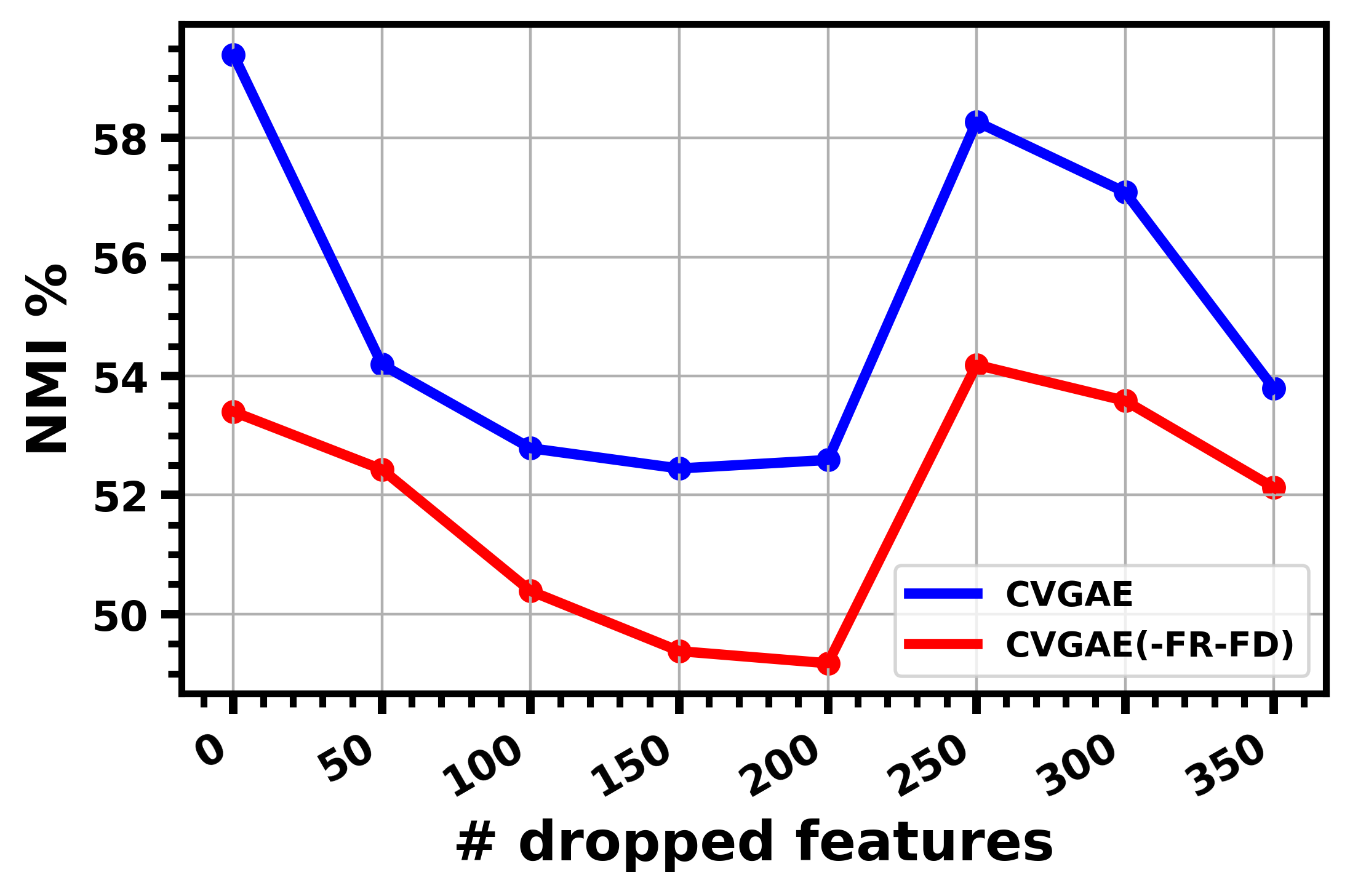}
     %\caption{ARI with noisy features.}
  \end{subfigure}
  \vskip 0.1in
  \caption{Performance of CVGAE and CVGAE(-FR-FD) on Cora in terms of ACC and NMI after adding Gaussian noise and dropping features.}
  \label{fig:robustness_features_CVGAE}
\end{figure*}

\textbf{Robustness:} We investigate the impact of alleviating FR and FD on the robustness of our model. We denote the model obtained after ablation of the FR and FD mechanisms by CVGAE(-FR-FD). In Figure \ref{fig:robustness_edges_CVGAE}, we illustrate the effect of randomly adding and dropping edges from the input structure on CVGAE and CVGAE(-FR-FD). In Figure \ref{fig:robustness_features_CVGAE}, we illustrate the effect of randomly dropping features and corrupting them with Gaussian noise on CVGAE and CVGAE(-FR-FD). The dropped and added edges are the same for the two models to ensure a fair comparison. Similarly, the dropped and corrupted features are also the same for both models. We observe that CVGAE consistently outperforms CVGAE(-FR-FD). Moreover, CVGAE is less affected by random modifications to the input features or graph structure. The FR and FD mechanisms allow the model to gradually accumulate and refine clustering-oriented information. Therefore, CVGAE has more capacity to filter the effect of random modifications to the input graph.

\begin{figure*}[t]
  \begin{subfigure}[b]{0.33\textwidth}
    \includegraphics[width=\linewidth]{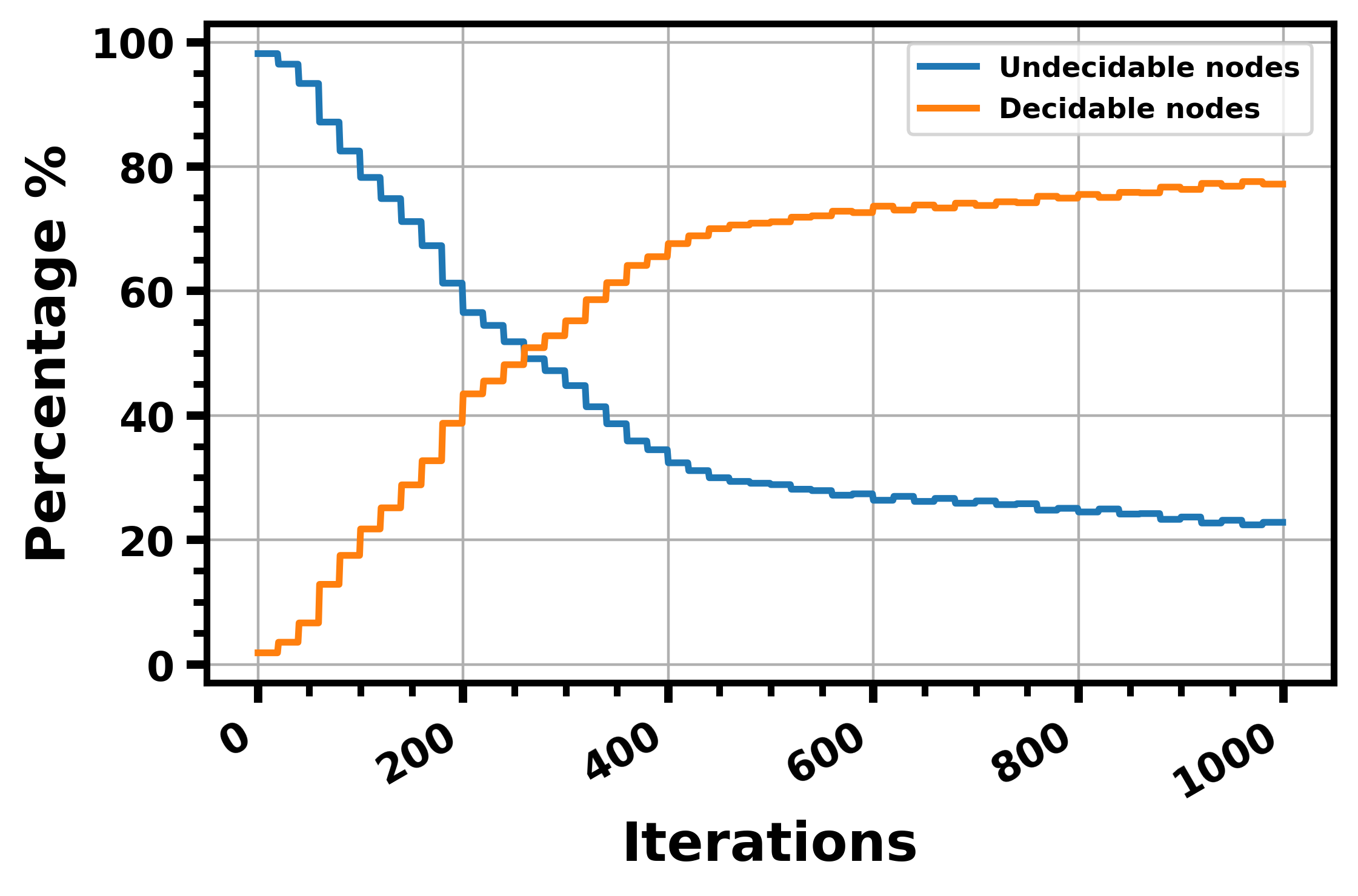}
    \caption{\% of decidable and undecidable nodes}
  \end{subfigure}
  \begin{subfigure}[b]{0.33\textwidth}
     \includegraphics[width=\linewidth]{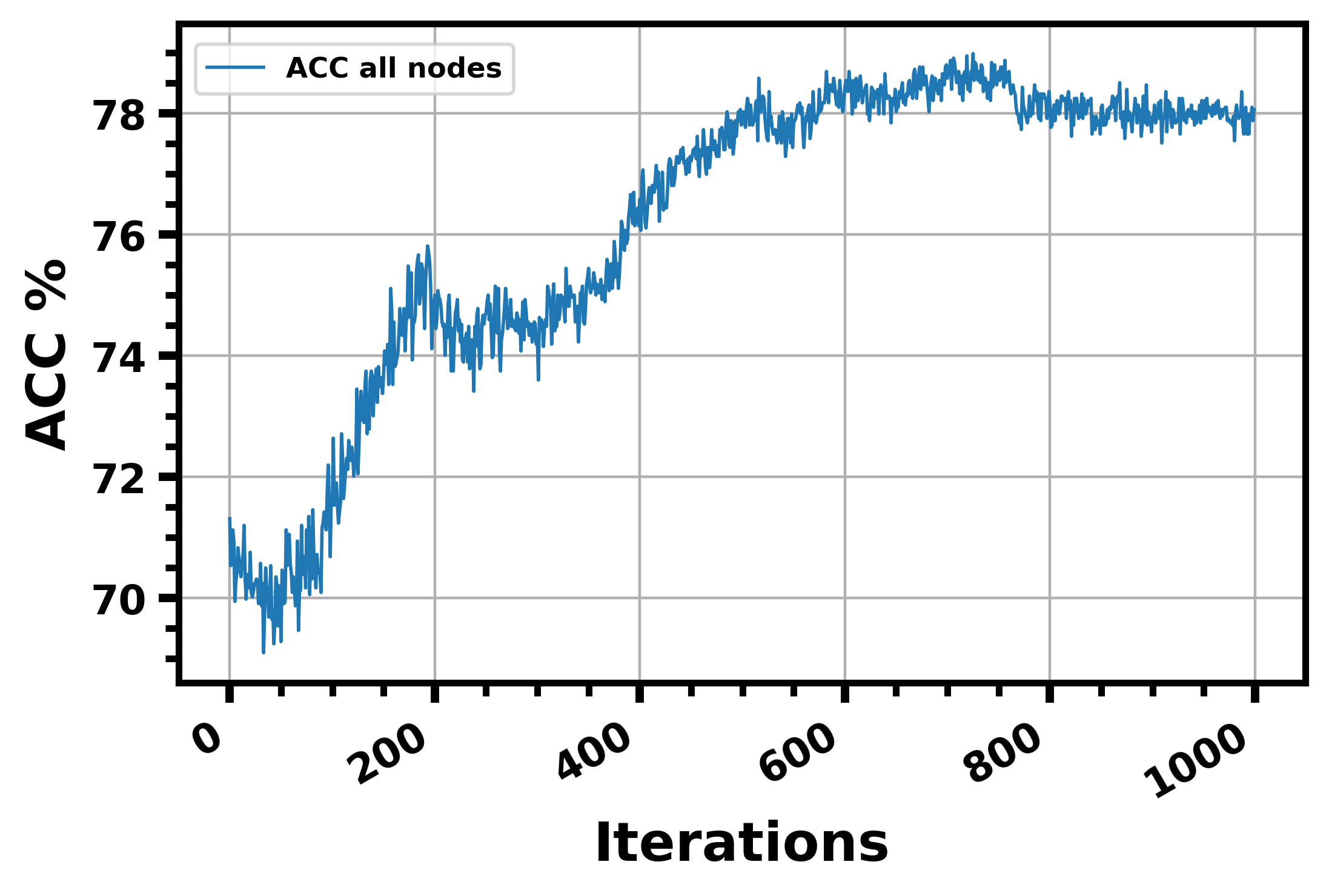}
     \caption{ACC: all nodes}
  \end{subfigure}
  \begin{subfigure}[b]{0.33\textwidth}
    \includegraphics[width=\linewidth]{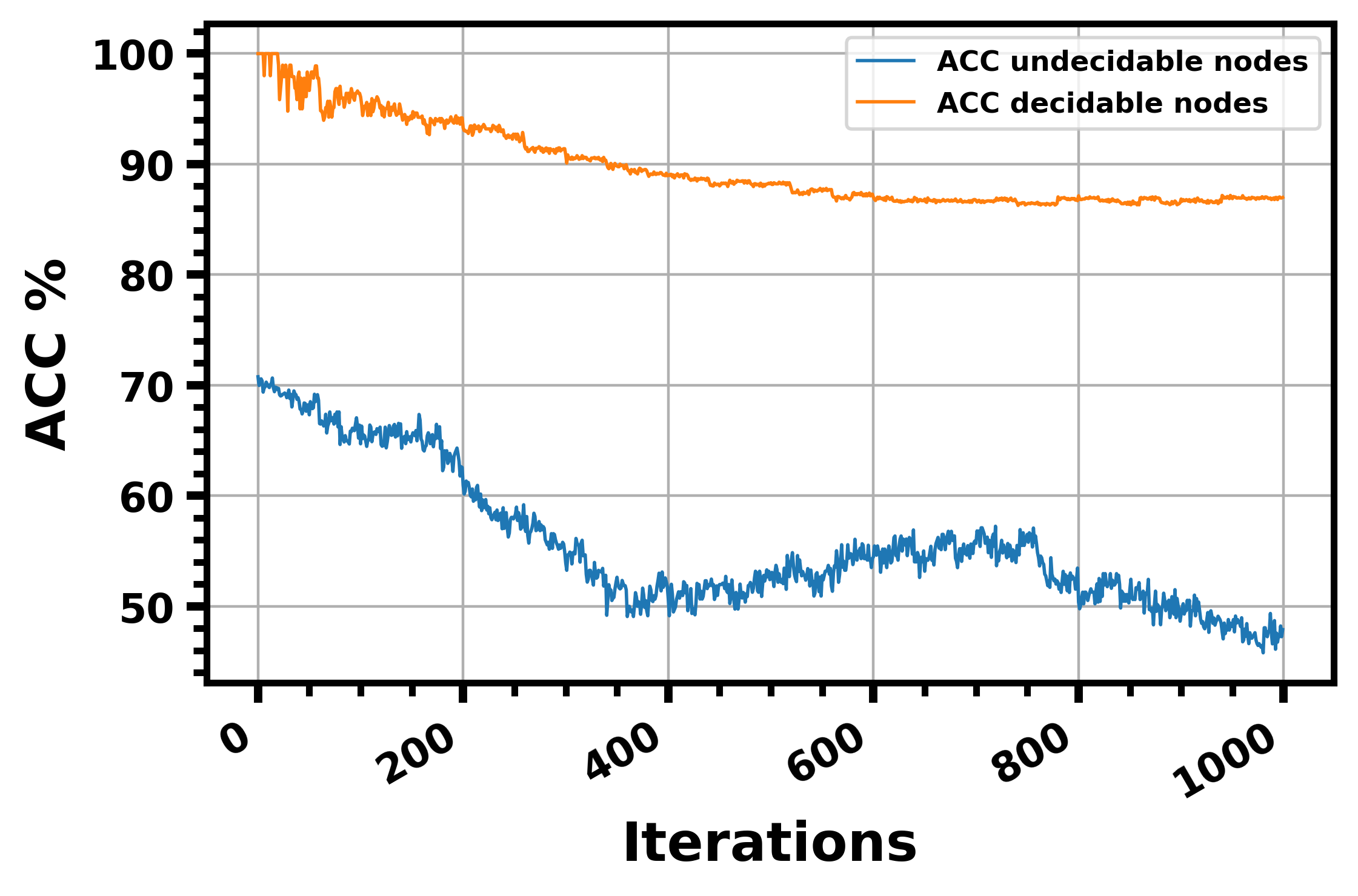}
    \caption{ACC: decidable and undecidable nodes}
  \end{subfigure}
  %\begin{subfigure}[b]{0.33\textwidth}
    %\includegraphics[width=\linewidth]{images/Loss_CVGAE_cora.png}
    %\caption{Loss}
  %\end{subfigure}
  %\begin{subfigure}[b]{0.33\textwidth}
     %\includegraphics[width=\linewidth]{images/NMI_CVGAE_cora.png}
     %\caption{NMI: all nodes}
  %\end{subfigure}
  %\begin{subfigure}[b]{0.33\textwidth}
    %\includegraphics[width=\linewidth]{images/NMI_unconflicted_data_CVGAE_cora.png}
    %\caption{NMI: decidable and undecidable nodes}
  %\end{subfigure}
  \begin{subfigure}[b]{0.33\textwidth}
    \includegraphics[width=\linewidth]{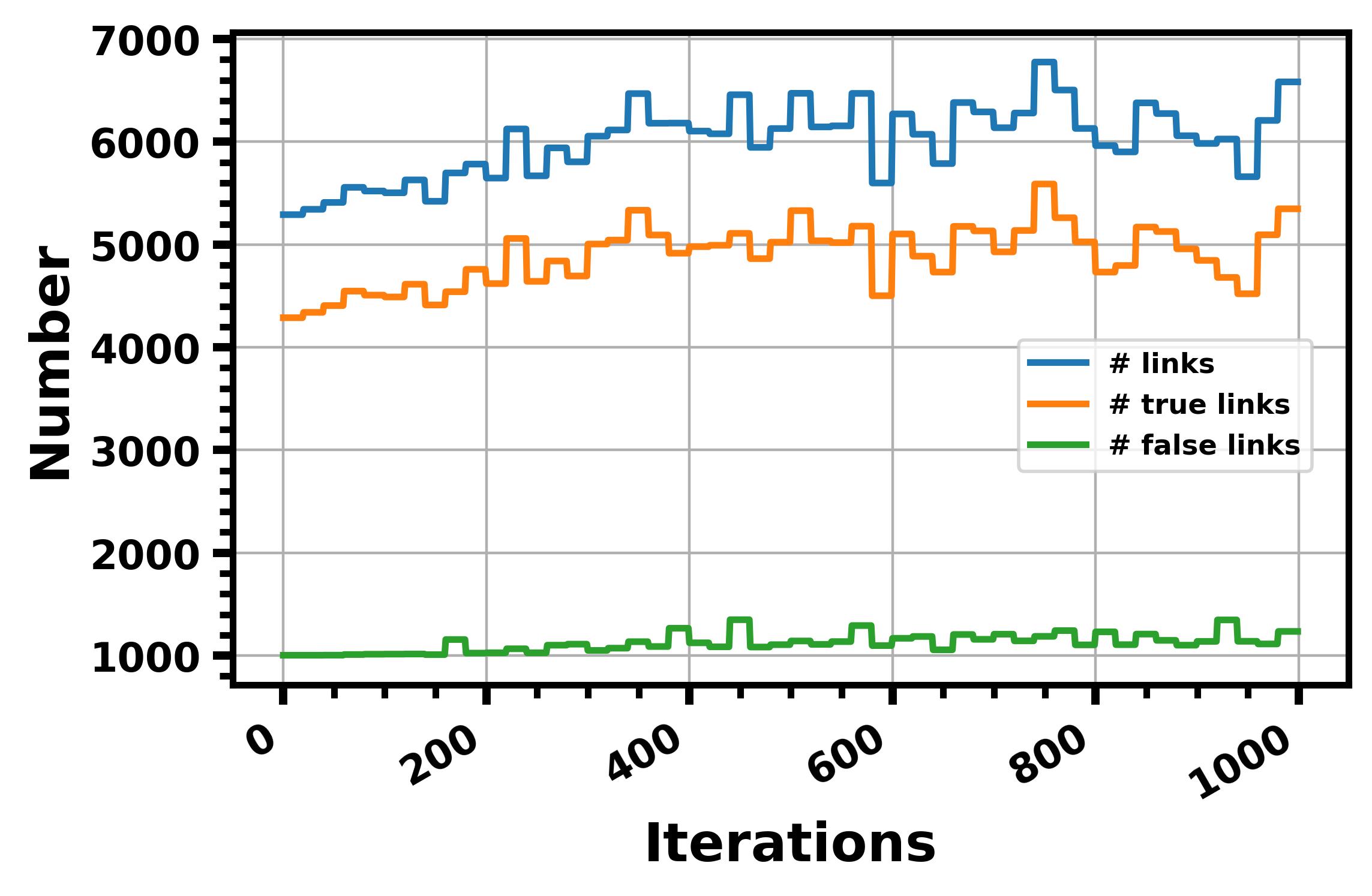}
    \caption{\# links $A^{pos}$}
  \end{subfigure}
  \begin{subfigure}[b]{0.33\textwidth}
     \includegraphics[width=\linewidth]{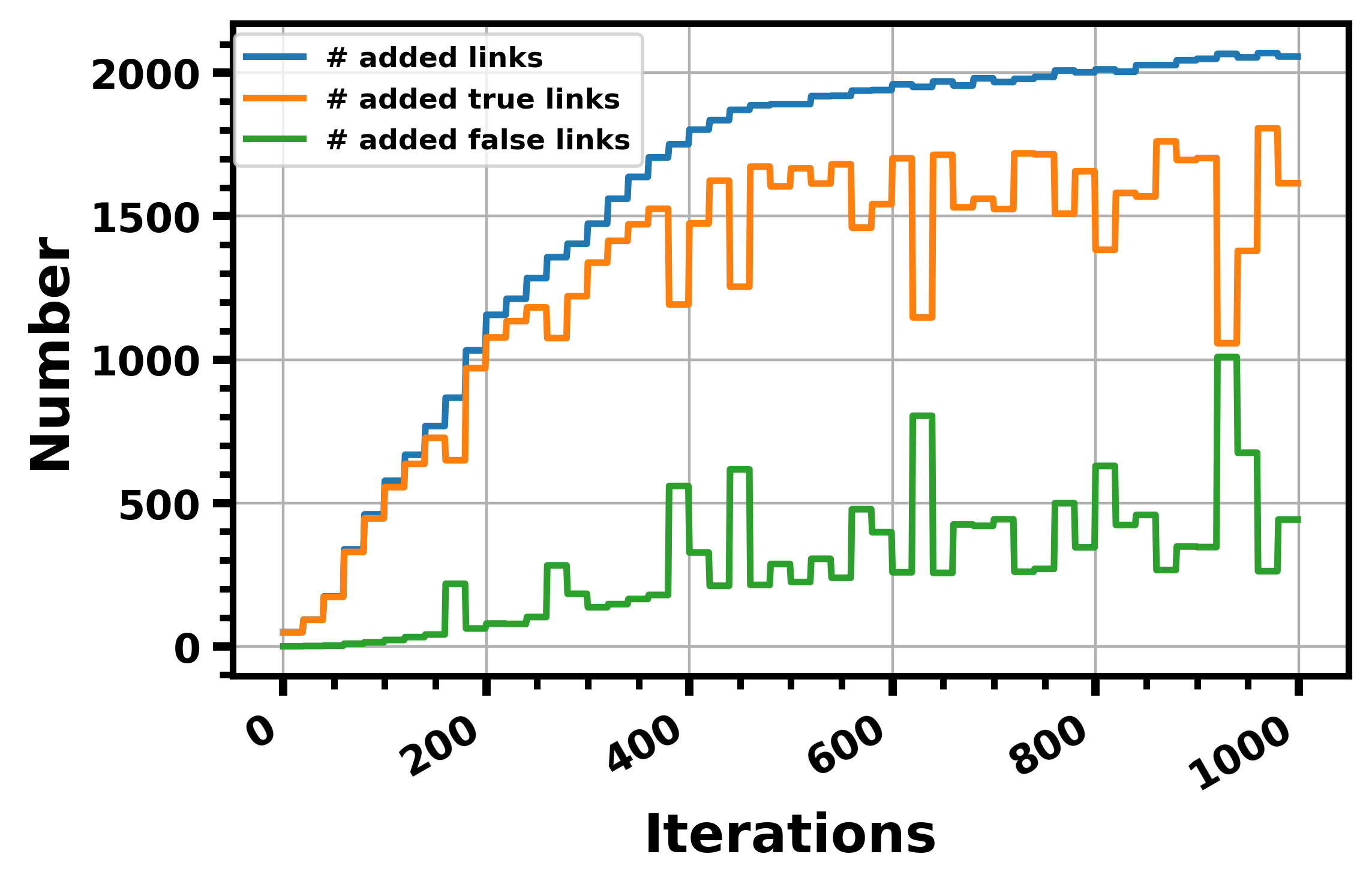}
     \caption{\# added links $A^{pos}$}
  \end{subfigure}
  \begin{subfigure}[b]{0.33\textwidth}
    \includegraphics[width=\linewidth]{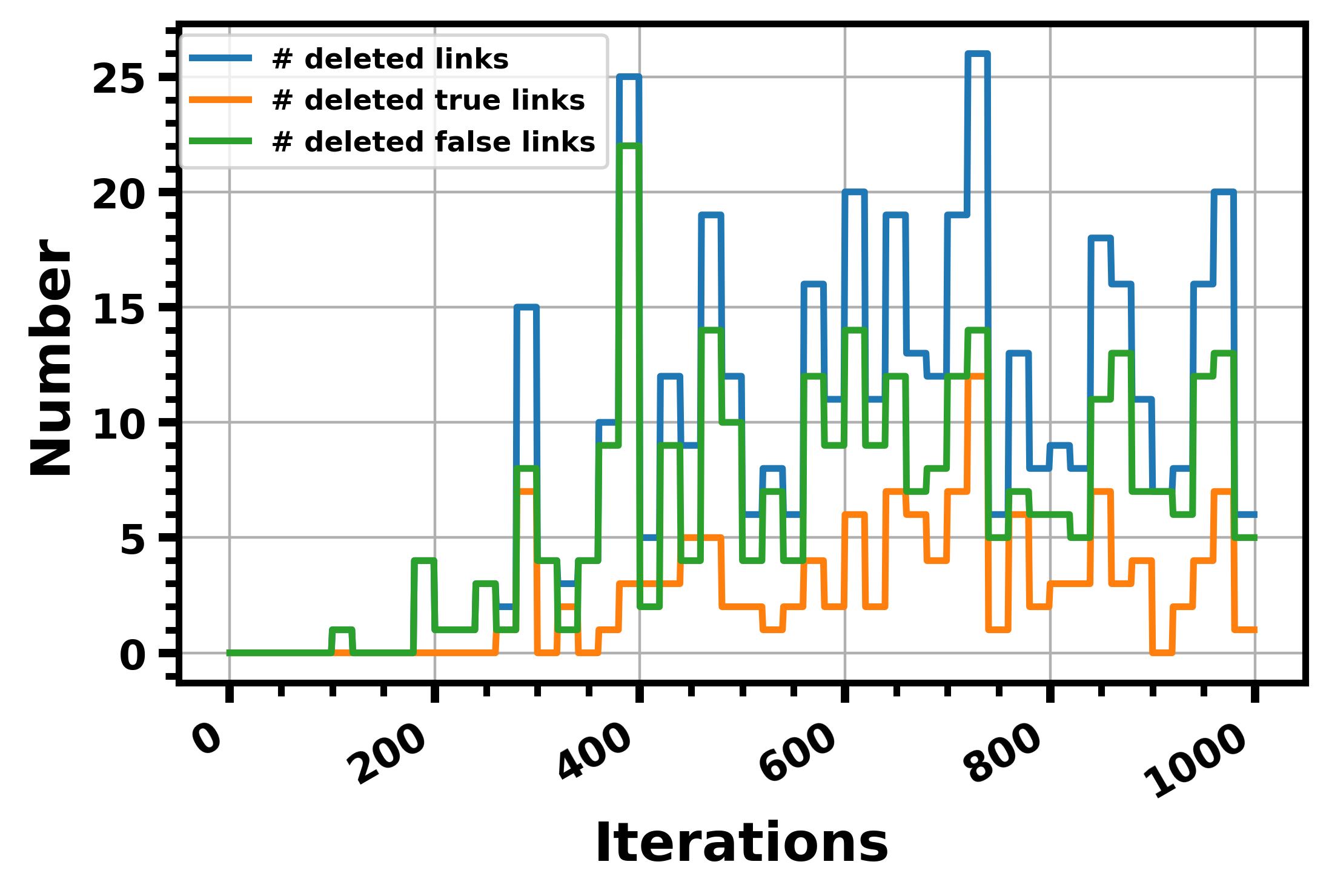}
    \caption{\# deleted links $A^{pos}$}
  \end{subfigure}
  %\vskip 0.1in
  \caption{Learning dynamics of CVGAE on Cora.}
  \label{fig:lr_dyn_ClusVGAE}
\end{figure*}

\textbf{Learning dynamics:} We investigate the learning dynamics of CVGAE on the Cora dataset. In Figure \ref{fig:lr_dyn_ClusVGAE} (a), we can see that the number of decidable nodes (nodes of $\Omega$) gradually increases. In Figure \ref{fig:lr_dyn_ClusVGAE} (b), we illustrate the evolution of ACC during the training process. In Figure \ref{fig:lr_dyn_ClusVGAE} (c), we show the evolution of ACC for two complementary sets (the set of decidable nodes $\Omega$ and the set of undecidable nodes $\mathcal{V} - \Omega$). At the end of the training process, $\Omega$ constitutes $80\%$ of the training set, and it has an ACC higher than $86\%$. The results obtained show that our model can identify \textit{gradually} and \textit{effectively} more reliable nodes for the clustering task.

For each training iteration, the clustering-oriented graph $\mathcal{G}^{pos}$ is constructed from the original graph $\mathcal{G}$. We denote an edge between two nodes in the same cluster as a true link. Similarly, we denote an edge between two nodes in different clusters as a false link. We investigate the evolution of $A^{pos}$ on the basis of three figures. In Figure \ref{fig:lr_dyn_ClusVGAE} (d), we can see that the number of true links generally increases, while the number of false links almost remains fixed. In Figure \ref{fig:lr_dyn_ClusVGAE} (e), we observe that the number of true links among the added links is greater than the number of the added false links. In Figure \ref{fig:lr_dyn_ClusVGAE} (f), we observe that the number of false links among the deleted links is higher than the number of deleted true links. Our results show that our method constructs a graph structure more suitable for the clustering task than the original structure. 

\begin{figure*}[t]
  \centering
  \begin{subfigure}[b]{0.329\textwidth}
     \includegraphics[width=\linewidth]{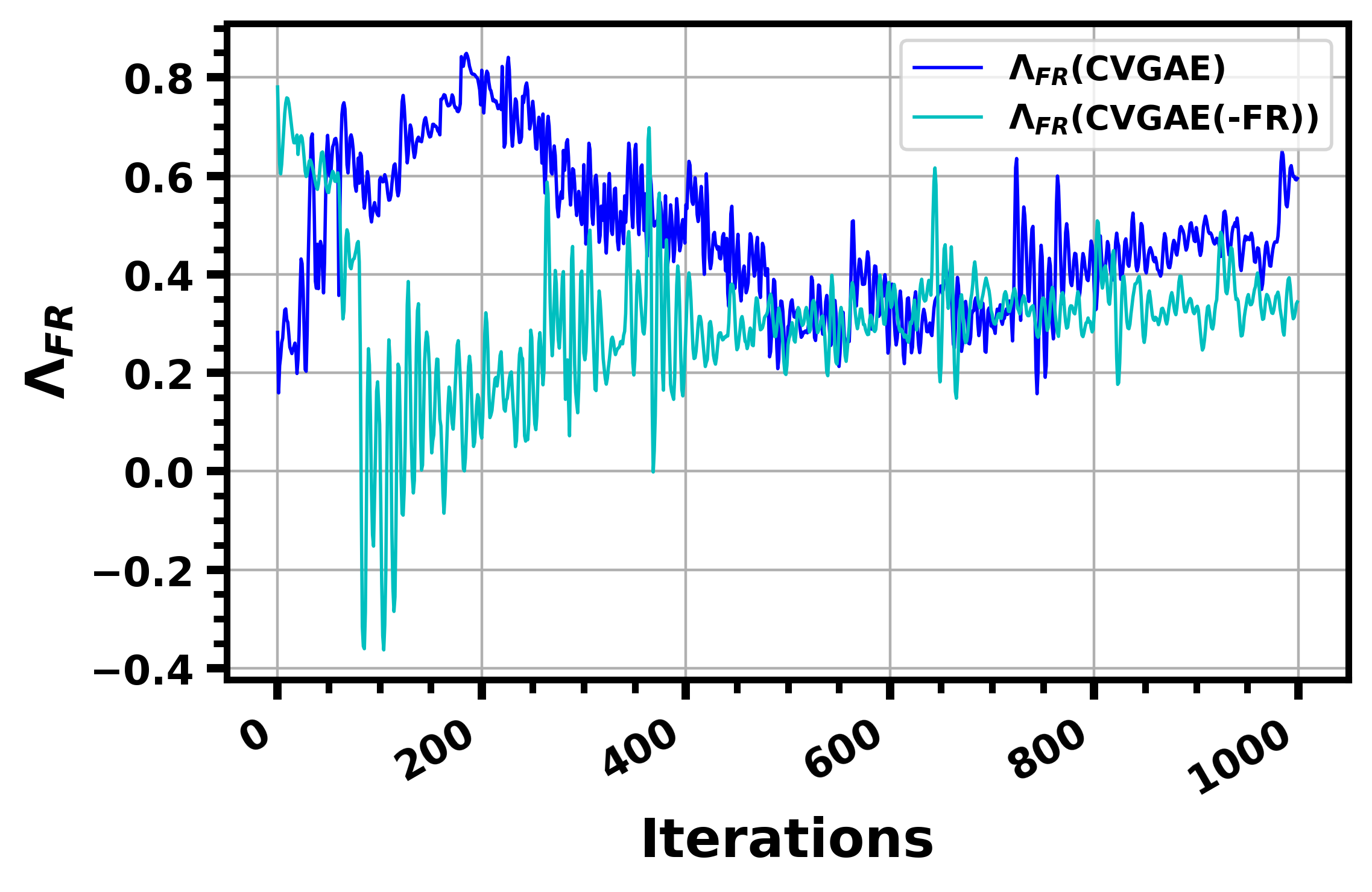}
  \end{subfigure} \hfil
  \begin{subfigure}[b]{0.329\textwidth}
     \includegraphics[width=\linewidth]{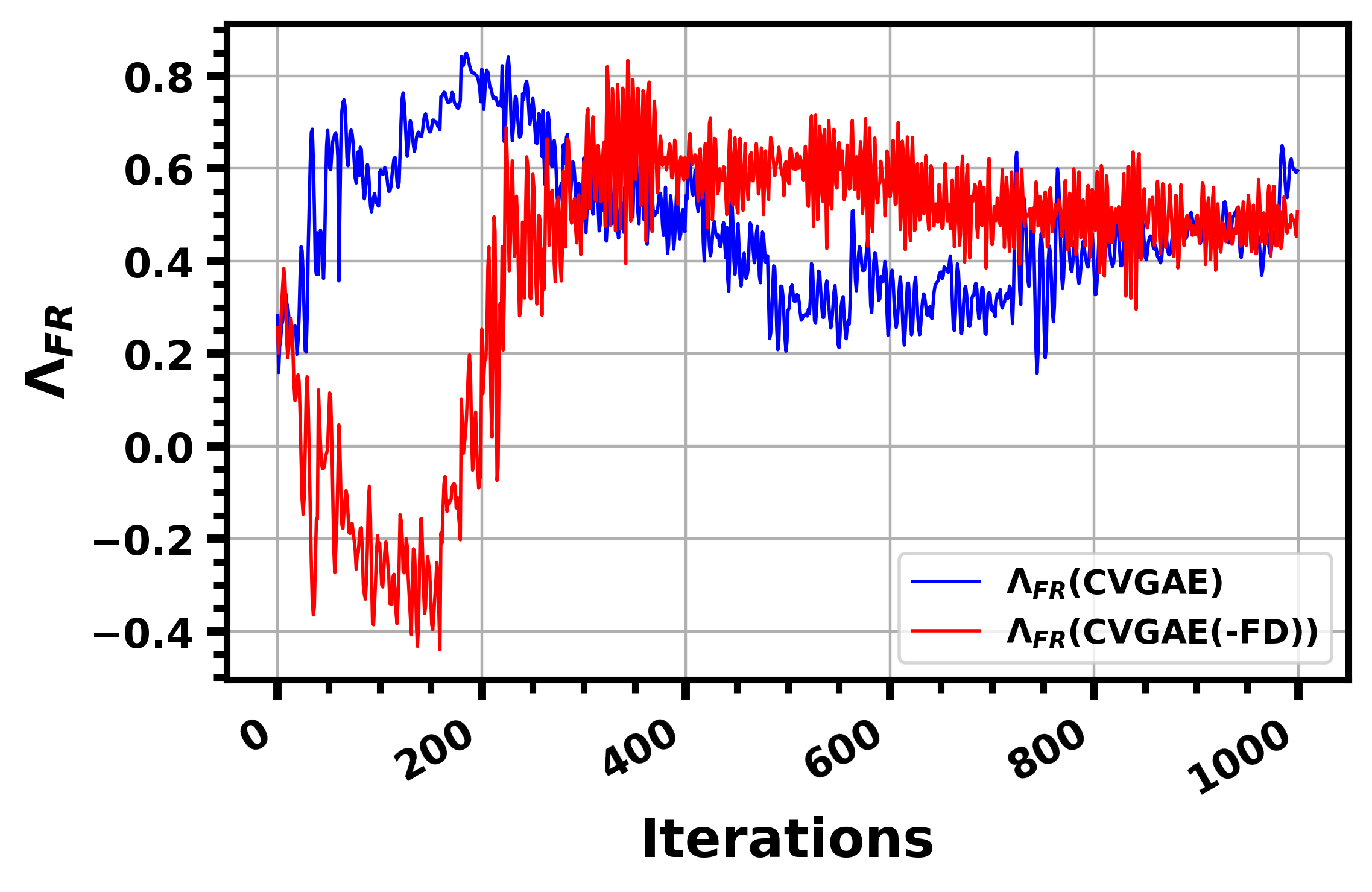}
  \end{subfigure} \hfil
  \begin{subfigure}[b]{0.329\textwidth}
    \includegraphics[width=\linewidth]{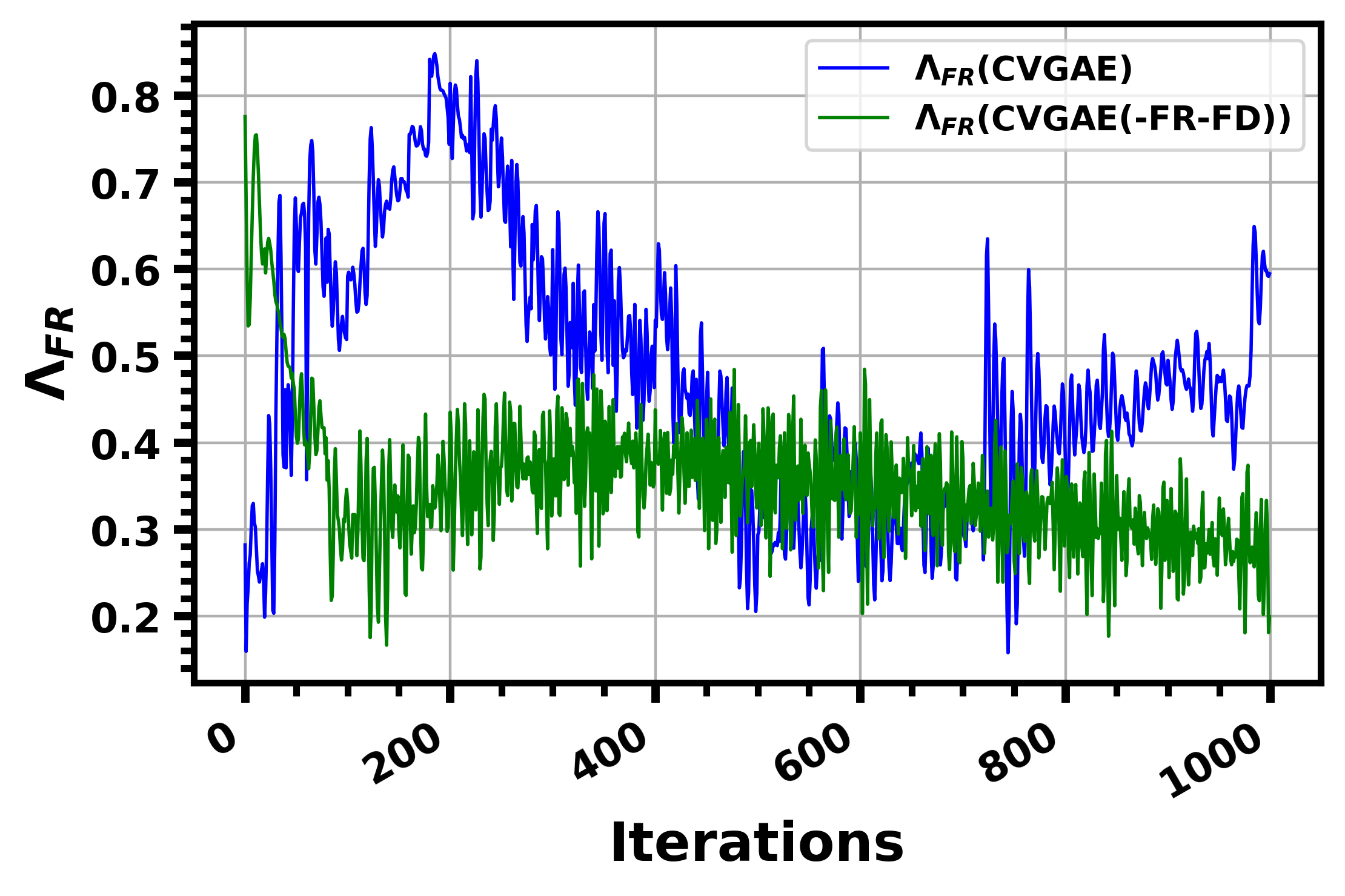}
  \end{subfigure} \hfil
  \medskip
  \begin{subfigure}[b]{0.329\textwidth}
     \includegraphics[width=\linewidth]{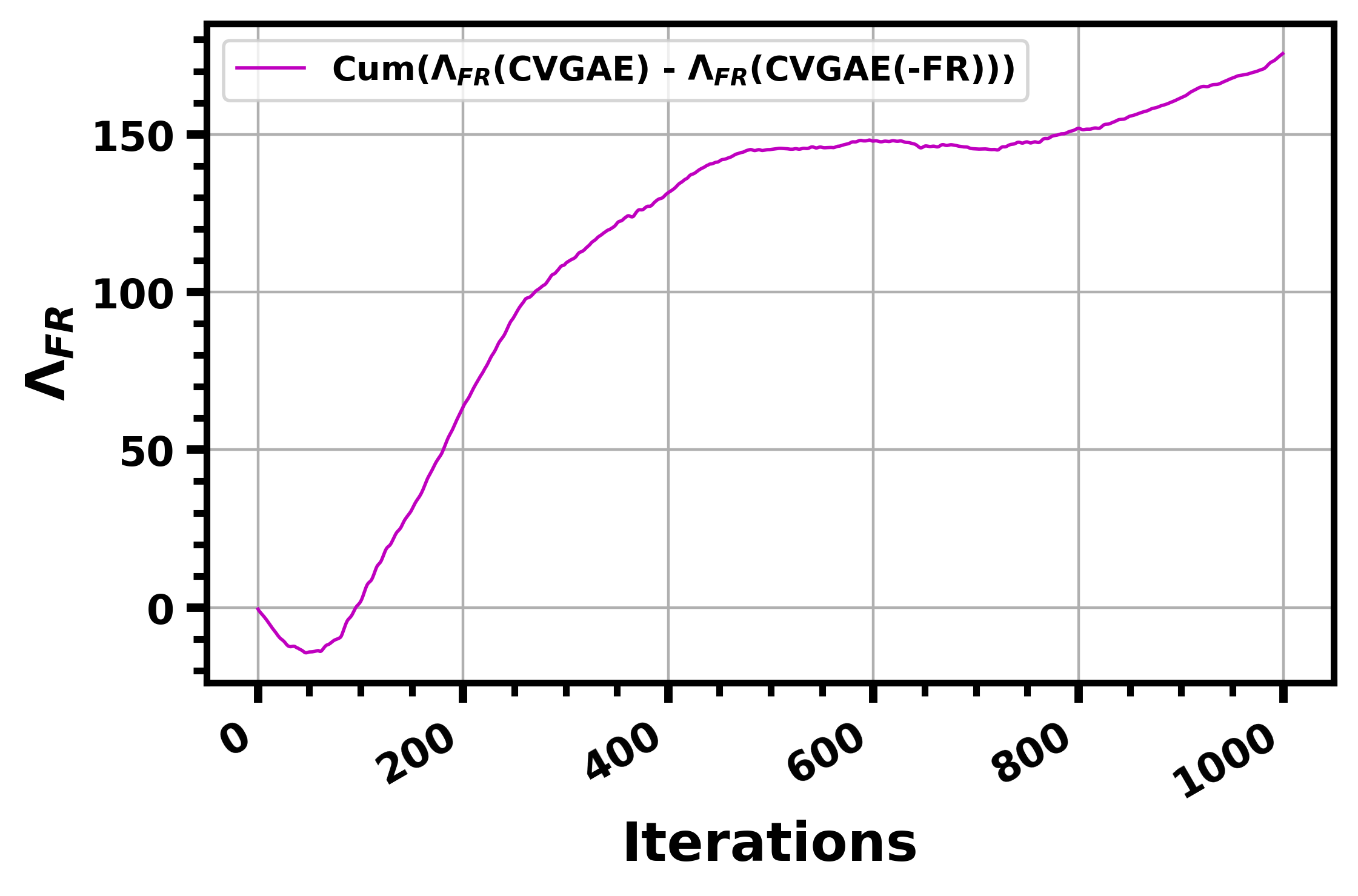}
     \caption{CVGAE vs CVGAE(-FR)}
  \end{subfigure}
  \begin{subfigure}[b]{0.329\textwidth}
     \includegraphics[width=\linewidth]{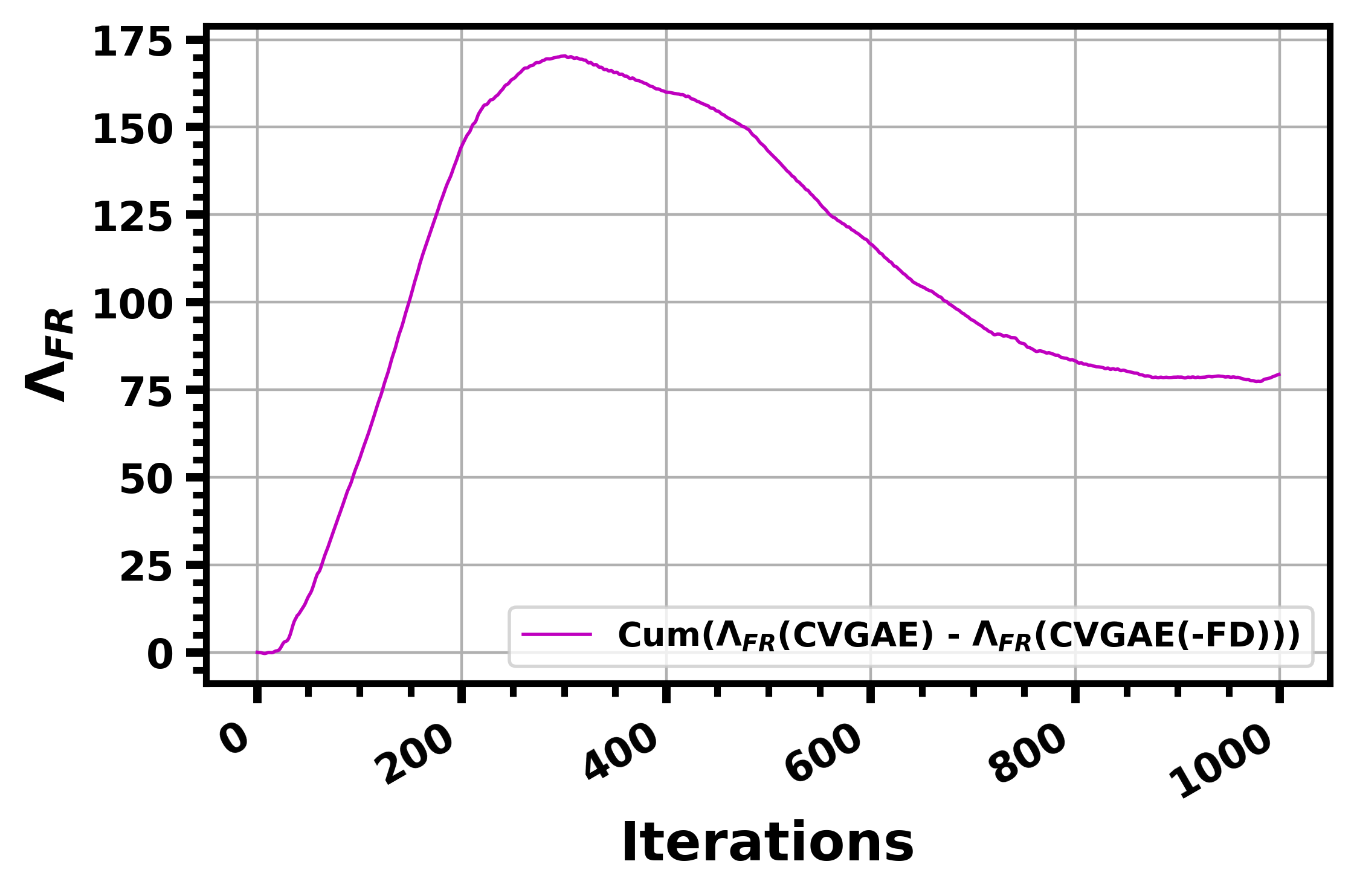}
     \caption{CVGAE vs CVGAE(-FD)}
  \end{subfigure} \hfil
  \begin{subfigure}[b]{0.329\textwidth}
    \includegraphics[width=\linewidth]{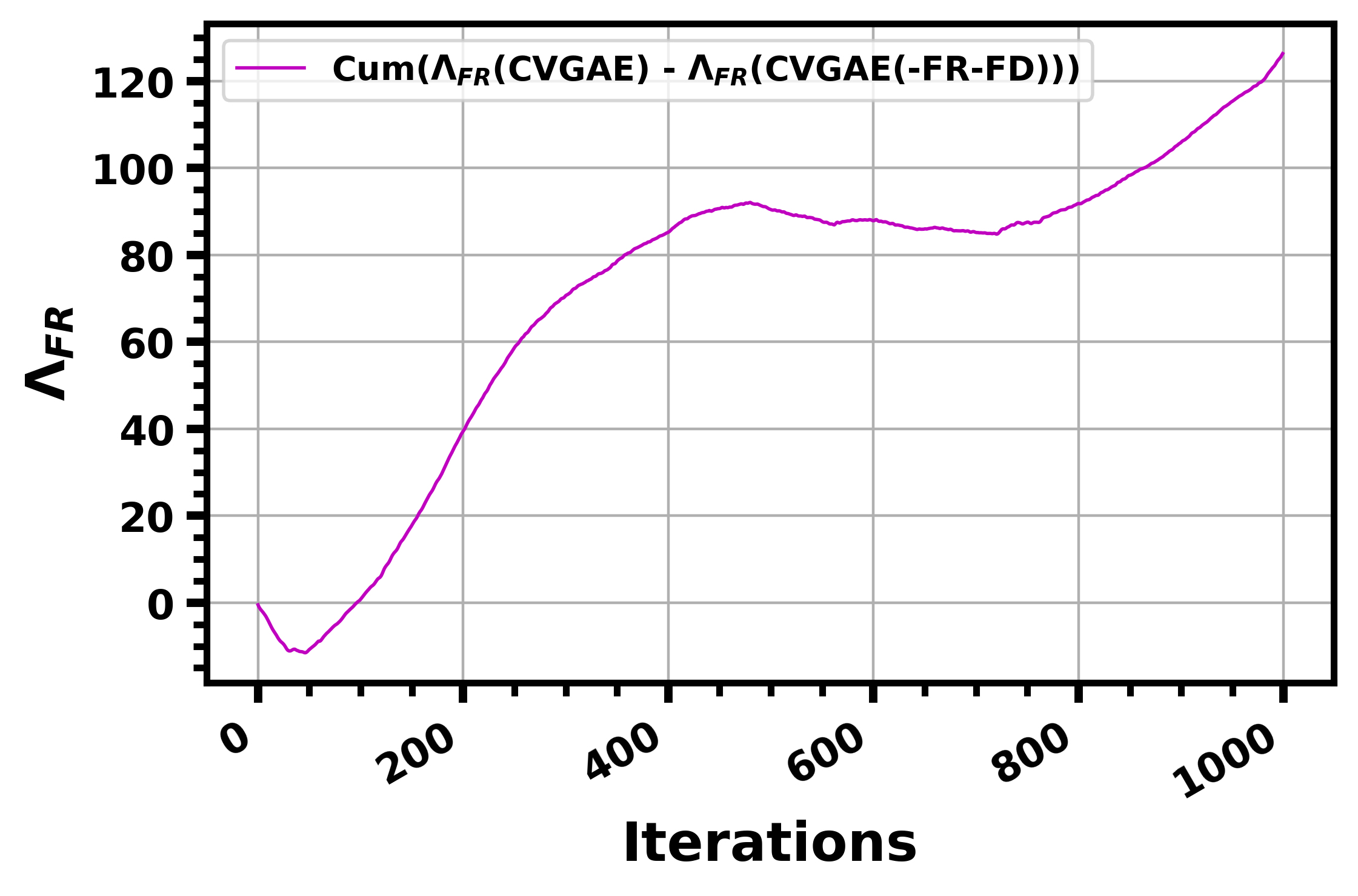}
    \caption{CVGAE vs CVGAE(-FR-FD)}
  \end{subfigure} \hfil
  \caption{Performance of CVGAE and its variants in terms of $\Lambda_{FR}$ on Cora. Blue line: $\Lambda_{FR}$ values of CVGAE. Cyan line: $\Lambda_{FR}$ values of CVGAE(-FR). Red line: $\Lambda_{FR}$ values of CVGAE(-FD). Green line: $\Lambda_{FR}$ values of CVGAE(-FR-FD). Purple line: cumulative difference between $\Lambda_{FR}$ values of CVGAE and $\Lambda_{FR}$ of the other variants.}
  \label{fig:FR_CVGAE}
\end{figure*}

\textbf{Feature Randomness:} We investigate the capacity of our model to mitigate FR. To this end, we evaluate CVGAE against three related variants CVGAE(-FR), CVGAE(-FD), and CVGAE(-FR-FD) in terms of $\Lambda_{FR}$ and the cumulative difference in $\Lambda_{FR}$. All methods are trained independently. In Figure \ref{fig:FR_CVGAE} (a), we compare CVGAE with CVGAE(-FR). At first (from iteration $0$ to $50$), CVGAE(-FR) has very high values of $\Lambda_{FR}$. Furthermore, it has higher $\Lambda_{FR}$ values than CVGAE. The clustering loss of CVGAE(-FR) is computed based on the entire set of nodes. Whereas for CVGAE, a very small set of reliable nodes $\Omega$ is used for computing the clustering loss. Due to the insufficiency of reliable nodes for CVGAE during the first few iterations, CVGAE(-FR) ensures a better approximation to the supervised gradient. Based on some recent findings \cite{paper101, paper102}, we know that neural networks start by learning simple patterns based on earlier layers. These simple patterns remain the same regardless of whether the model is trained with random or target labels \cite{paper103}. Therefore, we can understand why the gradient of the clustering loss for CVGAE(-FR) is a good approximation to the supervised gradient at the beginning of the training process. After that (from iteration $50$ to $1000$), we observe that CVGAE has higher $\Lambda_{FR}$ values than CVGAE(-FR). Globally, the cumulative difference in $\Lambda_{FR}$ has a glaring increasing tendency, which slightly stagnates for a short period of time (between iterations $500$ and $600$). At the end of the training process, the cumulative difference between CVGAE and CVGAE(-FR) in $\Lambda_{FR}$ reaches a score of $180$ in $1,000$ iterations. This result confirms that CVGAE outperforms CVGAE(-FR) in terms of $\Lambda_{FR}$.

In Figure \ref{fig:FR_CVGAE} (b), we compare CVGAE with CVGAE(-FD) in terms of $\Lambda_{FR}$ and the cumulative difference in $\Lambda_{FR}$. we observe that there are three stages. The first stage ranges between iterations $0$ and $300$. For this stage, CVGAE has higher $\Lambda_{FR}$ values than CVGAE(-FD). In other words, gradually building a clustering-oriented graph during the first stage is better at mitigating FR than reconstructing the input graph. The second stage spreads between iterations $300$ and $800$. For this stage, CVGAE (-FD) has higher values of $\Lambda_{FR}$ than CVGAE. The reconstruction loss has a more favorable effect in mitigating FR during the second stage. This is because the gradually constructed graph has already accumulated several noisy edges, making the training process more exposed to FR. The third stage spreads between iterations $800$ and $1,000$. For this stage, CVGAE and CVGAE(-FD) have the same level of $\Lambda_{FR}$. At the end of the training process, the cumulative difference between CVGAE and CVGAE(-FD) in $\Lambda_{FR}$ reaches a score of $75$ over $1,000$ iterations. This result indicates that CVGAE globally outperforms CVGAE(-FD) in terms of $\Lambda_{FR}$. Thus, adequately controlling FD contributes to reducing FR.

In Figure \ref{fig:FR_CVGAE} (c), we compare CVGAE with CVGAE(-FR-FD) in terms of $\Lambda_{FR}$ and the cumulative difference in $\Lambda_{FR}$. At the end of the training process, the cumulative difference between CVGAE and CVGAE(-FR-FD) in $\Lambda_{FR}$ reaches a score of $120$ in $1,000$ iterations. Furthermore, this cumulative difference has a similar variation compared to the cumulative difference in Figure \ref{fig:FR_CVGAE} (a). However, the final score achieved in Figure \ref{fig:FR_CVGAE} (a) ($150$) is higher than the one achieved in Figure \ref{fig:FR_CVGAE} (c) ($120$). We have also found that the cumulative difference cum($\Lambda_{FR}$(CVGAE(-FR-FD))-$\Lambda_{FR}$(CVGAE(-FR))) remains consistently positive. These results indicate that CVGAE(-FR-FD) is globally better than CVGAE(-FR) in mitigating FR, which can be explained by the trade-off between FR and FD.

\begin{figure*}[t]
  \centering
  \begin{subfigure}[b]{0.329\textwidth}
     \includegraphics[width=\linewidth]{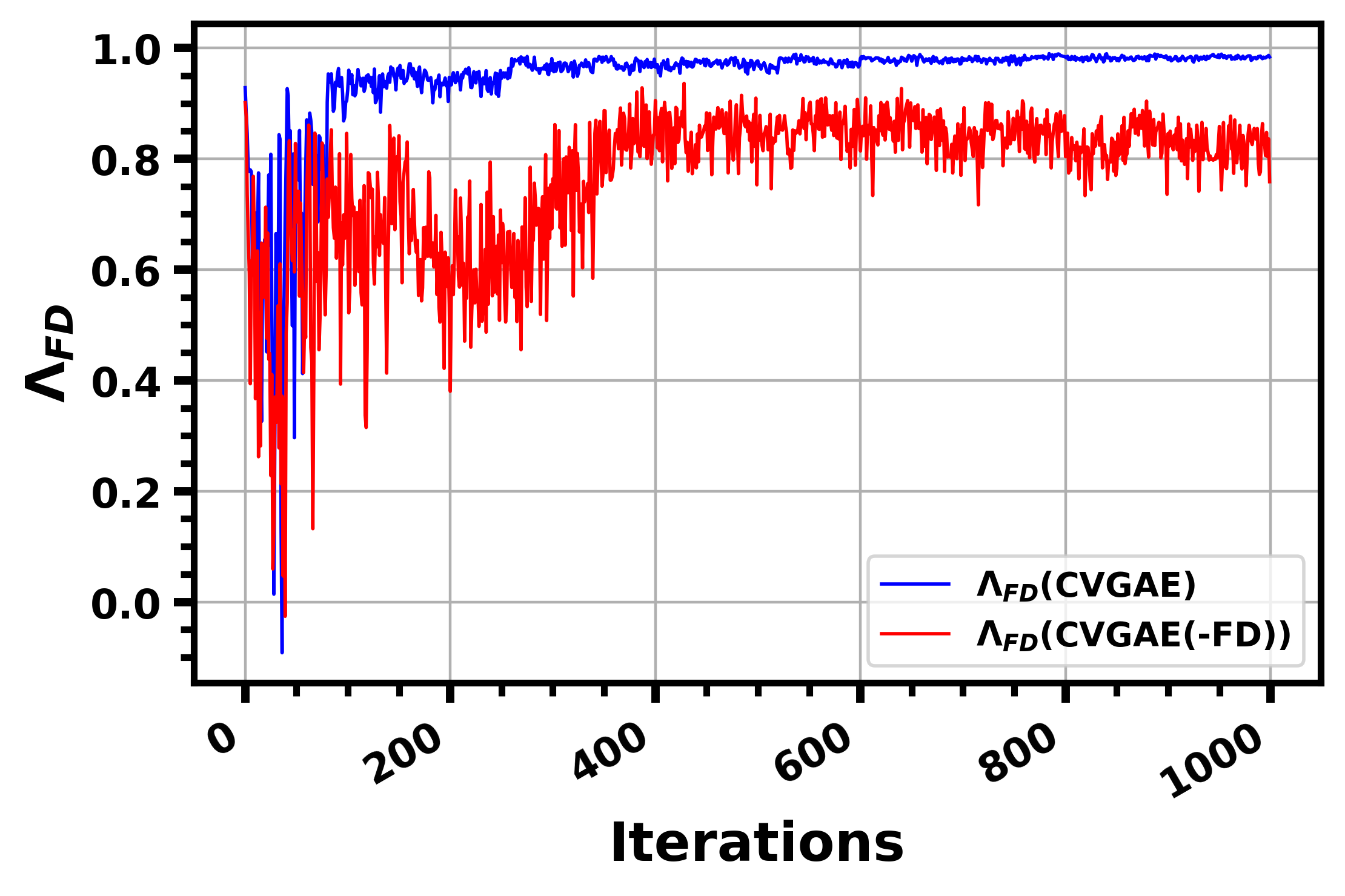}
  \end{subfigure} \hfil
  \begin{subfigure}[b]{0.329\textwidth}
     \includegraphics[width=\linewidth]{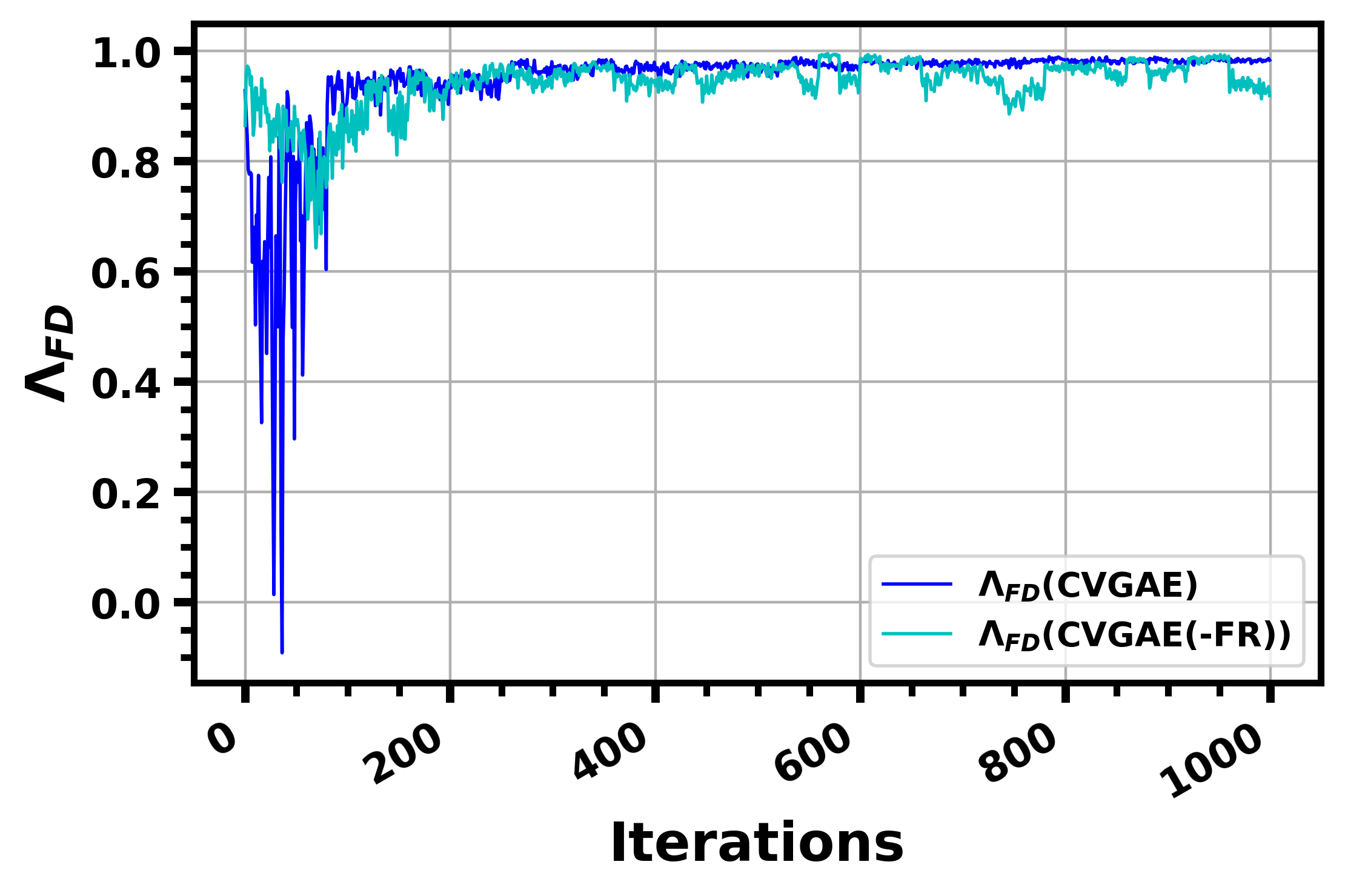}
  \end{subfigure} \hfil
  \begin{subfigure}[b]{0.329\textwidth}
    \includegraphics[width=\linewidth]{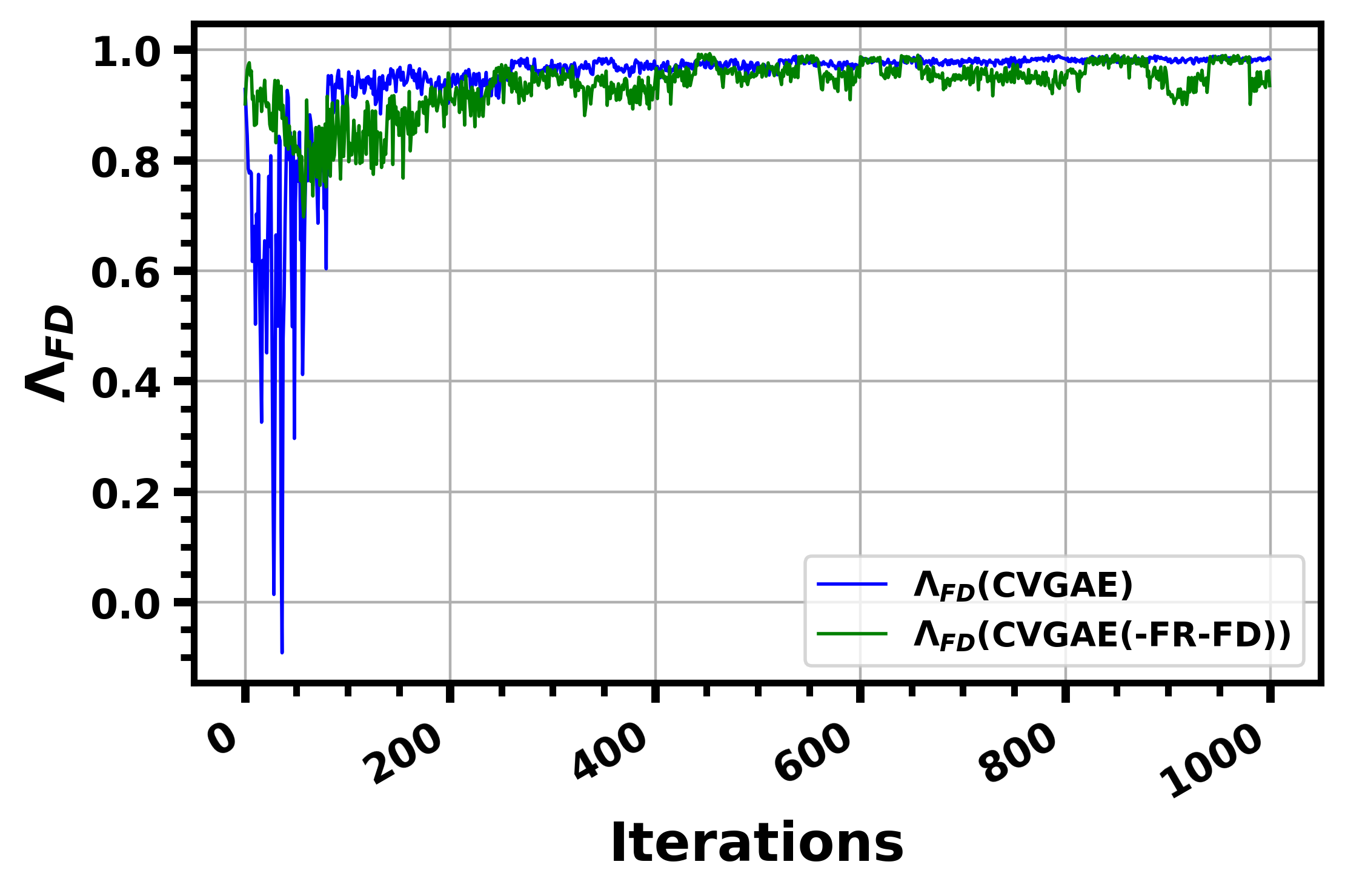}
  \end{subfigure} \hfil
  \medskip
  \begin{subfigure}[b]{0.329\textwidth}
     \includegraphics[width=\linewidth]{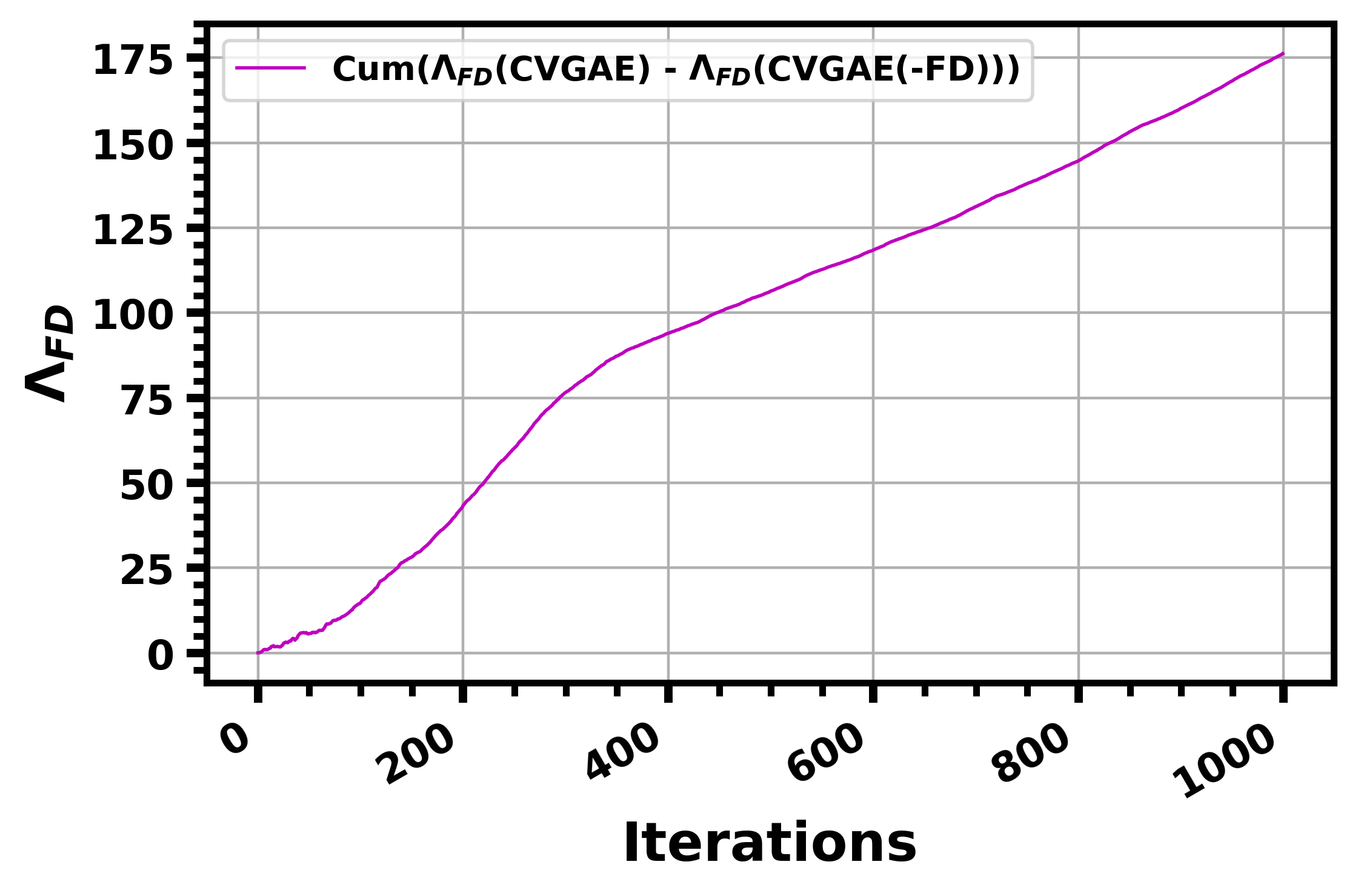}
     \caption{CVGAE vs CVGAE(-FD)}
  \end{subfigure} \hfil
  \begin{subfigure}[b]{0.329\textwidth}
     \includegraphics[width=\linewidth]{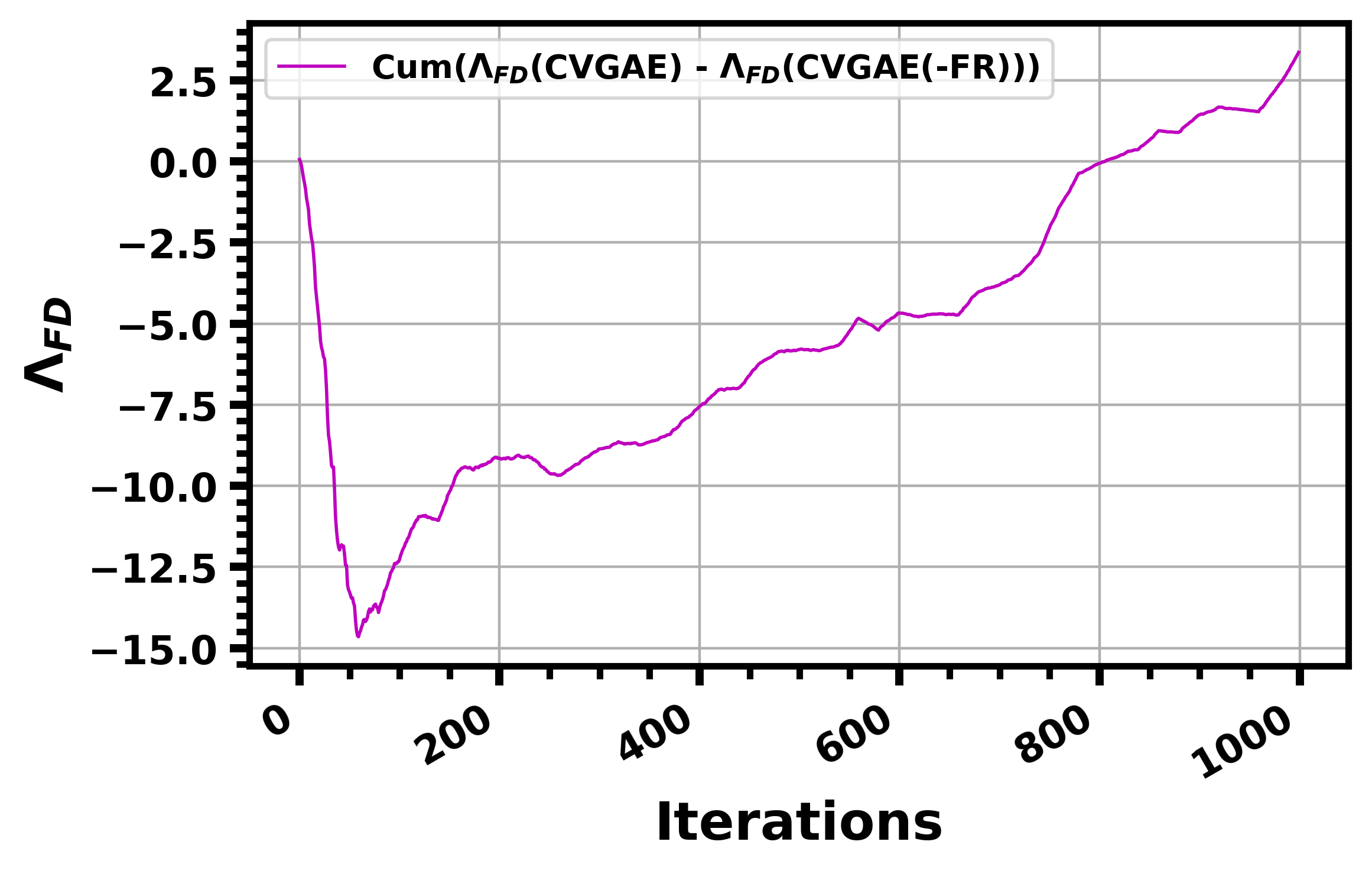}
     \caption{CVGAE vs CVGAE(-FR)}
  \end{subfigure}
  \begin{subfigure}[b]{0.329\textwidth}
    \includegraphics[width=\linewidth]{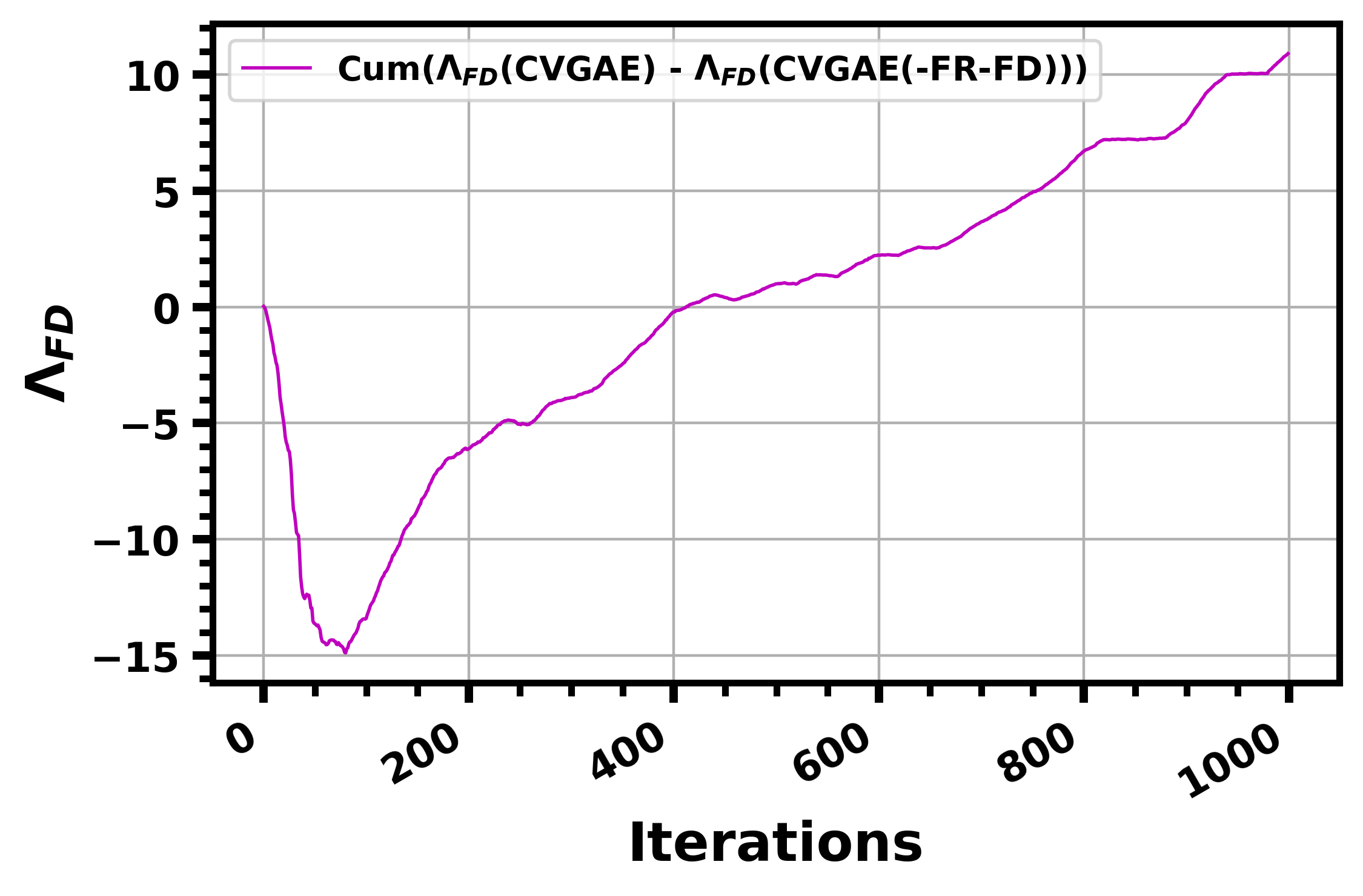}
    \caption{CVGAE vs CVGAE(-FR-FD)}
  \end{subfigure} \hfil
  \caption{Performance of CVGAE and its variants in terms of $\Lambda_{FD}$ on Cora. Blue line: $\Lambda_{FD}$ values of CVGAE. Red line: $\Lambda_{FD}$ values of CVGAE(-FD). Cyan line: $\Lambda_{FR}$ values of CVGAE(-FR). Green line: $\Lambda_{FD}$ values of CVGAE(-FR-FD).   Purple line: normalized cumulative difference between $\Lambda_{FD}$ values of CVGAE and $\Lambda_{FD}$ of the other variants.}
  \label{fig:FD_CVGAE}
\end{figure*}

\textbf{Feature Drift:} We investigate the capacity of our model to mitigate FD. To this end, we evaluate CVGAE against three related variants CVGAE(-FD), CVGAE(-FR), and CVGAE(-FR-FD) in terms of $\Lambda_{FD}$ and the cumulative difference in $\Lambda_{FD}$. All methods are trained independently. In Figure \ref{fig:FD_CVGAE} (a), we compare CVGAE with CVGAE(-FD). We observe that the cumulative difference between CVGAE and CVGAE(-FD) in $\Lambda_{FD}$ has a clear increasing tendency. At the end of the training process, this cumulative difference reaches a score of $175$ over $1,000$ iterations. This result confirms that CVGAE consistently outperforms CVGAE (-FD) in terms of $\Lambda_{FD}$. 

In Figure \ref{fig:FD_CVGAE} (b), we compare CVGAE with CVGAE(-FR) in terms of $\Lambda_{FD}$ and the cumulative difference in $\Lambda_{FD}$. Both models have a mechanism against FD. We observe that there are two stages. The first stage ranges between the iterations $0$ and $50$. For this stage, CVGAE(-FR) has higher $\Lambda_{FD}$ values than CVGAE. FD is an artificially created problem by a pretext task to alleviate FR. Hence, eliminating the FR mechanism strengthens the importance of the FD mechanism. For the second stage, CVGAE(-FD) has slightly lower $\Lambda_{FD}$ values than CVGAE. In Figure \ref{fig:FD_CVGAE} (a), the cumulative difference in $\Lambda_{FD}$ varies between $0$ and $175$. However, this cumulative difference varies in a much smaller interval (between $-15$ and $2.5$) in Figure \ref{fig:FR_CVGAE} (b). This result shows that the FR mechanism has a low impact on $\Lambda_{FD}$ compared with the FD mechanism. %The opposite is not true (the FD mechanism has a significant impact on $\Lambda_{FR}$).

In Figure \ref{fig:FD_CVGAE} (c), we compare CVGAE with CVGAE(-FR-FD) in terms of $\Lambda_{FD}$ and the cumulative difference in $\Lambda_{FD}$. In Figure \ref{fig:FD_CVGAE} (a), the cumulative difference in $\Lambda_{FD}$ varies between $0$ and $175$. However, this cumulative difference varies in a much smaller interval (between $-15$ and $10$) in Figure \ref{fig:FR_CVGAE} (c). This result shows that  the FD mechanism has a little impact on $\Lambda_{FD}$ \textit{in the absence} of a FR mechanism.

\begin{figure*}[!h]
  \centering
  \begin{subfigure}[b]{0.329\textwidth}
     \includegraphics[width=\linewidth]{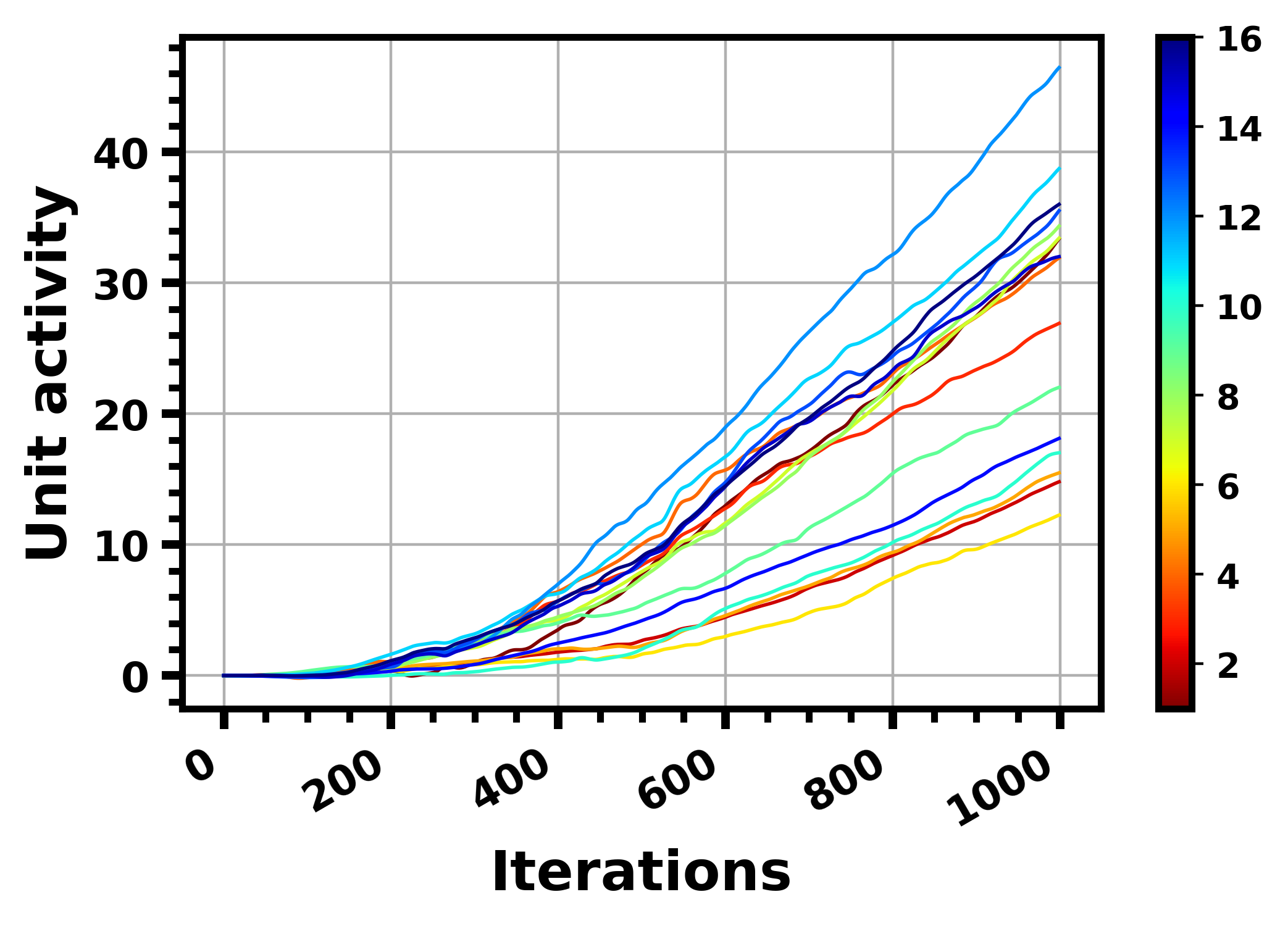}
     \caption{Cora}
  \end{subfigure} \hfil
  \begin{subfigure}[b]{0.329\textwidth}
     \includegraphics[width=\linewidth]{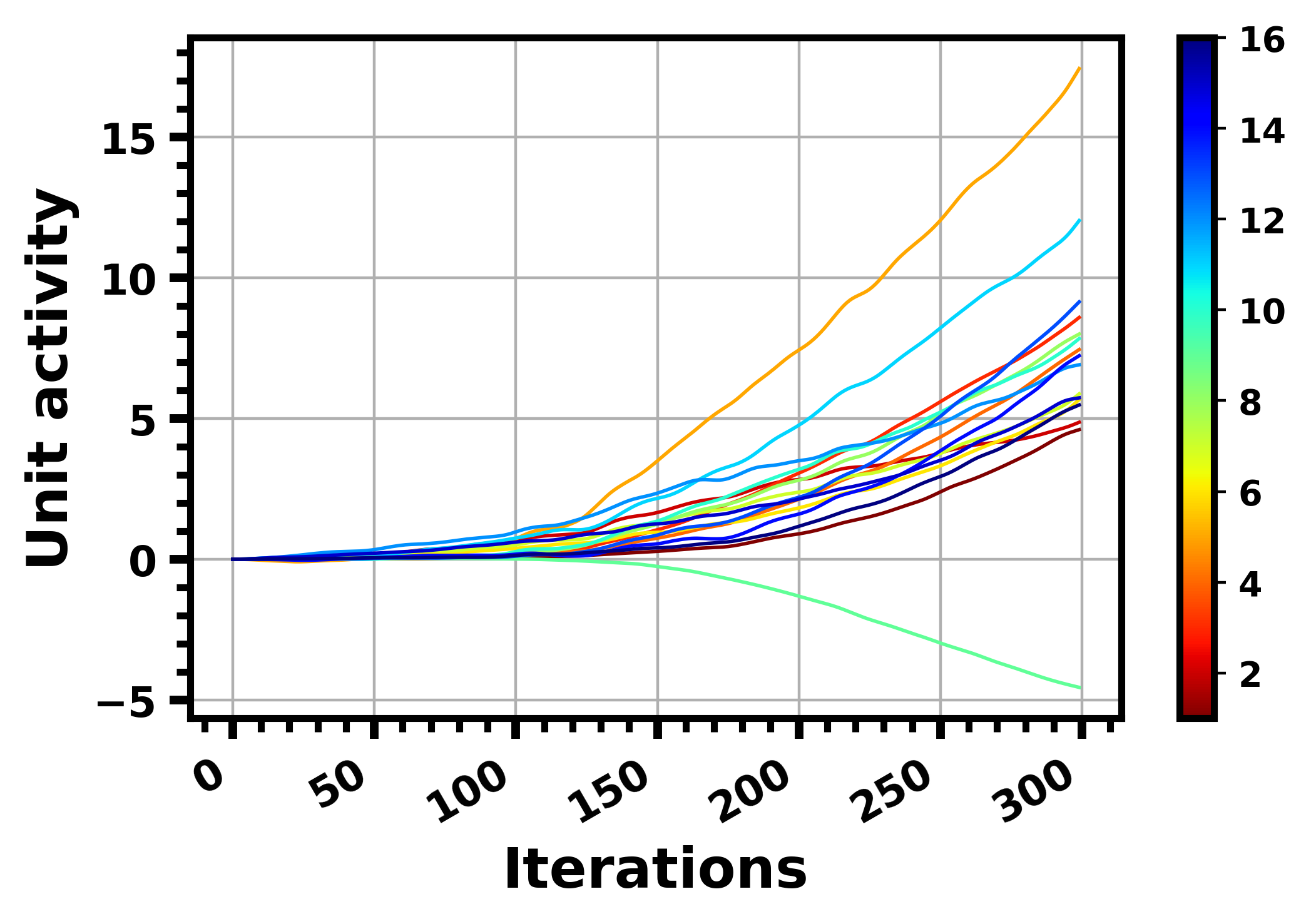}
     \caption{Citeseer}
  \end{subfigure} \hfil
  \begin{subfigure}[b]{0.329\textwidth}
    \includegraphics[width=\linewidth]{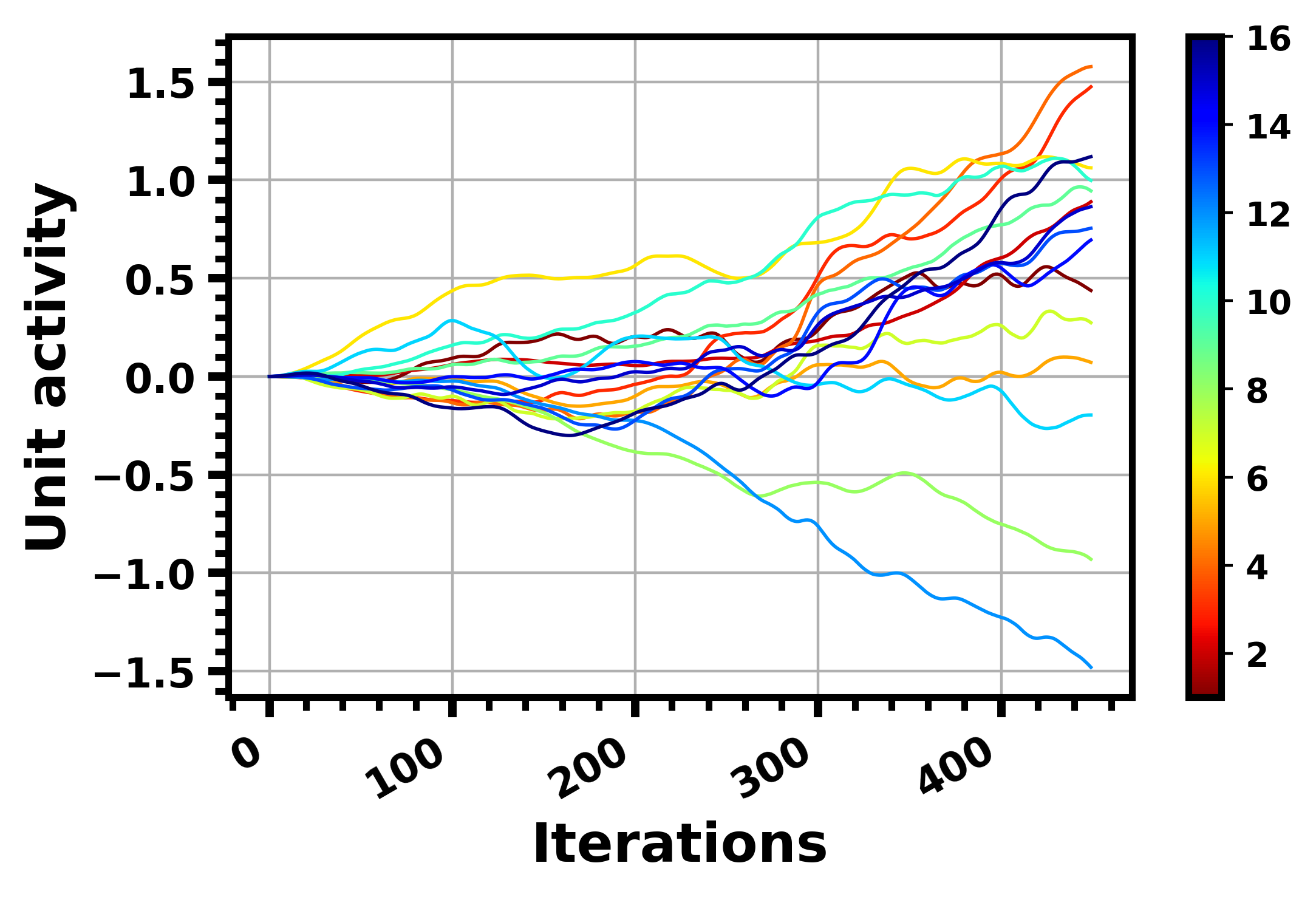}
    \caption{Wiki}
  \end{subfigure} \hfil
  \caption{Unit activity of CVGAE and its variant CVGAE(-PC). Each line represents the cumulative difference between $AU$(CVAGE) and $AU$(CVAGE(-PC)) for a single unit among the $16$ neurons of the latent space.}
  \label{fig:PC_CVGAE}
\end{figure*}

\textbf{Posterior Collapse:} We investigate the capacity of our model to mitigate PC. In Figure \ref{fig:PC_CVGAE}, we illustrate the cumulative difference between $AU$(CVAGE) and $AU$(CVAGE(-PC)) for each individual unit among the $16$ neurons of the latent space. We observe that this cumulative difference has an increasing tendency for most units. These results confirm that the term $KL(q(z_{i} | X^{pos}, A^{pos}) \: \|  \: q(z_{i} | X^{neg}, A^{neg}))$ increases the activity of most latent units. Consequently, our model alleviates the PC problem compared to CVAGE(-PC).
%decrease the number of inactive units, which are limited to minimizing the regularization term. Therefore, our model has the capacity to alleviate the PC problem.

\begin{table*}[!h]
  \caption{Ablation study of CVGAE on Cora, Citeseer, and Pubmed.}
  \begin{center}
  \begin{small}
  \scalebox{0.89}{\begin{tabular}{|c|c|c|c|c|c|c|c|c|c|c|c|c|}
    \hline
    \multicolumn{4}{|c|}{\textbf{Method}} & \multicolumn{3}{c|}{\textbf{Cora}} & \multicolumn{3}{c|}{\textbf{Citeseer}} & \multicolumn{3}{c|}{\textbf{Pubmed}}  \\
    \cline{1-13}
    \textbf{FR mechanism} & \textbf{FD mechanism} & \textbf{PC mechanism} & \textbf{Contrastive learning} & \textbf{ACC} & \textbf{NMI} & \textbf{ARI} & \textbf{ACC} & \textbf{NMI} & \textbf{ARI} & \textbf{ACC} & \textbf{NMI} & \textbf{ARI}  \\ \hline
    \xmark  & \xmark & \xmark & \xmark & 73.2 & 53.4 & 50.1 & 70.4 & 44.0 & 46.3 & 69.5 & 35.4 & 30.2 \\ \hline
    \xmark  & \xmark & \cmark & \xmark & 73.3 & 53.5 & 50.1 & 70.4 & 44.1 & 46.3 & 69.6 & 35.3 & 30.4 \\ \hline
    \xmark  & \cmark & \xmark & \cmark & 73.1 & 53.3 & 49.9 & 70.4 & 43.9 & 46.2 & 71.0 & 36.3 & 34.1 \\ \hline
    \xmark  & \cmark & \cmark & \cmark & 73.5 & 53.6 & 50.6 & 70.5 & 44.3 & 46.5 & 71.1 & 36.5 & 34.2 \\ \hline
    \cmark  & \xmark & \xmark & \xmark & 75.7 & 55.2 & 51.5 & 66.2 & 38.4 & 40.1 & 70.0 & 35.2 & 30.9 \\ \hline
    \cmark  & \xmark & \cmark & \xmark & 75.8 & 55.3 & 51.8 & 67.3 & 39.8 & 41.7 & 70.7 & 32.4 & 33.4 \\ \hline
    \cmark  & \cmark & \xmark & \xmark & 76.1 & 55.5 & 52.9 & 69.7 & 42.7 & 44.9 & 73.8 & 33.5 & 37.0 \\ \hline
    \cmark  & \cmark & \xmark & \cmark & 77.3 & 57.3 & 55.3 & 71.2 & 44.5 & 46.8 & 74.2 & 34.8 & 37.9 \\ \hline
    \cmark  & \cmark & \cmark & \cmark & \textbf{79.0} & \textbf{59.9} & \textbf{58.8} & \textbf{71.8} & \textbf{45.4} & \textbf{47.7} & \textbf{74.4} & \textbf{34.8} & \textbf{38.2} \\ \hline
  \end{tabular}}
  \end{small} 
  \end{center}
  \label{Table:ablation_study}
\end{table*}

\begin{figure*}[!h]
  \begin{subfigure}[b]{0.19\textwidth}
    \includegraphics[width=\linewidth]{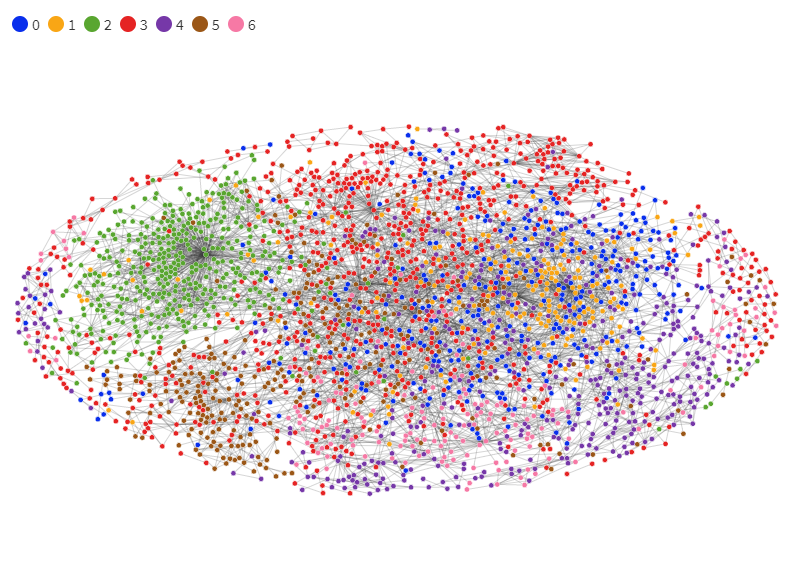}
    \caption{Epoch 0}
  \end{subfigure}
  \begin{subfigure}[b]{0.19\textwidth}
    \includegraphics[width=\linewidth]{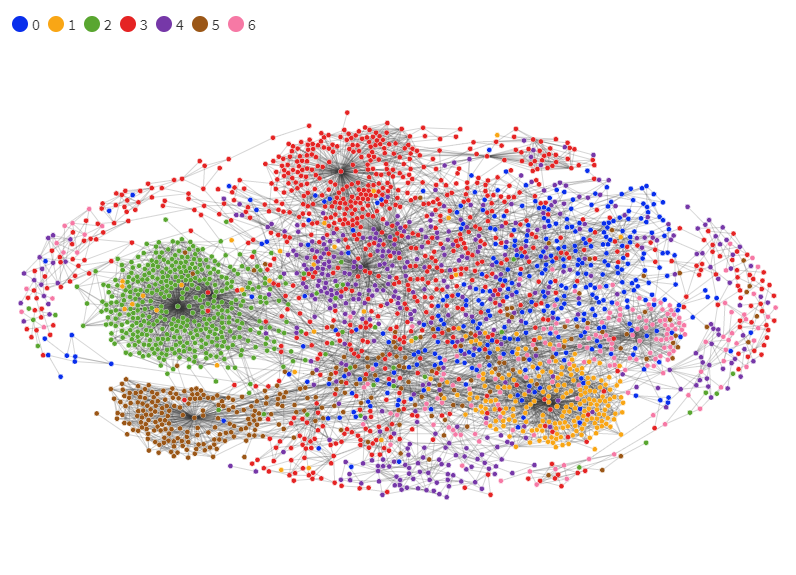}
    \caption{Epoch 200}
  \end{subfigure}
  \begin{subfigure}[b]{0.19\textwidth}
    \includegraphics[width=\linewidth]{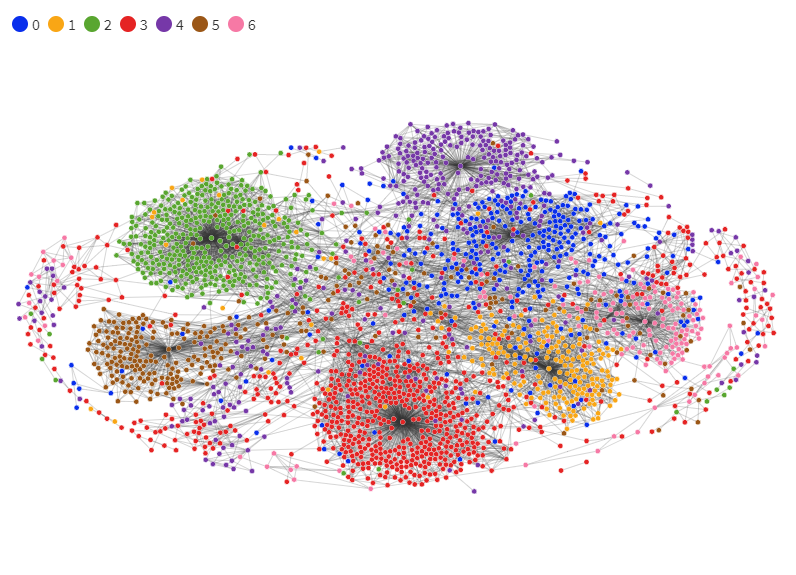}
    \caption{Epoch 400}
  \end{subfigure}
  \begin{subfigure}[b]{0.19\textwidth}
    \includegraphics[width=\linewidth]{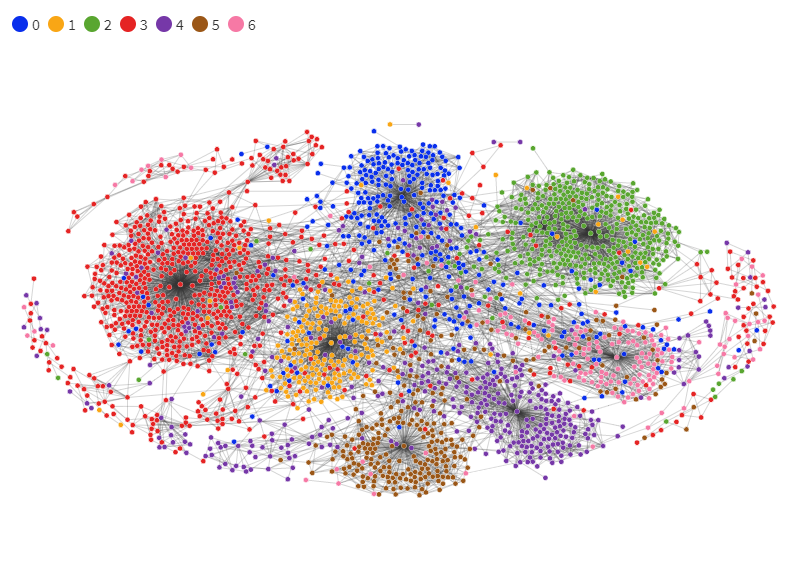}
    \caption{Epoch 600}
  \end{subfigure}
  \begin{subfigure}[b]{0.19\textwidth}
    \includegraphics[width=\linewidth]{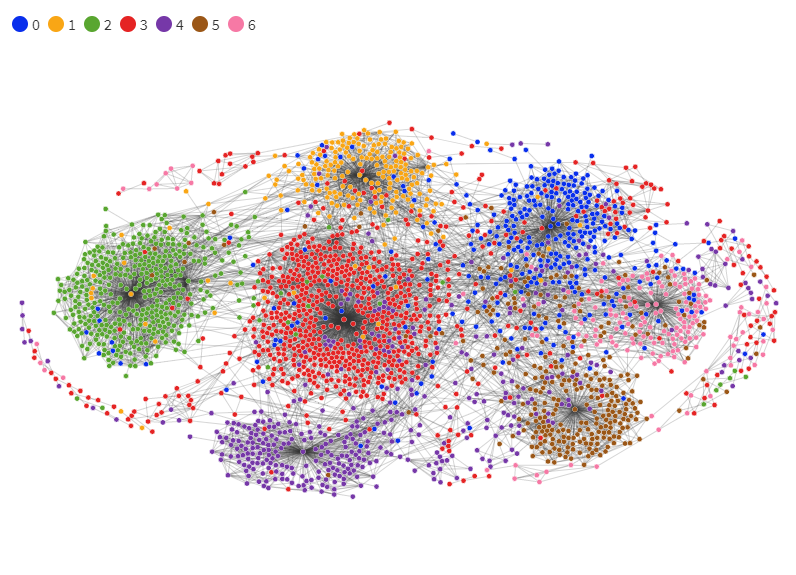}
    \caption{Epoch 800}
  \end{subfigure}
  \medskip
  \begin{subfigure}[b]{0.19\textwidth}
    \includegraphics[width=\linewidth]{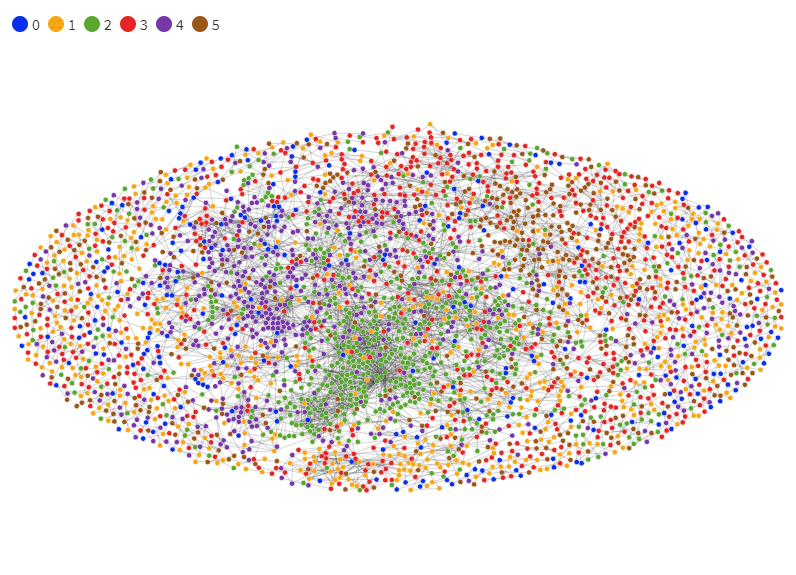}
    \caption{Epoch 0}
  \end{subfigure}
  \begin{subfigure}[b]{0.19\textwidth}
    \includegraphics[width=\linewidth]{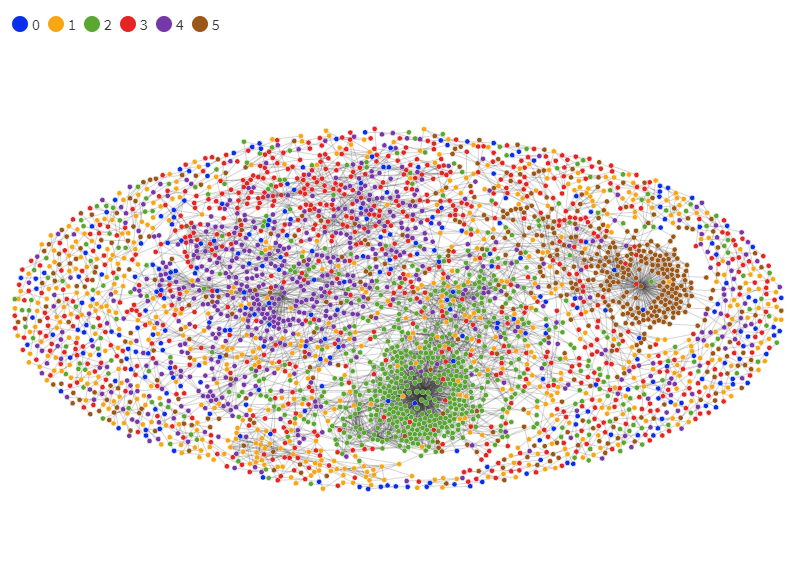}
    \caption{Epoch 100}
  \end{subfigure}
  \begin{subfigure}[b]{0.19\textwidth}
    \includegraphics[width=\linewidth]{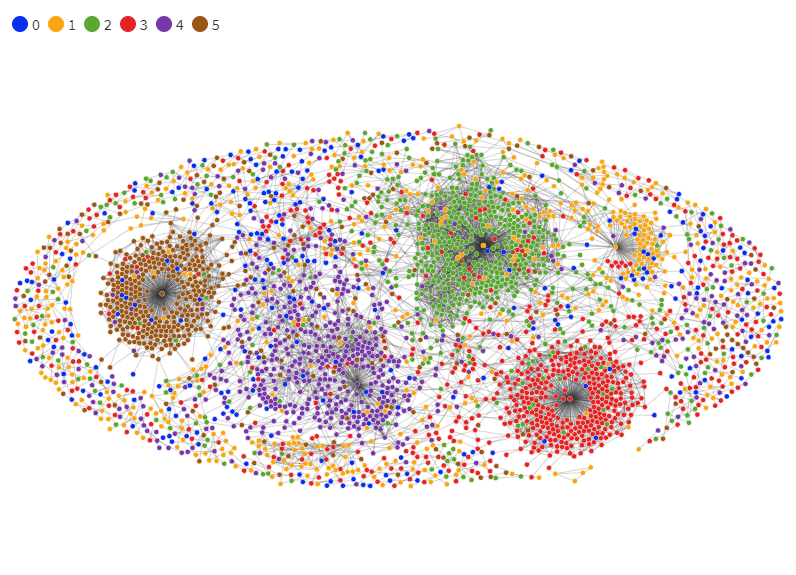}
    \caption{Epoch 200}
  \end{subfigure}
  \begin{subfigure}[b]{0.19\textwidth}
    \includegraphics[width=\linewidth]{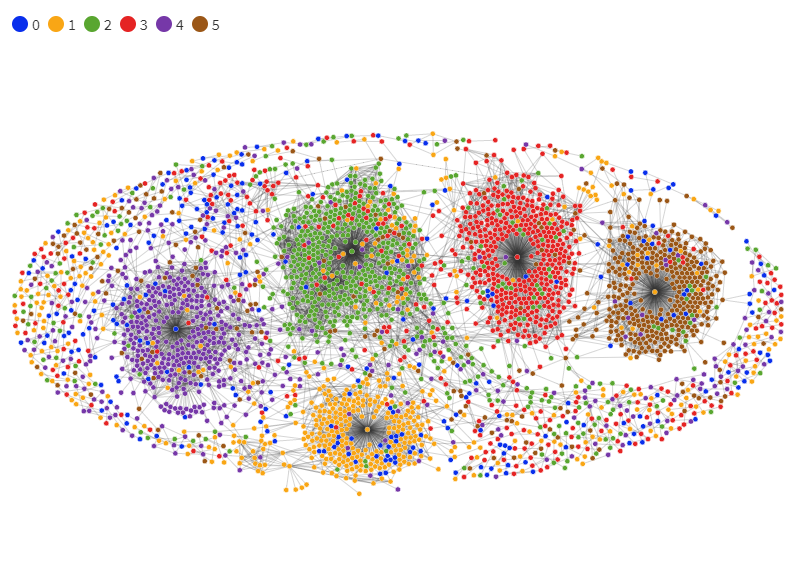}
    \caption{Epoch 300}
  \end{subfigure}
  \begin{subfigure}[b]{0.19\textwidth}
    \includegraphics[width=\linewidth]{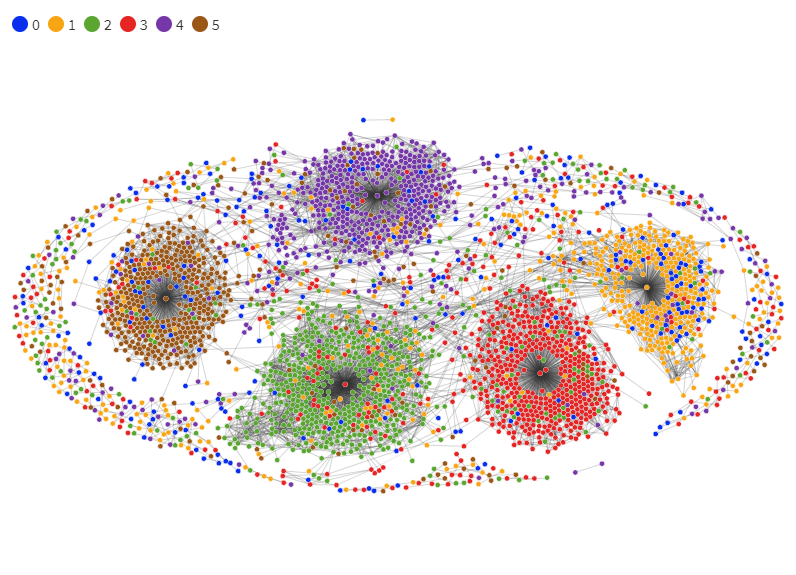}
    \caption{Epoch 400}
  \end{subfigure}
  \caption{Visualizing the graph structure $A^{pos}$, on Cora and Citeseer. Top row: results on Cora; bottom row: results on Citeseer.}
  \label{fig:vis_graphs}
\end{figure*}

\begin{figure*}[!h]
  \centering
  \begin{subfigure}[b]{0.19\textwidth}
    \includegraphics[width=\linewidth]{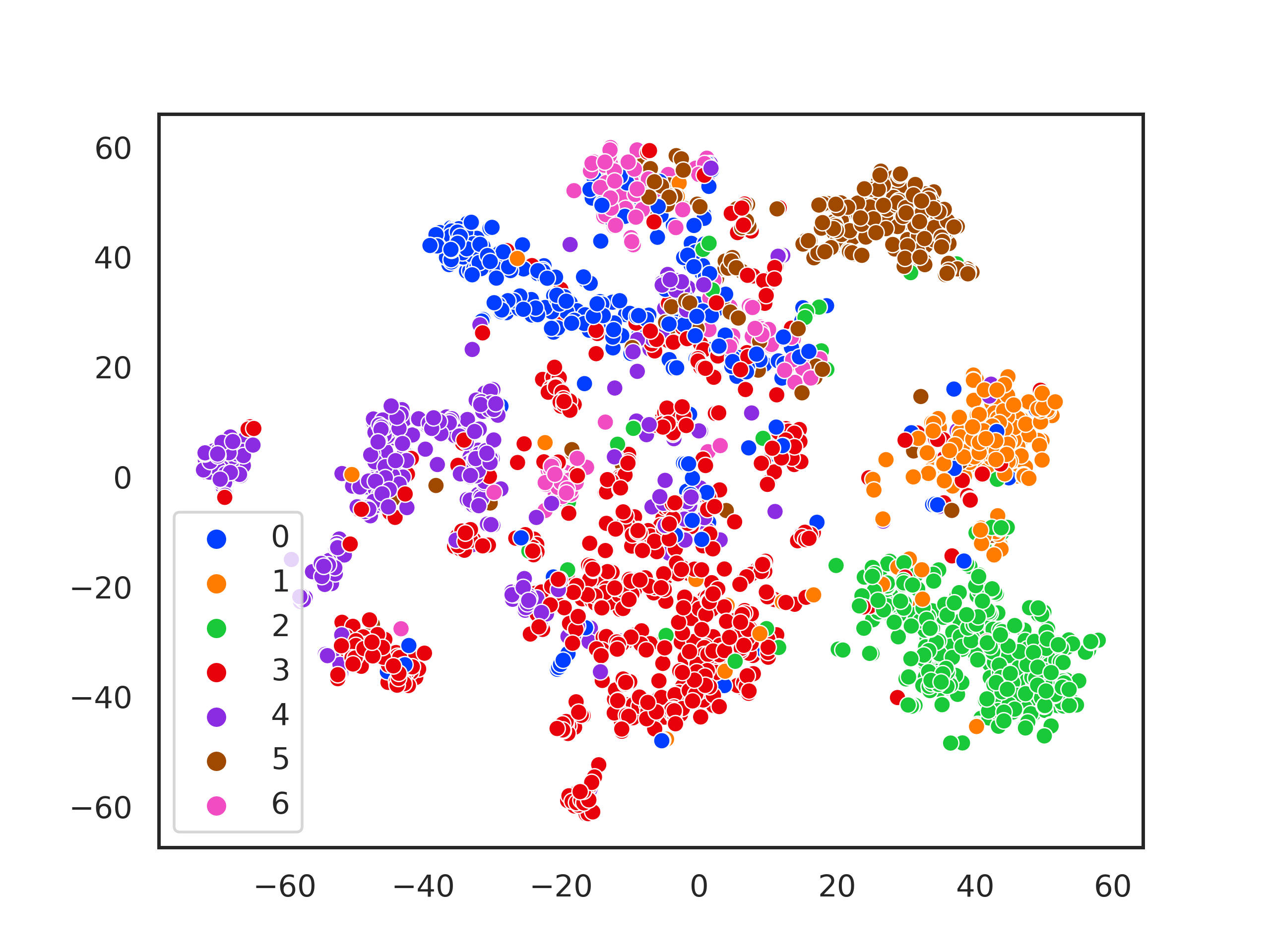}
    \caption{Epoch 0}
 \end{subfigure} \hfil
  \begin{subfigure}[b]{0.19\textwidth}
     \includegraphics[width=\linewidth]{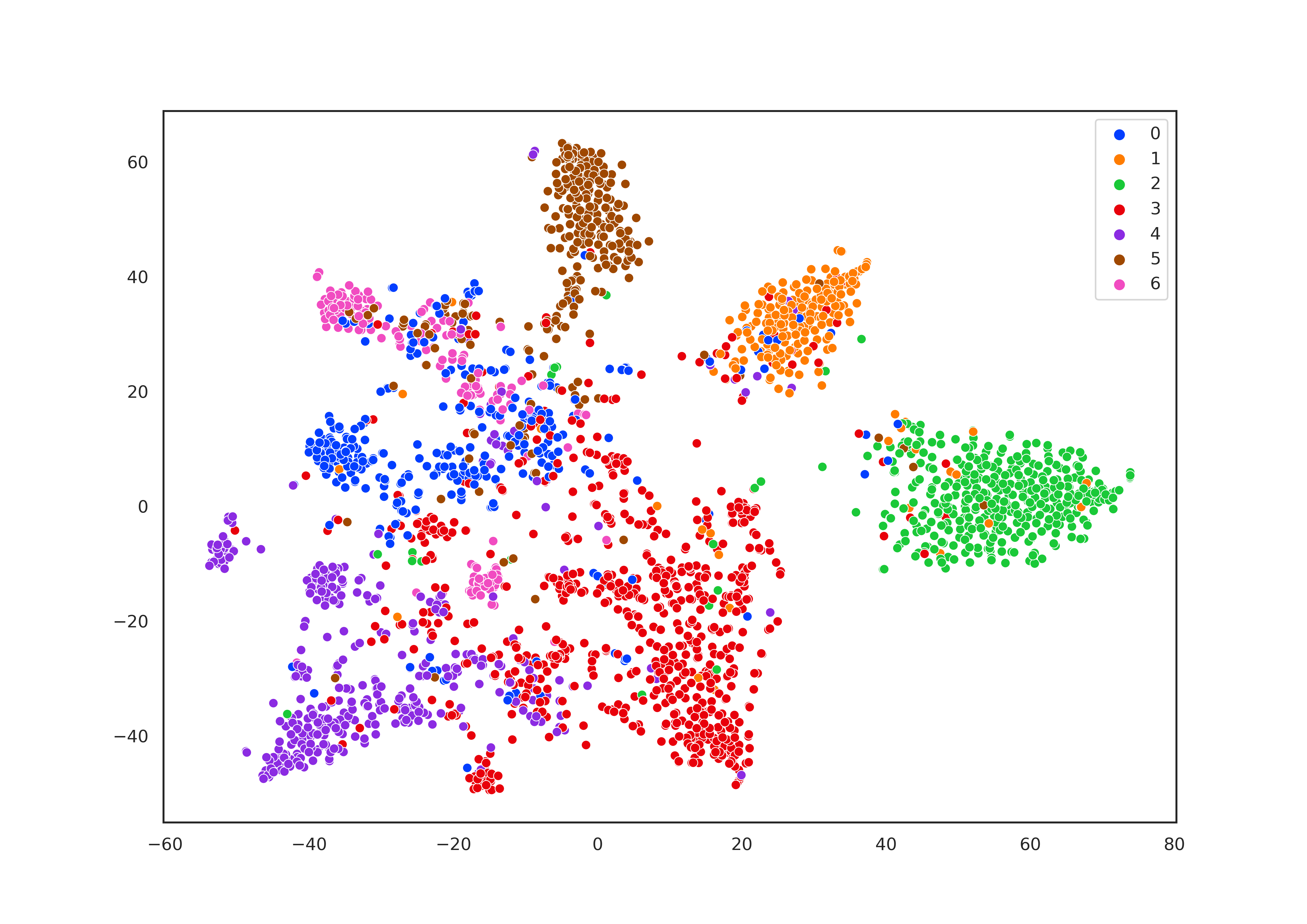}
     \caption{Epoch 200}
  \end{subfigure} \hfil
  \begin{subfigure}[b]{0.19\textwidth}
    \includegraphics[width=\linewidth]{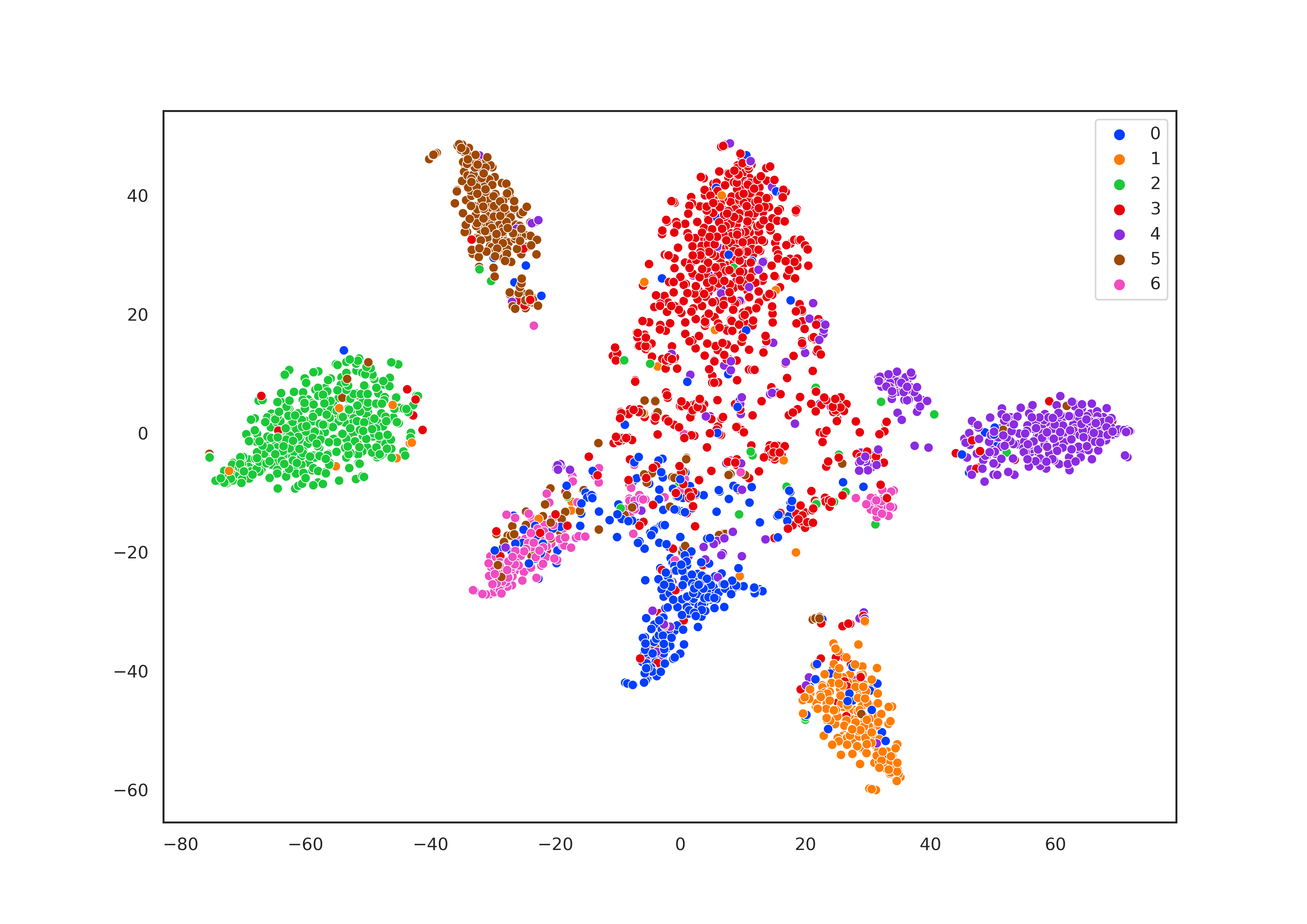}
    \caption{Epoch 400}
  \end{subfigure} \hfil
  \begin{subfigure}[b]{0.19\textwidth}
    \includegraphics[width=\linewidth]{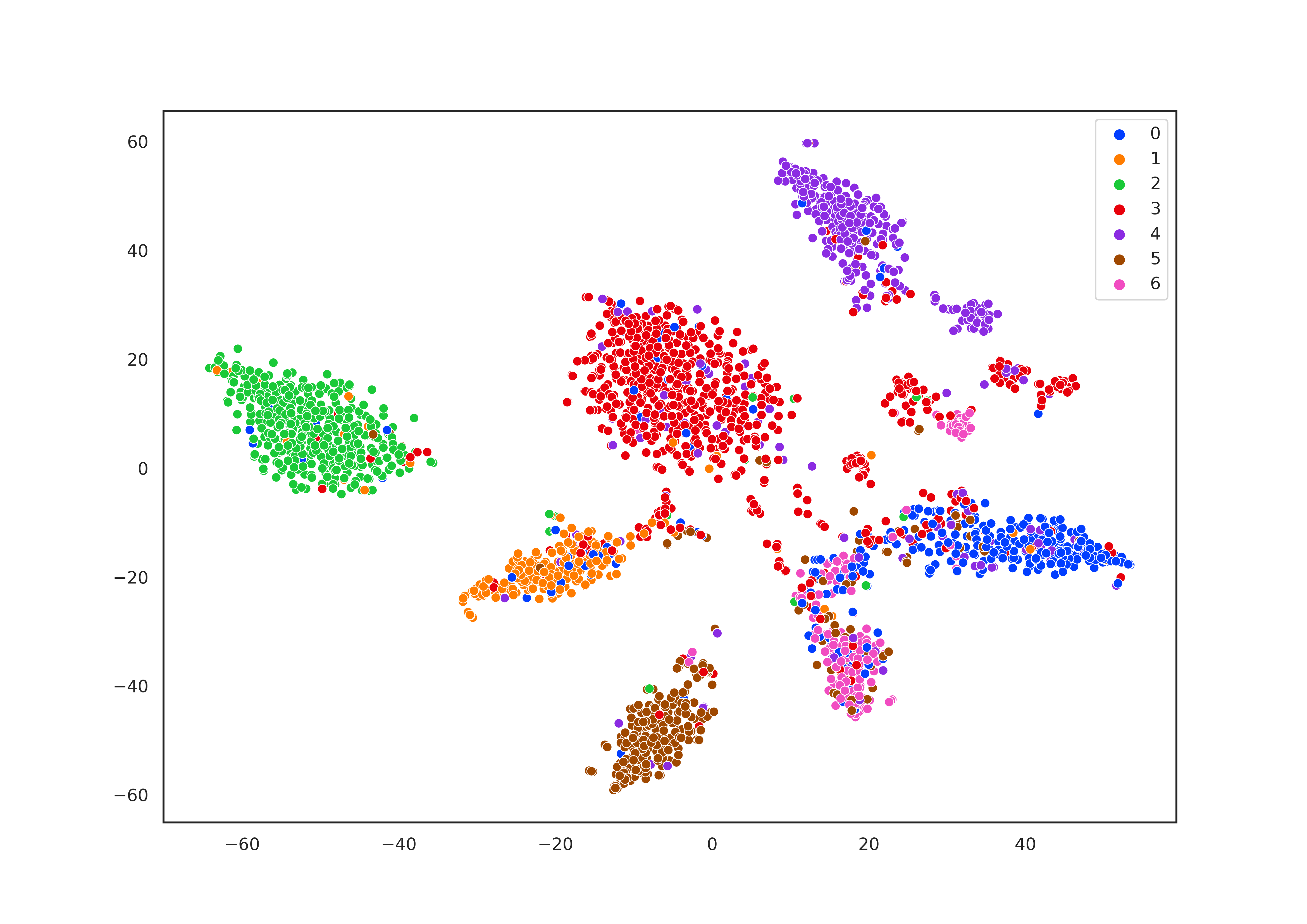}
    \caption{Epoch 600}
  \end{subfigure} \hfil
  \begin{subfigure}[b]{0.19\textwidth}
    \includegraphics[width=\linewidth]{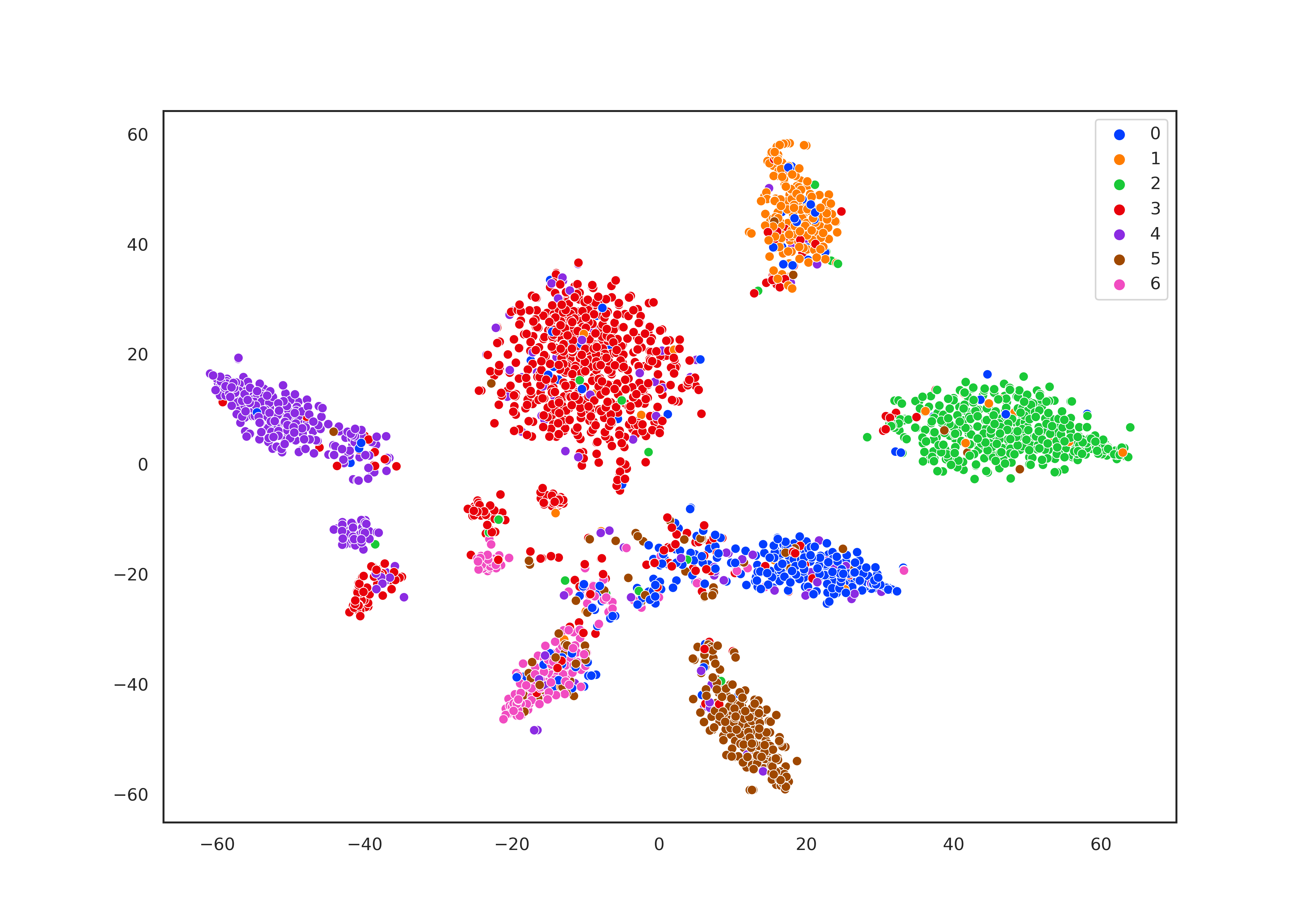}
    \caption{Epoch 800}
  \end{subfigure} \hfil
  \medskip
  \begin{subfigure}[b]{0.19\textwidth}
    \includegraphics[width=\linewidth]{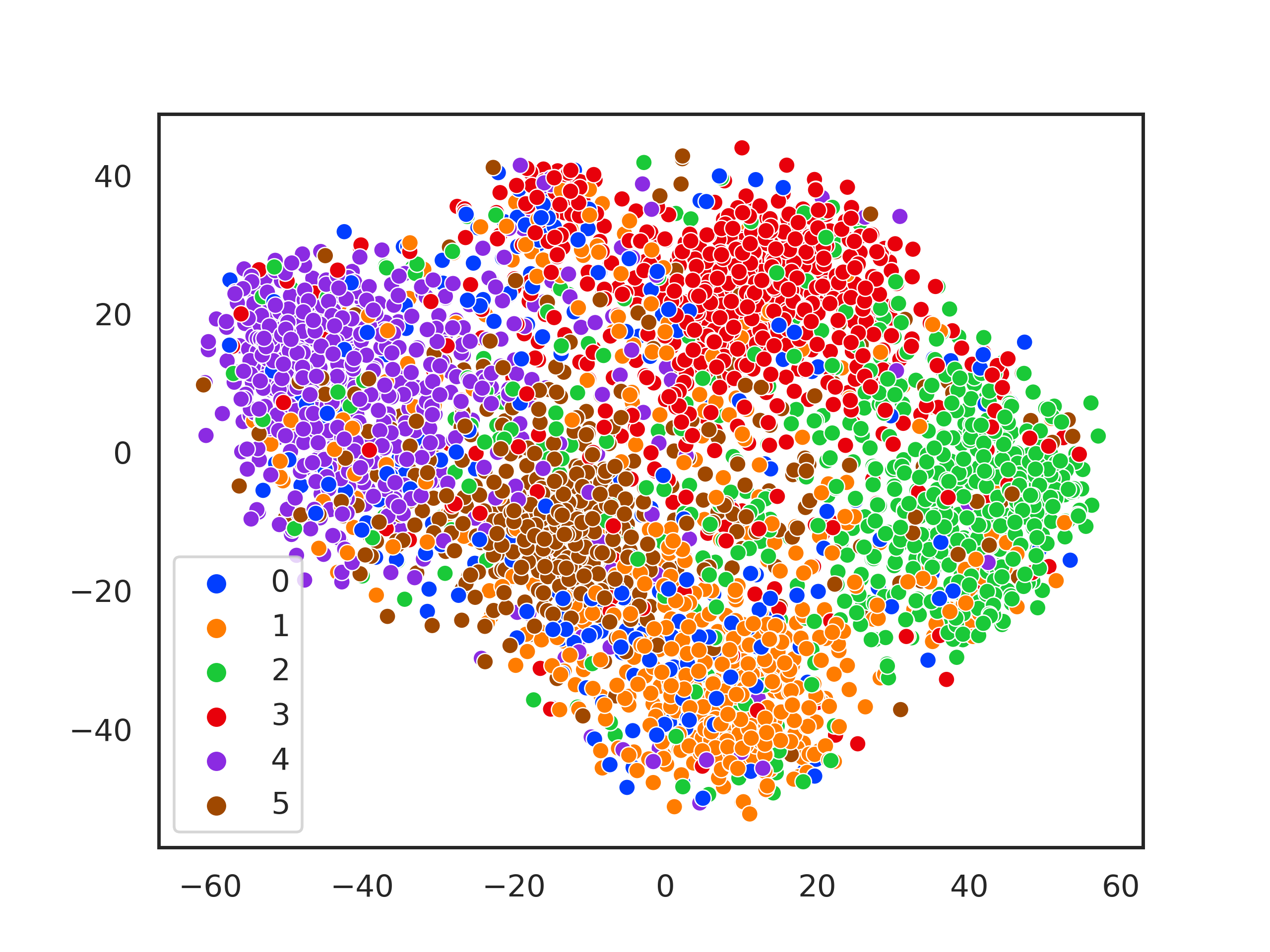}
    \caption{Epoch 0}
  \end{subfigure} \hfil
  \begin{subfigure}[b]{0.19\textwidth}
     \includegraphics[width=\linewidth]{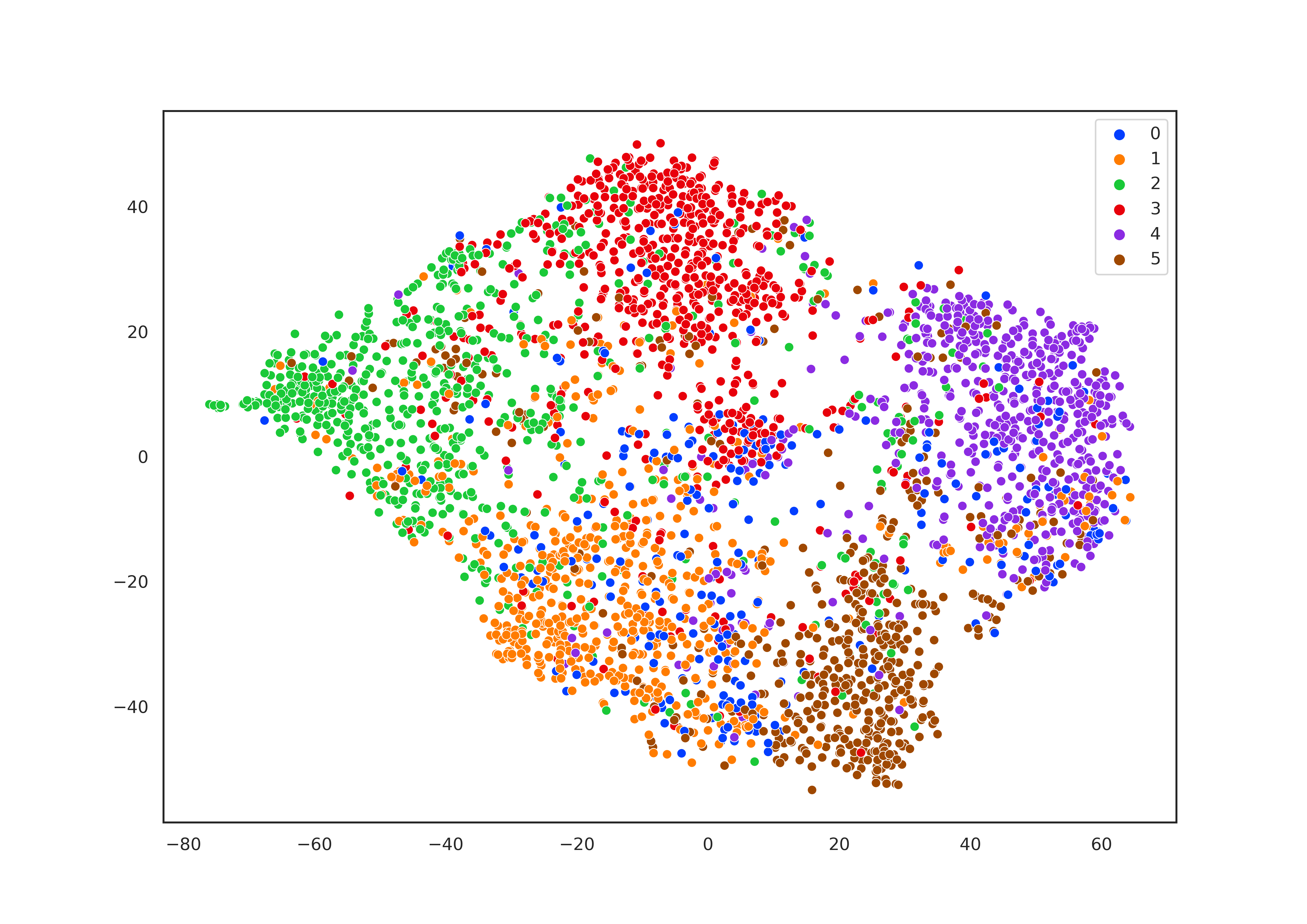}
     \caption{Epoch 100}
  \end{subfigure} \hfil
  \begin{subfigure}[b]{0.19\textwidth}
    \includegraphics[width=\linewidth]{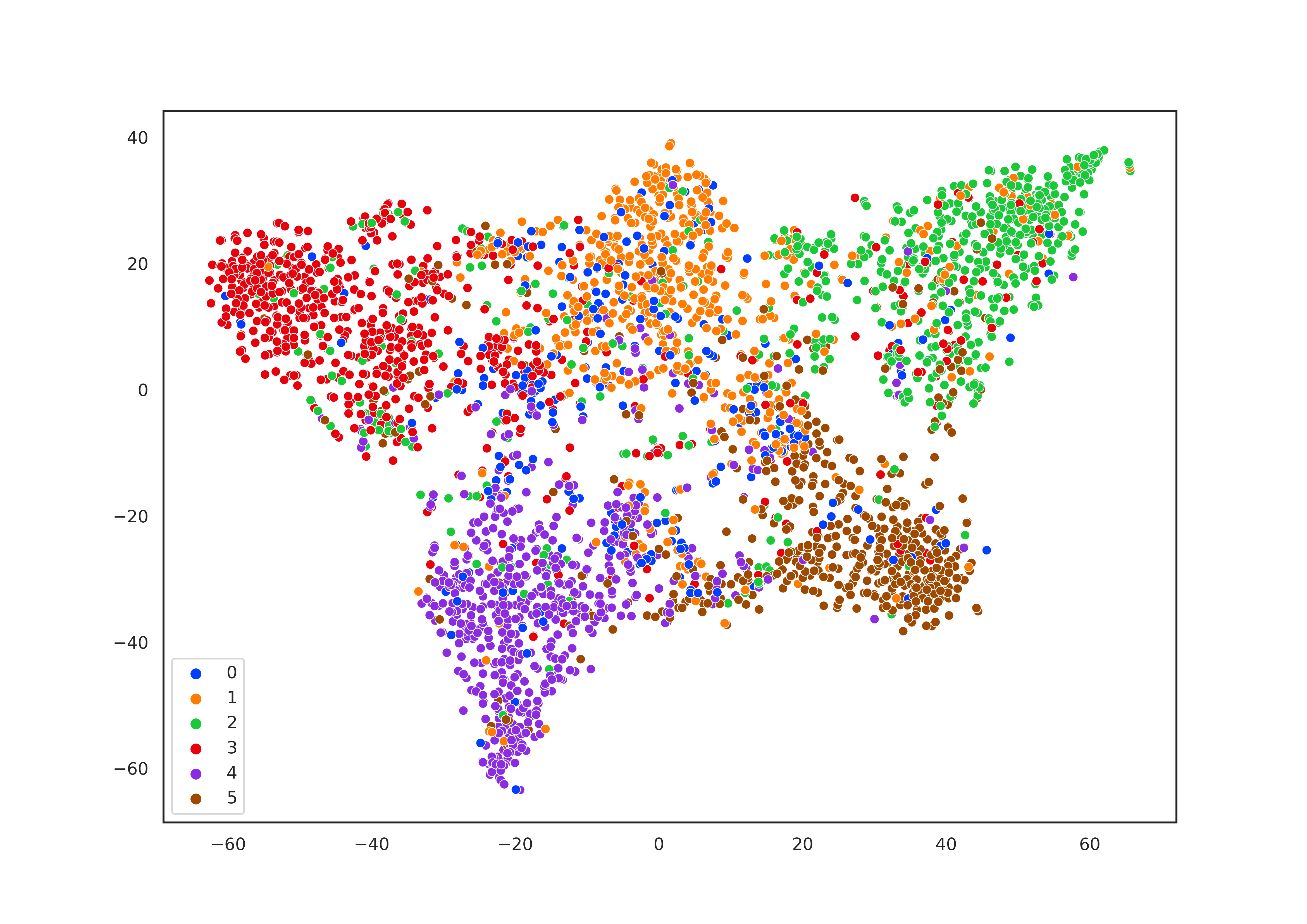}
    \caption{Epoch 200}
  \end{subfigure} \hfil
  \begin{subfigure}[b]{0.19\textwidth}
    \includegraphics[width=\linewidth]{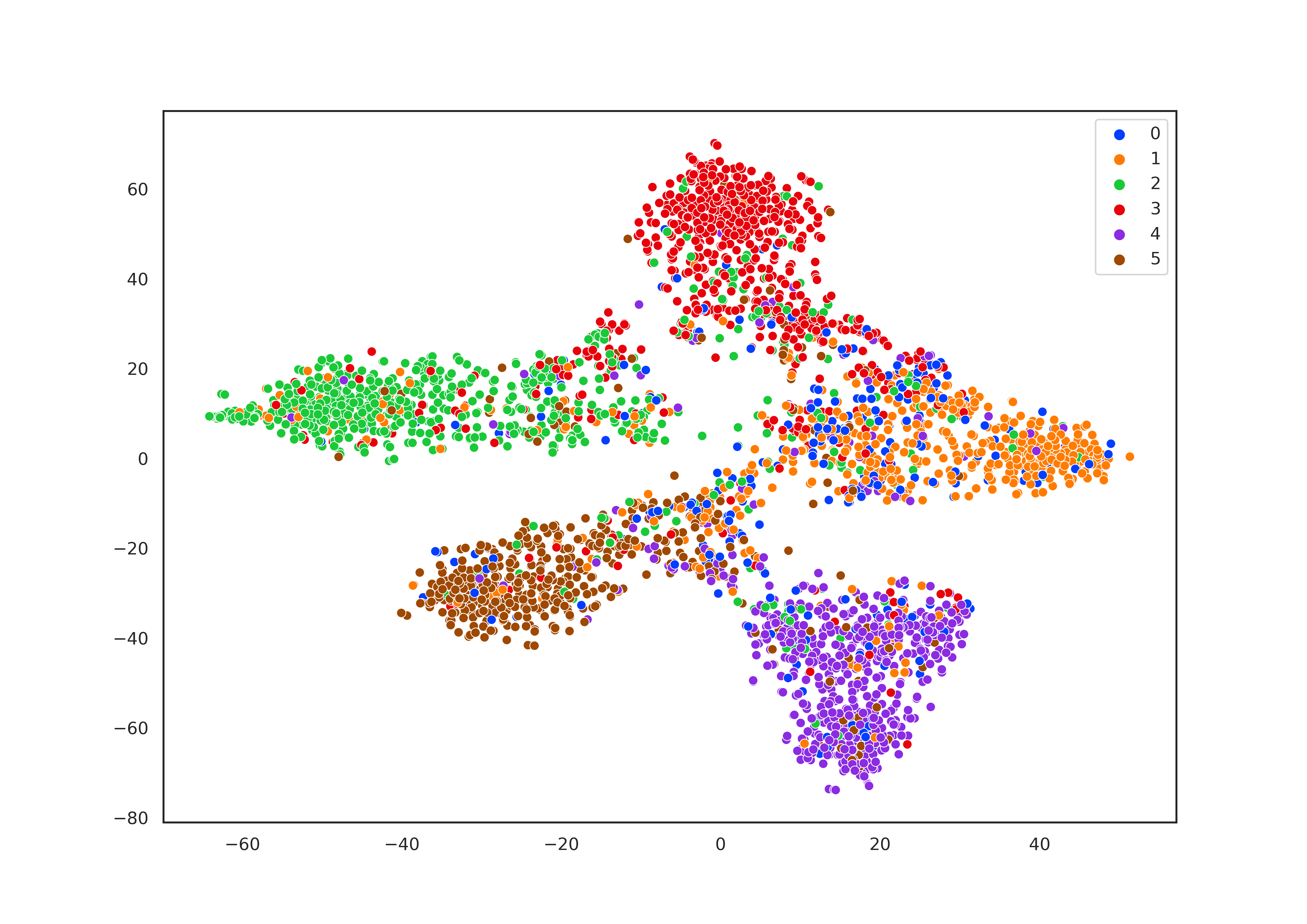}
    \caption{Epoch 300}
  \end{subfigure} \hfil
  \begin{subfigure}[b]{0.19\textwidth}
    \includegraphics[width=\linewidth]{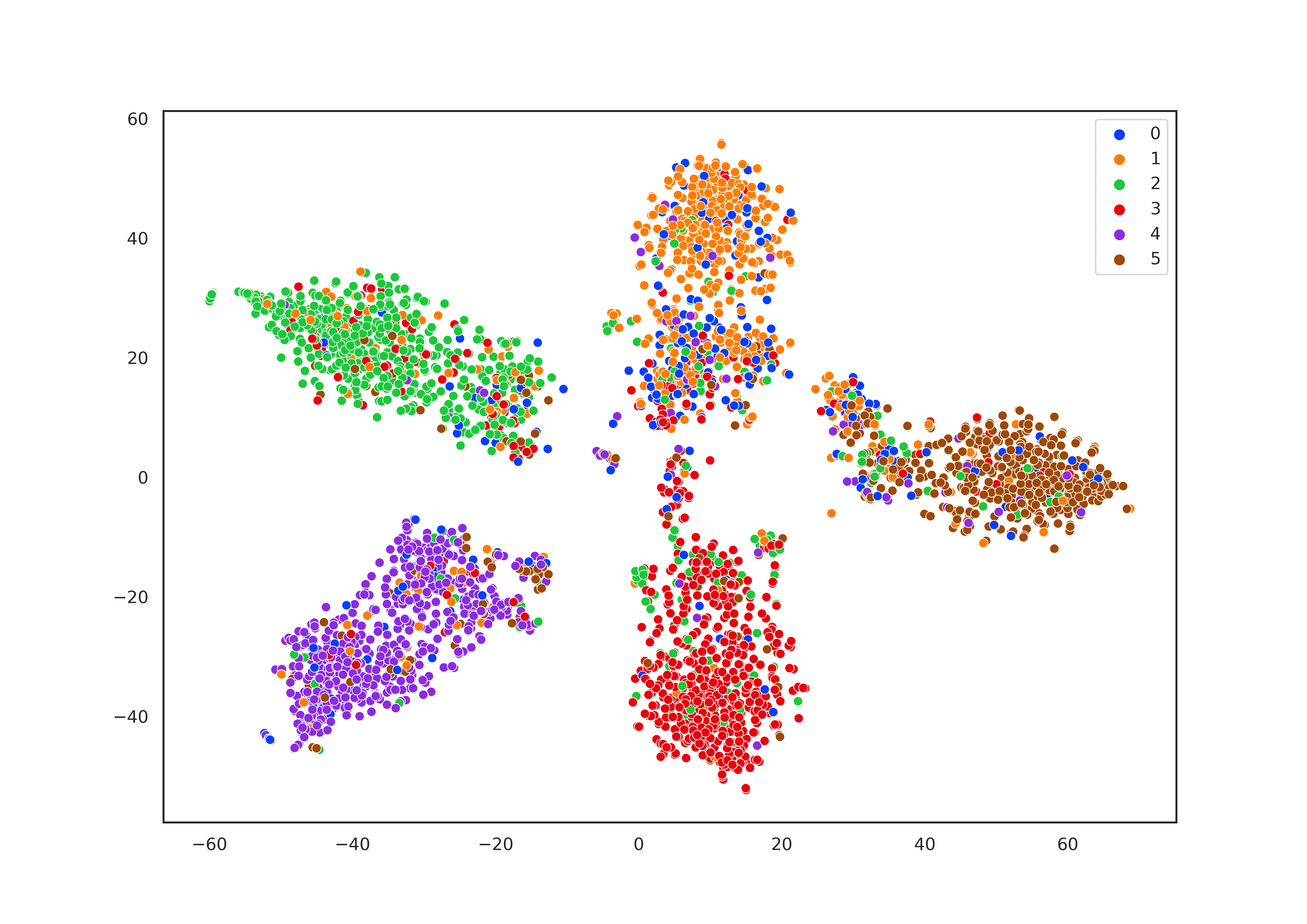}
    \caption{Epoch 400}
  \end{subfigure} \hfil
\vskip 0.1in
  \caption{T-SNE visualizations of the latent representations of CVGAE, on Cora and Citeseer. Top row: results on Cora; bottom row: results on Citeseer.}
  \label{fig:vis_tnse}
\end{figure*}

\begin{figure*}
  \centering
  \captionsetup[subfigure]{labelformat=empty}
  \begin{subfigure}[b]{0.245\textwidth}
    \includegraphics[width=\linewidth]{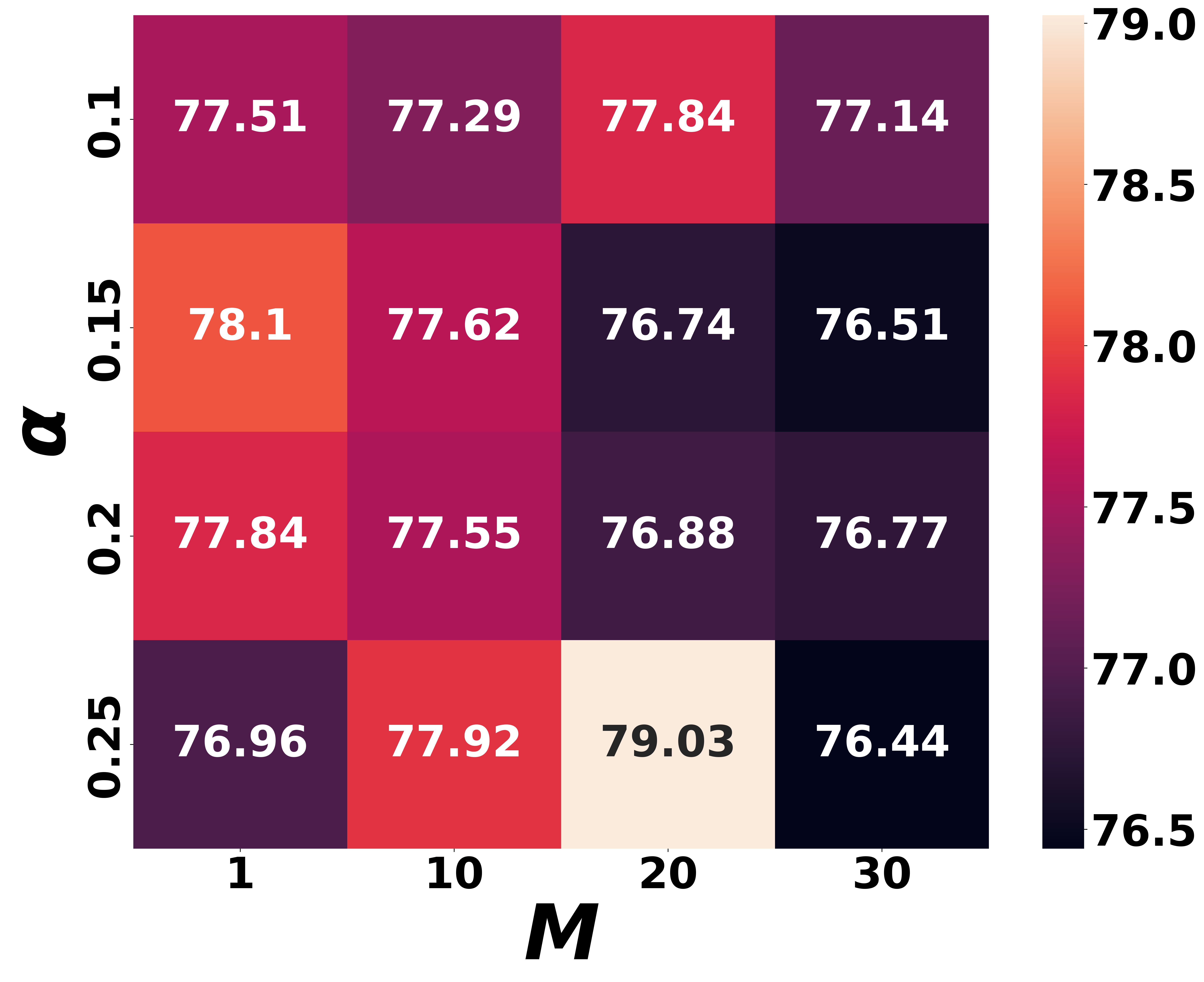}
    \caption{ACC on Cora}
  \end{subfigure}
  \begin{subfigure}[b]{0.245\textwidth}
     \includegraphics[width=\linewidth]{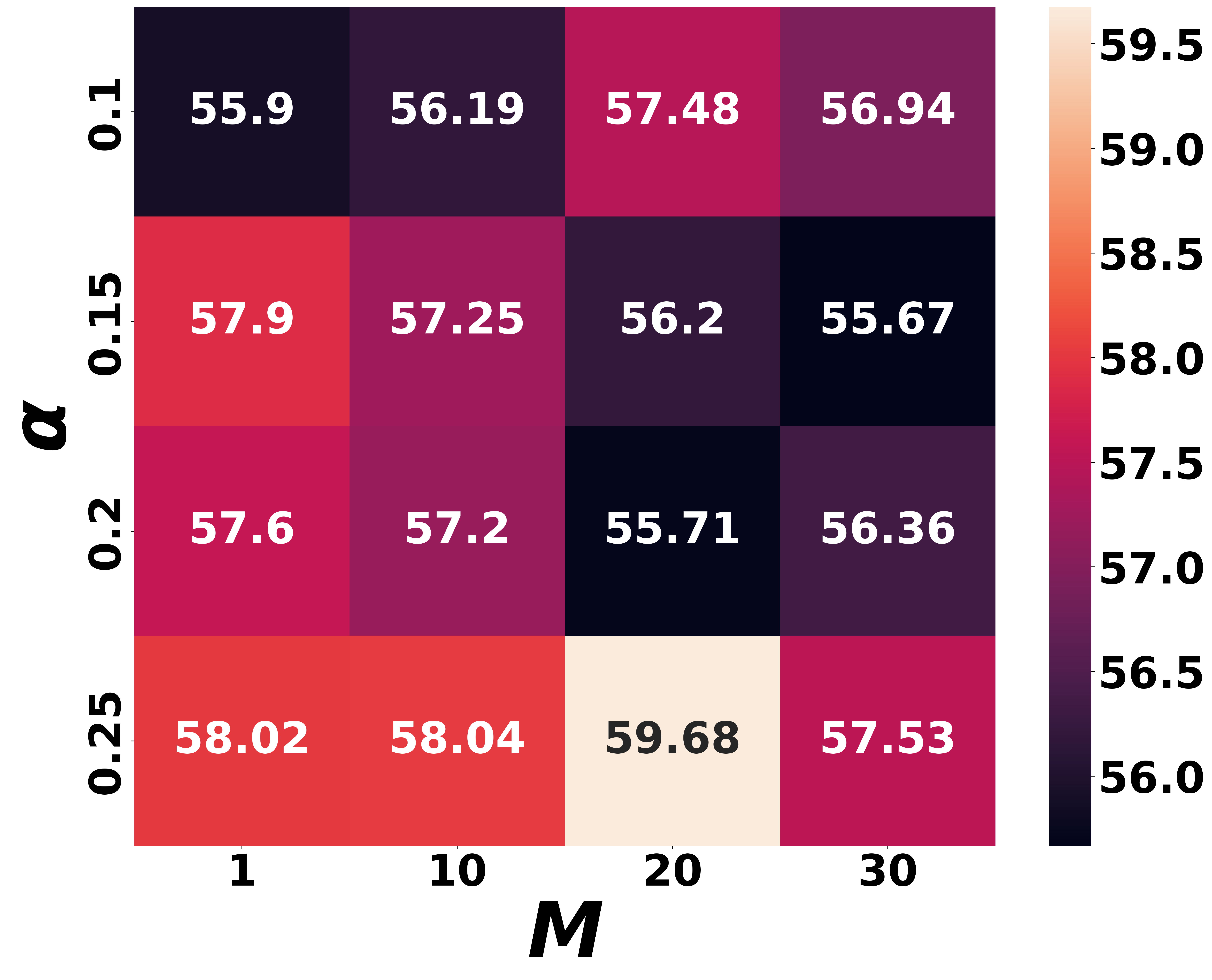}
     \caption{NMI on Cora}
  \end{subfigure}
  \begin{subfigure}[b]{0.245\textwidth}
     \includegraphics[width=\linewidth]{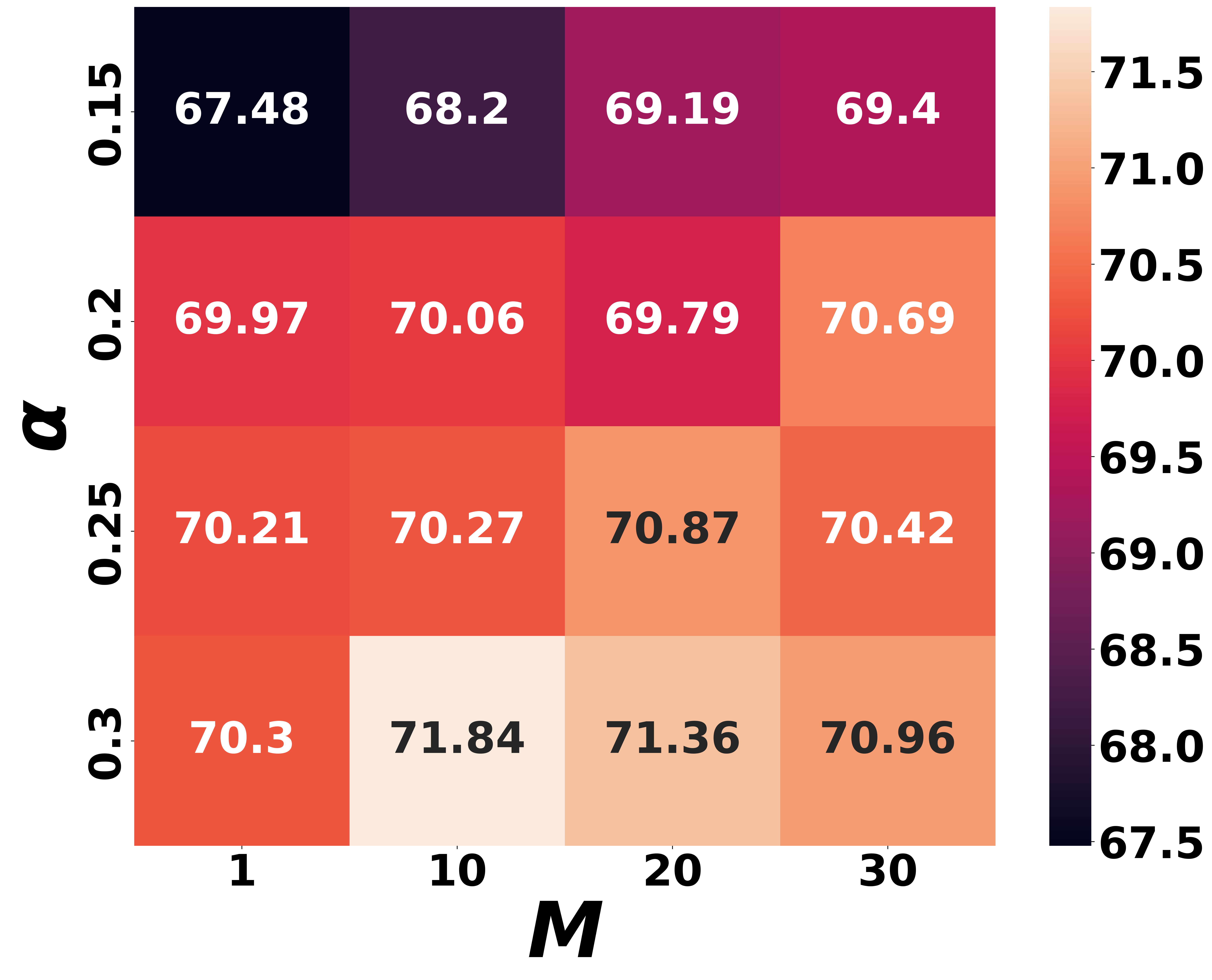}
     \caption{ACC on Citeseer}
  \end{subfigure}
  \begin{subfigure}[b]{0.245\textwidth}
     \includegraphics[width=\linewidth]{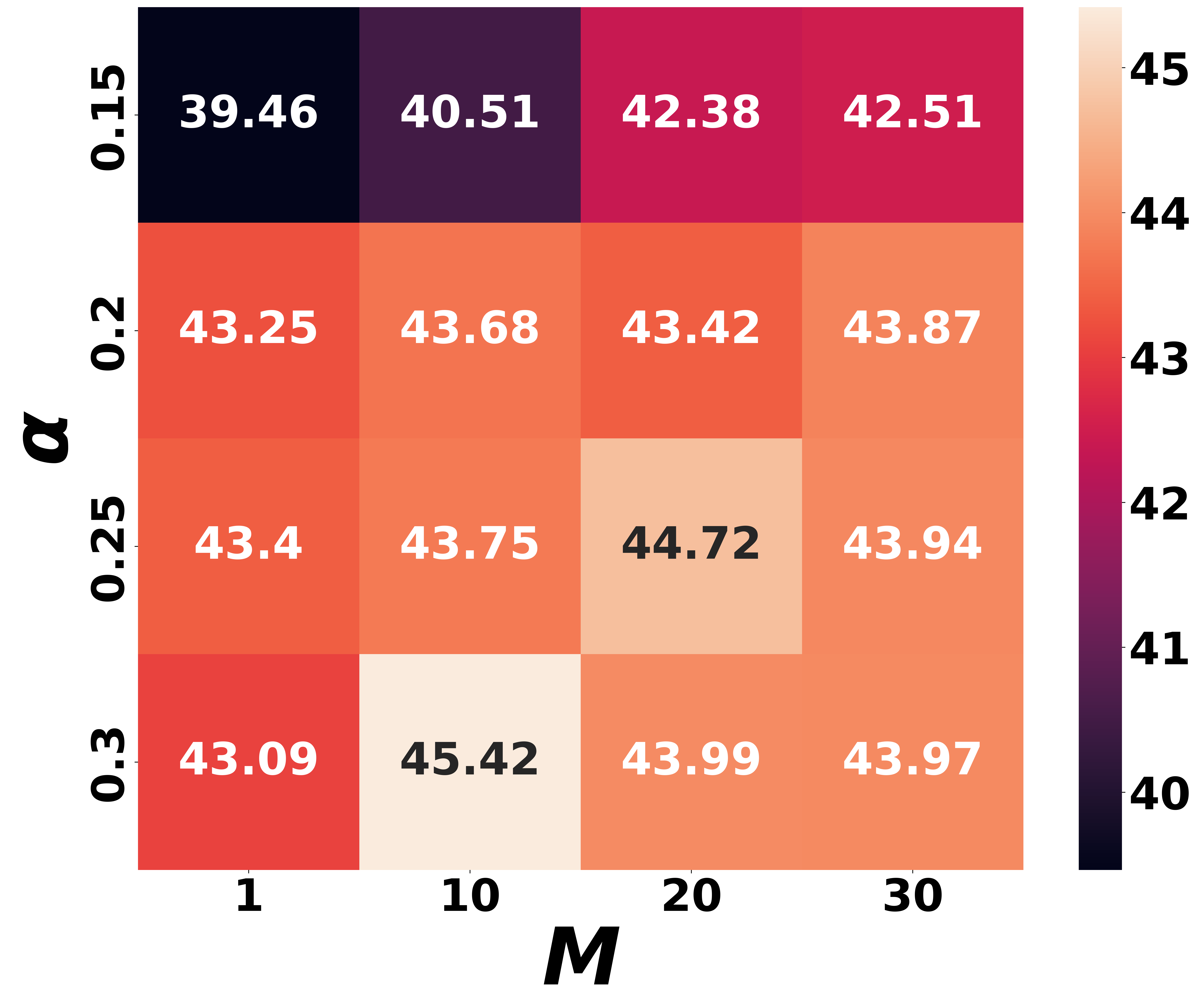}
     \caption{NMI on Citeseer}
  \end{subfigure}
  \caption{Sensitivity of CVGAE to the hyperparameters $\alpha$ and $M$ in terms of ACC and NMI.}
  \label{fig:sensitivity}
  %\vspace*{-3mm}
\end{figure*}

\textbf{Ablation study:} In Table \ref{Table:ablation_study}, we show the impact of each contribution on clustering quality. To this end, we tried several variants of our model. In particular, we have CVGAE(-FR) in the $4^{\text{th}}$ row, CVGAE(-FD) in the $6^{\text{th}}$ row, CVGAE(-CL) in the $7^{\text{th}}$ row, and CVGAE(-PC) in the $8^{\text{th}}$ row. As we can see, CVGAE outperforms all the other variants on three datasets (Cora, Citeseer, and Pubmed) in terms of ACC, NMI, and ARI. These results validate that the strong performance of CVGAE is attributed to four aspects: integrating contrastive learning and alleviating FR, FD, and PC. 

%\begin{table*}[!h]
%  \caption{Ablation study of CVGAE on Cora, Citeseer, and Pubmed.}
%  \begin{center}
%  \begin{small}
%  \scalebox{0.95}{\begin{tabular}{|c|c|c|c|c|c|c|c|c|c|c|c|}
%    \hline
%    \multicolumn{3}{|c|}{\textbf{Method}} & \multicolumn{3}{c|}{\textbf{Cora}} & \multicolumn{3}{c|}{\textbf{Citeseer}} & \multicolumn{3}{c|}{\textbf{Pubmed}}  \\
%    \cline{1-12}
%    \textbf{FR mechanism} & \textbf{FD mechanism} & \textbf{PC mechanism} & \textbf{ACC} & \textbf{NMI} & \textbf{ARI} & \textbf{ACC} & \textbf{NMI} & \textbf{ARI} & \textbf{ACC} & \textbf{NMI} & \textbf{ARI}  \\ \hline
%    \xmark  & \xmark & \xmark & 73.2 & 53.4 & 50.1 & 70.4 & 44.0 & 46.3 & 69.5 & 35.4 & 30.2 \\ \hline
%    \xmark  & \xmark & \cmark & 73.3 & 53.5 & 50.1 & 70.4 & 44.1 & 46.3 & 69.6 & 35.3 & 30.4 \\ \hline
%    \xmark  & \cmark & \xmark & 73.1 & 53.3 & 49.9 & 70.4 & 43.9 & 46.2 & 71.0 & 36.3 & 34.1 \\ \hline
%    \xmark  & \cmark & \cmark & 73.5 & 53.6 & 50.6 & 70.5 & 44.3 & 46.5 & 71.1 & 36.5 & 34.2 \\ \hline
%    \cmark  & \xmark & \xmark & 75.7 & 55.2 & 51.5 & 66.2 & 38.4 & 40.1 & 70.0 & 35.2 & 30.9 \\ \hline
%    \cmark  & \xmark & \cmark & 75.8 & 55.3 & 51.8 & 67.3 & 39.8 & 41.7 & 70.7 & 32.4 & 33.4 \\ \hline
%    \cmark  & \cmark & \xmark & 77.3 & 57.3 & 55.3 & 71.2 & 44.5 & 46.8 & 74.2 & 34.8 & 37.9 \\ \hline
%    \cmark  & \cmark & \cmark & \textbf{79.0} & \textbf{59.4} & \textbf{59.1} & \textbf{71.7} & \textbf{45.5} & \textbf{47.8} & %\textbf{74.4} & \textbf{34.8} & \textbf{38.2} \\ \hline
%  \end{tabular}}
%  \end{small}
%  \end{center}
%  \label{Table:ablation_study}
%\end{table*}

\textbf{Visualization:} We perform a qualitative study to support the quantitative findings. In Figure \ref{fig:vis_graphs}, we illustrate the structure of the clustering-oriented graph $\mathcal{G}^{pos}$ during the training process. At first, the initial structure lacks clustering-oriented edges and has several noisy links. As training progresses, the clustering structures become significantly more pronounced. More precisely, we observe that different nodes in each cluster are connected to a central node (cluster representative). Additionally, several edges that connect nodes in different clusters disappear. In Figure \ref{fig:vis_tnse}, we illustrate 2D T-SNE visualizations of CVGAE latent codes. The results obtained confirm that our model gradually increases the separation and compactness of the different clusters.

\textbf{Sensitivity analysis:} Our model has two data-dependent hyperparameters ($\alpha$ and $M$). All the other hyperparameters do not depend on the input data. In figure \ref{fig:sensitivity}, we illustrate the sensitivity of CVGAE to these data-dependent hyperparameters. Changing $\alpha$ and $M$ within the ranges $[0.1, 0.15, 0.2, 0.3]$ and $[1, 10, 20, 30]$, respectively, does not drastically affect the clustering results.

%\begin{figure*}[!h]
%  \begin{subfigure}[b]{0.33\textwidth}
%    \includegraphics[width=\linewidth]{images/hyperparameters_acc_cora.png}
%    \caption{Cora}
%  \end{subfigure}
%  \begin{subfigure}[b]{0.33\textwidth}
%     \includegraphics[width=\linewidth]{images/hyperparameters_acc_citeseer.png}
%     \caption{Citeseer}
%  \end{subfigure}
%  \begin{subfigure}[b]{0.33\textwidth}
%     \includegraphics[width=\linewidth]{images/hyperparameters_acc_pubmed.png}
%     \caption{Pubmed}
%  \end{subfigure}
%  \begin{subfigure}[b]{0.33\textwidth}
%    \includegraphics[width=\linewidth]{images/hyperparameters_nmi_cora.png}
%    \caption{Cora}
%  \end{subfigure}
%  \begin{subfigure}[b]{0.33\textwidth}
%     \includegraphics[width=\linewidth]{images/hyperparameters_nmi_citeseer.png}
%     \caption{Citeseer}
%  \end{subfigure}
%  \begin{subfigure}[b]{0.33\textwidth}
%     \includegraphics[width=\linewidth]{images/hyperparameters_nmi_pubmed.png}
%     \caption{Pubmed}
%  \end{subfigure}
%  \caption{Clustering results with different values of $\beta$ and $T_{u}$. Top row:  ACC results; bottom row: NMI %results.}
%  \label{fig:sens_hyperparameter_gamma_GMM-VGAE}
%\end{figure*}

\section{Conclusion}
This article proposes a novel variational lower bound of the graph log-likelihood function. Our method incorporates the clustering inductive bias in a principled way. Unlike the previous variational auto-encoders, CVGAE establishes a contrastive learning framework to account for the discrepancy between the inference and generative models after integrating the clustering inductive bias. Additionally, our lower bound is designed in a way to mitigate three problems, namely, Posterior Collapse, Feature Randomness, and Feature Drift. Theoretically, we find that the proposed formulation constitutes a tighter lower bound than the corresponding evidence lower bound. Empirically, our results show that the proposed model outperforms state-of-the-art graph variational auto-encoders and graph self-supervised methods in the node clustering task. Furthermore, our results provide strong evidence that our model can mitigate Feature Randomness, Feature Drift, and Posterior Collapse. Finally, our ablation study shows that the improvement in clustering results is imputed to four different aspects: introducing contrastive learning and alleviating Feature Randomness, Feature Drift, and Posterior Collapse. 

\bibliographystyle{IEEEtran}
\bibliography{CVGAE.bbl}

\end{document}